\documentclass{article}

\usepackage[nonatbib,preprint]{neurips_2020}
\usepackage[utf8]{inputenc} 
\usepackage[T1]{fontenc}    
\usepackage{hyperref}       
\usepackage{url}            
\usepackage{booktabs}       
\usepackage{amsfonts}       
\usepackage{nicefrac}       
\usepackage{microtype}      

\usepackage{graphicx}

\usepackage{amsmath,amssymb,bbm,amsthm}
\usepackage{comment}
\usepackage{epsfig} 
\usepackage{multicol}
\usepackage{multirow}
\usepackage{color}
\usepackage{mathrsfs}
\usepackage{xparse}
\usepackage{algorithm}
\usepackage{algorithmicx}
\usepackage{algpseudocode}
\usepackage{wrapfig}
\usepackage[font=small,labelfont=bf,tableposition=top]{caption}
\usepackage[font=footnotesize]{subcaption}

\newtheorem{theorem}{Theorem}
\newtheorem{lemma}{Lemma}

\newtheorem{definition}{Definition}
\newtheorem{corollary}{Corollary}

\usepackage{mathtools}

\usepackage{optidef}
\usepackage{gensymb}

\usepackage{etoolbox}
\newtoggle{comments}
\settoggle{comments}{true} 
\iftoggle{comments}{
  \newcommand {\alberto}[1]{{\color{orange}{~Alberto: #1}\normalfont}}
  \newcommand {\bjin}[1]{{\color{blue}{~Baihong: #1}\normalfont}}
  \newcommand {\yuxin}[1]{{\color{red}{~Yuxin: #1}\normalfont}}
  \newcommand {\yingshui}[1]{{\color{cyan}{~Yingshui: #1}\normalfont}}
  \newcommand {\xyyue}[1]{{\color{cyan}{~Xiangyu: #1}\normalfont}}
  \newcommand {\red}[1]{{\color{red}{#1}\normalfont}}
}{
  \newcommand {\alberto}[1]{{}}
  \newcommand {\bjin}[1]{{}}
  \newcommand {\yuxin}[1]{{}}
  \newcommand {\yingshui}[1]{{}}
  \newcommand {\xyyue}[1]{{}}
  \newcommand {\red}[1]{{}}
}

\usepackage{acronym}
\acrodef{ML}{Machine Learning}
\acrodef{AI}{Artificial Intelligence}
\acrodef{TTA}{Test-time Augmentation}
\acrodef{SL}{Severity Level}
\acrodef{i.d.}{in-distribution}
\acrodef{o.o.d.}{out-of-distribution}
\acrodef{DL}{Deep Learning}
\acrodef{KL}{Kullback-Leibler}
\acrodef{DSE}{Design Space Exploration}

\acrodef{FPR}{False Positive Rate}
\acrodef{FNR}{False Negative Rate}
\acrodef{TPR}{True Positive Rate}
\acrodef{TNR}{True Negative Rate}

\acrodef{FNP}{False Negative Precision}
\acrodef{AUC}{Area Under Curve}
\acrodef{FDD}{Fault Detection and Diagnosis}

\acrodef{CNN}{Convolutional Neural Network}

\newcommand{\denselist}{\itemsep 0pt\topsep-6pt\partopsep-6pt}

\newcommand{\expct}[1]{\mathbb{E}\left[#1\right]}

\newcommand{\abs}[1]{\left\vert#1\right\vert}

\newcommand{\rvdiff}{Z}

\title{Exploiting Uncertainties from Ensemble Learners to Improve Decision-Making in Healthcare AI}

\author{
  Yingshui Tan$^1$\thanks{Equal Contributions}\quad Baihong Jin$^{1,3\ast}$ \quad Xiangyu Yue$^1$\\
  \textbf{Yuxin Chen}$^2$\quad \textbf{Alberto Sangiovanni-Vincentelli}$^1$ \\
  $^1$ University of California, Berkeley\\ 
  $^2$ University of Chicago 
  $^3$ Lawrence Berkeley National Laboratory
}

\begin{document}

\maketitle

\begin{abstract}
Ensemble learning is widely applied in \ac{ML} to improve model performance and to mitigate decision risks. In this approach, predictions from a diverse set of learners are combined to obtain a joint decision. Recently, various methods have been explored in literature for estimating decision uncertainties using ensemble learning; however, determining which metrics are a better fit for certain decision-making applications remains a challenging task. In this paper, we study the following key research question in the selection of uncertainty metrics: when does an uncertainty metric outperforms another? We answer this question via a rigorous analysis of two commonly used uncertainty metrics in ensemble learning, namely \emph{ensemble mean} and \emph{ensemble variance}. We show that, under mild assumptions on the ensemble learners, ensemble mean is preferable with respect to ensemble variance as an uncertainty metric for decision making. We empirically validate our assumptions and theoretical results via an extensive case study: the diagnosis of referable diabetic retinopathy.
\end{abstract}

\section{Introduction}

\acf{DL} has revolutionized the field of computer vision, and has found wide applications in commerce, security and healthcare. Still, models make mistakes, and the ability to evaluate the uncertainties associated with a \ac{DL} model decisions is of paramount importance as high prediction accuracy itself, especially for healthcare applications where mistakes can be very costly.

In recent years, we have seen a stream of work proposed in the \ac{DL} community advocating the use of ensemble models to estimate prediction uncertainties. Popular approaches in this category include deep ensemble~\cite{lakshminarayanan2017simple}, Monte-Carlo (MC) dropout~\cite{gal2016dropout,jin2019augmenting}, and \ac{TTA}~\cite{ayhan2018test,wang2018automatic}.
These methods can be used for 1) identifying potentially misclassified examples and 2) detecting \ac{o.o.d.} examples, making them powerful tools for managing the risks in using AI for decision-making.

In this paper, we investigate an issue that often arises in medical image diagnostic models: how to deal with ``incipient diseases''---early stages of a disease that develops in a progressive manner~\cite{jin2020ensemble,blennow2003csf}. The symptoms of such diseases naturally distribute over a ``severity spectrum'' ranging from ``healthy'' to ``severe''. As can be expected, the incipient disease cases  suffer from higher risks of being wrongly classified. On top of that, these cases are often high-stake ones as well since early identification of incipient diseases is of great value by itself since preventive measures can be taken to avoid  severe consequences. Therefore, it is important to develop methods to mitigate the risks of misclassifying incipient diseases. 

When using AI for diagnostic applications, false negative decisions are often concerning and costly. If a disease case is mistaken for an healthy condition, the patient may miss the best time for treatment and may suffer from more severe consequences. We consider an uncertainty-informed human-AI collaborative scheme, where a set of ``uncertain negative'' decisions are first identified based on the negative decisions made by an ensemble model, and then human diagnosticians are involved to re-examine these uncertain cases to identify false negatives. We formulate our research problem under the above scheme, and aim to design approaches to maximize the number of \textit{corrected} wrong predictions. The contributions of this paper are three:
\begin{itemize}\denselist
    \item We compare and analyze two commonly used uncertainty metrics, one based on the ensemble mean and the other based on the ensemble variance. Our theoretical analysis shows that the uncertainty metric based on ensemble mean is preferable to that based on ensemble variance. 
    \item To validate our theoretical findings, we conducted a case study on  diabetic retinopathy, and evaluated the performance of three ensemble approaches and two uncertainty metrics. Our experimental results confirmed our theoretical findings and showed that the number of false negative decisions can be greatly reduced with our proposed \textit{stacking ensemble} approach at a modest increase in human re-examination efforts.
    \item Our comparative studies on three ensemble methods also suggest that incipient diseases as a type of ambiguous inputs may represent a more difficult case to use decision uncertainties for decision making than the case of \ac{o.o.d.} inputs. 
\end{itemize}

\section{Problem Formulation}\label{sec:formulation}

We formulate the uncertainty-informed disease diagnosis problem in a binary classification setting. Consider an ensemble $\mathcal{E}$ comprising a diverse set of binary classifiers, $\mathcal{M}^{(1)},\mathcal{M}^{(2)},\ldots,\mathcal{M}^{(K)}$, that have been trained for the same diagnosis task. Let  $z_i\in\{0,1\}$  represent the ground-truth label of input $x_i$, and  $\hat{y}_i^{(k)}$  denote the output of the $k$th classifier where $\hat{y}_i^{(k)}\in[0,1]$. By using a threshold $\tau$ to dichotomize the continuous output $y_i^{(k)}$, each classifier $\mathcal{M}^{(k)}$  produces a predicted class label $\hat{z}_i^{(k)}$ for input $x_i$.

To identify false negatives in the classification, we use prediction uncertainties with an \textit{uncertainty metric} $U$ for ranking the negative examples\footnote{Examples that are classified as negative, i.e. $\{x_i\,\vert\,\hat{z}_i=0\}$.}. The uncertainty metric $U: \mathbb{R}^K\rightarrow \mathbb{R}$ takes as input the ensemble predictions on $x_i$, and outputs an \textit{uncertainty score} $s(x_i)\doteq U(y_i^{(1)},y_i^{(2)},\ldots,y_i^{(K)})$. The interpretation of $s(x_i)$ depends on the task. In our application, we seek to utilize prediction uncertainties to identify false negative decisions: a higher $s(x_i)$ indicates that $x_i$ is more likely to be a false negative decision.

As in classification tasks, we need a threshold to resolve a dichotomy between ``uncertain''' and ``certain'; we define the \textit{uncertain negative rate} as a percentage $q\%$ such that only a fraction $q\%$ of the negative examples will be deemed \textit{uncertain negatives}. To evaluate how many of the uncertain negatives are actually false negatives, we define the following performance measure,
\begin{definition}[False Negative Precision]
We define the \acf{FNP} $R_\text{FNP}(U,q)$ to be the ratio of false negative decisions among uncertain negative decisions, under a given uncertainty metric $U$ and a given uncertain negative rate $q\%$. The \ac{FNP} takes a value between 0 and 1. Written in mathematical form, 
$R_\text{FNP} \doteq 
{\left\vert\left\{x_i\,\vert\, x_i\in\mathcal{I}^{-}_{q\%},z_i=1\right\}\right\vert}/
{\left\vert\mathcal{I}^{-}_{q\%}\right\vert}\in[0,1],$ where $\mathcal{I}^{-}_{q\%}$ is the index set of negative examples whose uncertainty scores are within the top-$q$ percentile.
\end{definition}
The \ac{FNP} metric can also be interpreted as the probability of an identified uncertain example being an actual false negative. Our goal in this paper is to rigorously analyze several commonly used uncertainty metrics (see Section \ref{sec:methodology}). Formally, we seek the uncertainty metric $U$ that maximizes $R_\text{FNP}(U,q)$, i.e., $U^*=\arg\max_U R_\text{FNP}(U,q)$.

\section{Ensemble-based Uncertainty Estimation Methods}\label{sec:methodology}

\subsection{Ensemble Methods for Uncertainty Estimation}\label{sec:ensemble-method}

We first give a brief overview of popular \ac{DL}-based ensemble methods  that can be used for estimating decision uncertainties.
The two methods described below can be categorized into the class of \textit{explicit ensemble} models, where a diverse set of individual models are combined to make a joint decision:

\noindent\textbf{Deep Ensemble} Proposed by Lakshminarayanan~et~al.~\cite{lakshminarayanan2017simple,fort2019deep}, a \textit{deep ensemble} is made up of multiple neural networks of the same architecture; the individual learners are diversified by random initialization as well as random shuffling of training examples. The deep ensemble method has been shown to be effective in detecting \ac{o.o.d.} inputs from image datasets.

\noindent\textbf{Stacking Ensemble}
\textit{stacking ensemble}~\cite{wolpert1992stacked,sikora2015modified} is a more general ensemble approach for \ac{DL}, where the outputs from a wide range of decision models can be ``stacked'' into an ensemble decision. The stacking ensemble is a flexible approach that is amenable to real-world applications since little restriction is imposed on individual learners.

The next two methods fall under the category of \textit{implicit ensemble} methods~\cite{huang2017snapshot} due to their ``train once get many for free'' nature, where multiple predictions can be generated from one single trained model; the diversity among predictions either comes from the stochasticity inherent to the network (as in MC-dropout) or from perturbations to the input data (as in \ac{TTA}). 

\noindent\textbf{Monte Carlo Dropout (MC-dropout)} Dropout~\cite{srivastava2014dropout} is a popular and powerful regularization technique to prevent overfitting neural network parameters. 
Recently, Gal and Ghahramani proposed using MC-dropout~\cite{gal2016uncertainty} to estimate a network's prediction uncertainty by using dropout not only at training time but also at test time. By sampling a dropout model $\mathcal{M}$ using the same input for $T$ times, we can obtain an ensemble of prediction results with $T$ individual probability vectors. The dropout technique provides an inexpensive approximation to training and evaluating an ensemble of exponentially many similar yet different neural networks.

\noindent\textbf{\acf{TTA}} Similar to the MC-dropout technique, a network with \ac{TTA}~\cite{ayhan2018test,wang2018automatic} produces a different result each time we ``sample'' the same network with the same given input $x$. Different from MC-dropout networks, \ac{TTA} adds randomness to the test input $x$ through data augmentation as is often performed during training, e.g., adjustment of brightness, image cropping and image flipping. This creates an ensemble of exponentially many predictors as in MC-dropout networks.

Implicit ensemble methods are considered appealing due to the reduced training costs since only one model needs to be trained. However, the use of explicit ensembles itself does not incur much additional cost in reality~\cite{zhou2012ensemble}, as compared to single learners or implicit ensembles. The development of \ac{ML} models (including implicit ensembles) usually involves \ac{DSE}, e.g., architecture search~\cite{zoph2016neural}, hyperparameter tuning~\cite{bardenet2013collaborative}, (training-time) data augmentation~\cite{wong2016understanding} and K-fold cross-validation~\cite{rodriguez2009sensitivity}. The model instances generated during the \ac{DSE} processes can be used to construct explicit ensembles; in this respect, the advantages of implicit ensembles over explicit ones is not so significant.
\vspace{-3mm}

\subsection{Metrics for Capturing and Ranking Prediction Uncertainties}\label{sec:metrics}
A number of metrics have been proposed in literature for estimating the prediction uncertainties of ensemble learners. In Lakshminarayanan~et~al.'s paper~\cite{lakshminarayanan2017simple}, the metrics are broadly classified into two categories: \textit{confidence-based} and \textit{disagreement-based} metrics. The former is designed to capture the consensus of the individual learners in an ensemble, while the latter is designed to measure the degree of disagreement among their predictions; however, the two seemingly unrelated goals can have a significant  overlap. In this paper, we propose a more rigorous categorization for these uncertainty metrics depending on their mathematical forms to unveil their differences and to enable further analyses. Some metrics (hereafter referred to as type-1 metrics) rely only on the ensemble output $\hat{y}_i^e$, while others (referred to as type-2 metrics) take all single learner's outputs into account.

Type-1 metrics use the ensemble output $\hat{y}_i^e$ to compute the confidence level, without the need to know what the individual predictions are. A negative aspect of these metrics is that the disagreement among individual learners can be hidden beneath the ensemble output $\hat{y}_i^e$.

\noindent\textbf{Margin Metric (\textsc{mean})}
An intuitive metric that measures the confidence of a classifier on input $x$ is the difference between the output probability and the decision threshold $\tau$. Since we use the convention that larger function values of $s_\textsc{mean}(x_i)$ corresponds to larger uncertainties, the margin metric can be formulated as $s_\textsc{mean}(x_i)=U_\textsc{mean}(\hat{y}_i^e)\doteq 1-\left\vert \hat{y}_i^e - \tau\right\vert$, which will be later referred to as \textsc{mean}.

\noindent\textbf{Binary Cross-Entropy Metric (\textsc{entropy})}
The binary cross-entropy $s_\textsc{entropy}$ as a function of $x_i$ takes the form $s_\textsc{entropy}(x_i)=U_\textsc{entropy}(\hat{y}_i^e)\doteq -\left[\hat{y}_i^e\log \hat{y}_i^e + \left(1-\hat{y}_i^e\right)\log\left(1-\hat{y}_i^e\right)\right]$.
It can be shown that the entropy metric is equivalent to the margin metric when $\tau=0.5$. The entropy metric can be useful for evaluating decision uncertainties when no decision threshold is a priori assigned.

Type-2 metrics have the potential to give a more comprehensive characterization of the individual predictions (e.g., the disagreement). The following two existing type-2 metrics that are often used in literature,  focus on quantifying the \textit{disagreement} among individual learners in an ensemble and for this reason,  address the shortcomings of type-1 metrics.
\begin{wrapfigure}{r}{0.40\textwidth}  
    \includegraphics[height=4.9cm]{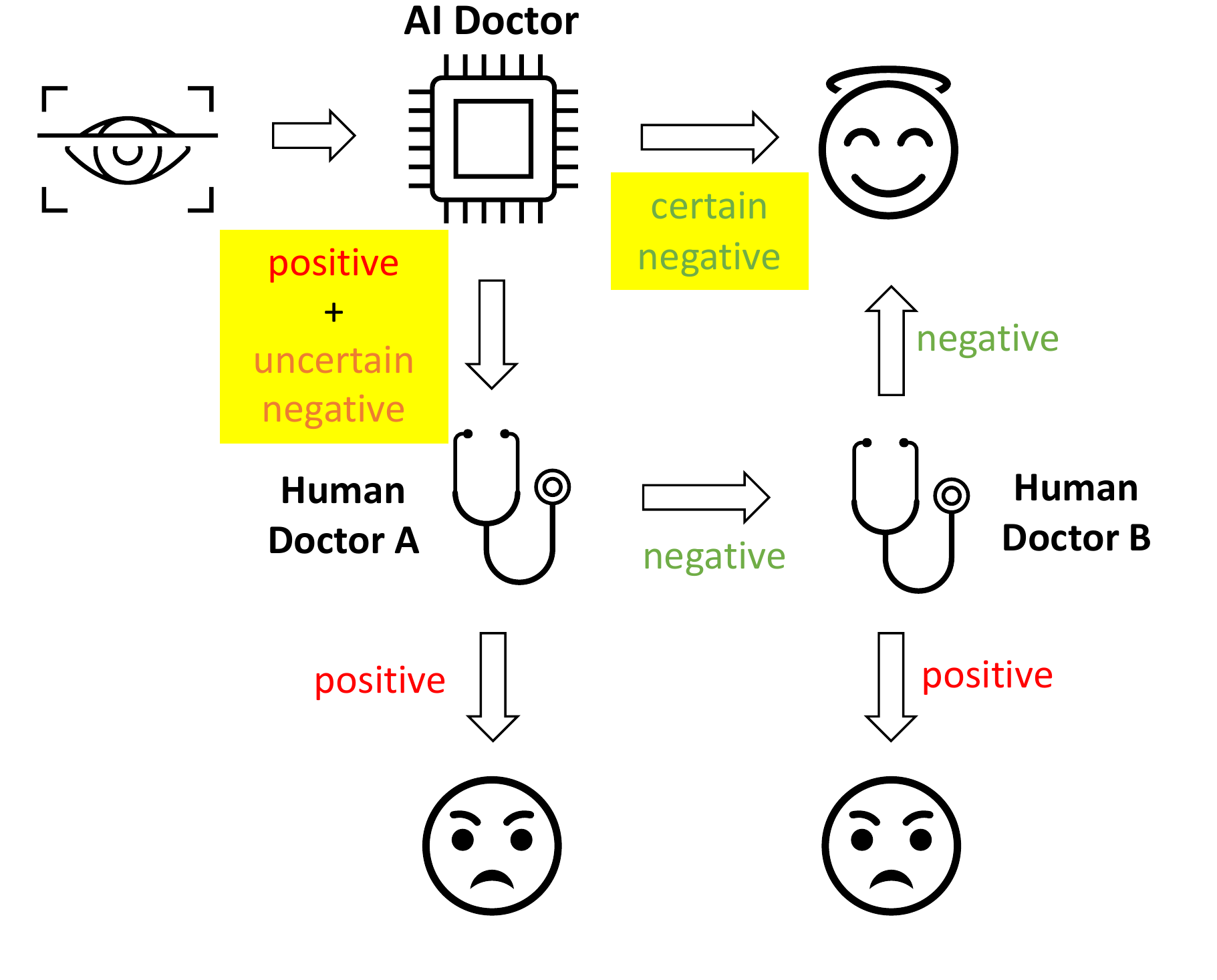}
    \caption{Illustration of the uncertainty-informed human-AI diagnostic scheme.}
    \label{fig:Human-AI-scheme}
    \vspace{-15mm}
\end{wrapfigure}

\noindent\textbf{Variance Metric (\textsc{var})}
The variance (or standard deviation) metric~\cite{leibig2017leveraging,jin2019detecting} measures how spread out the individual learners' predictions are from the ensemble prediction $\hat{y}_i^e$. The uncertainty score of input $x_i$ based on \textit{sample variance} can be written as $s_\text{Var}(x_i)=U_\textsc{var}\left(\hat{y}_i^{(1)},\hat{y}_i^{(2)},\ldots,\hat{y}_i^{(K)}\right)\doteq\frac{1}{K-1}\sum_{k=1}^K \left[\hat{y}_i^{(k)} - \hat{y}_i^e\right]$. This metric will be later referred to as \textsc{var}.


\noindent\textbf{\ac{KL} Divergence Metric (\textsc{kl})}
Similar to the variance metric, the \ac{KL} divergence metric~\cite{goldberger2003efficient} measures the deviation of individual learner's predictions from the ensemble output $\hat{y}_i^e$. The uncertainty score $s_\textsc{kl}(x_i)$ of input $x_i$ under the \ac{KL} divergence metric can be written as $s_\textsc{kl}(x_i)=U_\textsc{kl}\left(\hat{y}_i^{(1)},\hat{y}_i^{(2)},\ldots,\hat{y}_i^{(K)}\right) \doteq \frac{1}{K}\sum_{k=1}^K D_\textsc{kl}(y_i^{(k)} \Vert \hat{y}_i^e) = \sum_{k=1}^K \hat{y}_i^{(k)}\log\frac{\hat{y}_i^{(k)}}{\hat{y}_i^e}.$
This metric will be later referred to as \textsc{kl}.

A problem with \textsc{var} and \textsc{kl} is that they focus mainly on the disagreement among ensemble predictions but do not take in consideration  the value of $\hat{y}_i^e$. Consider a scenario where the all ensemble members predict a probability of $0.5$. Both \textsc{var} and \textsc{kl} will produce an uncertainty score of $0$ and thus will not be able to capture any decision uncertainties; in fact, this case where all learners give an output of $0.5$ is highly uncertain. 

\vspace{-2mm}
\subsection{Uncertainty-Informed Disease Diagnosis Scheme}\label{sec:Human-AI-scheme}
We consider an \textit{uncertainty-informed} diagnostic scheme as an application of prediction uncertainties that fosters the collaboration between human and AI diagnosticians, illustrated in Figure~\ref{fig:Human-AI-scheme}. In this scheme, an AI model is first used to screen the images. The cases diagnosed as positive will be referred to a human diagnostician, who will confirm the case as positive if she agrees with the AI (that the image corresponds to a positive case). If the AI and the human diagnostician disagree, another human diagnostician will be involved to make the final decision (as an arbitrator between AI and the first human diagnostician). For comparison, we will consider a baseline scheme where only the positive cases will be sent to human diagnosticians. The baseline scheme suffers from the problem that false negatives from the AI's diagnoses would never be reviewed by human diagnosticians.

\section{Theoretical Analysis on Uncertainty Metrics \textsc{mean} and \textsc{var}}\label{sec:theory}

In this section, we will perform theoretical analysis on two representative and commonly used uncertainty metrics, \textsc{mean} and \textsc{var}. We will compare the performance of \textsc{mean} to that of \textsc{var} under 1) an infinite-ensemble-size scenario and 2) a finite-ensemble-size scenario.

To model how different classifiers will respond to a given input $x_i$, we assume that the prediction $\hat{y}_i^{(k)}$ from classifier $\mathcal{M}^{(k)}$ is sampled from a beta distribution $\mathcal{B}(\alpha_i,\beta_i)$ that is characterized by two parameters by $\alpha_i$ and $\beta_i$. We further assume that $\alpha_i+\beta_i$ is fixed to the same constant value for all $i$'s.
Under this assumption, each input can be described by $\alpha_i$ ($\beta_i$ can be calculated since $\alpha_i+\beta_i$ is fixed), easing further analysis. The \ac{SL} of the case represented by image $x_i$ can be characterized by the parameter $\alpha_i$ . The larger the value of $\alpha_i$, the more severe the case of $x_i$ is. When $\alpha_i$ and $\beta_i$ are close, the case is ambiguous as the distribution shifts towards being symmetric (i.e. signifying much disagreement among classifiers) rather than being one-sided (i.e. consensus among classifiers 
that $x_i$ is negative or positive). We provide a set of examples in Figure~\ref{fig:beta-dist-five-SL} and also Figure~\ref{fig:beta-dist-five-SL-plus} in the supplementary materials showing how the beta distribution can be used to capture diverse predictions given by an ensemble learner.

As discussed in previous sections, the choice of uncertainty metric $U$ determines how examples are ranked and therefore affects the detection performance of false negatives. We expect the final ranking negative examples due to the uncertainty metric $U$ matches the true severity ranking given by $\alpha_i$. Taking a microscopic view into the ranking process, we consider two negative examples $x_i$ and $x_j$, and assume $x_i$ represents a less severe case than $x_j$. Under the above beta distribution assumption, we will have $\alpha_i<\alpha_j\leq\beta_j$. Our theoretical analysis will focus on the chance that $x_i$ (the less ambiguous or more normal case) is considered more uncertain than $x_j$ (the more ambiguous case). If the following theorem holds, then those correctly ranked by \textsc{var} are also likely to be correctly ranked by \textsc{mean}, indicating that \textsc{mean} is a preferable uncertainty metric to \textsc{var}.

\begin{lemma}\label{lm:sample-uncertainty}
Consider two inputs $x_i, x_j$ with uncertainty score $s(x_i)$ and $s(x_j)$ estimated from $n$ \emph{i.i.d.} ensemble learners, and denote by $\Delta_{ij}(s) := \mathbb{E}[s(x_j)-s(x_i)]$ the difference of expected uncertainty score.
If $\Delta_{ij}(s) > 0$, then $P\left(s(x_i)>s(x_j)\right) = \mathcal{O}\left(\frac{\text{Var}(s(x_i)+\text{Var}(s(x_j))}{n \Delta_{ij}^2(s)}\right)$.
\end{lemma}
The proof is deferred to the supplemental materials in Section~\ref{sec:lemma}. Intuitively, Lemma~\ref{lm:sample-uncertainty} states that if input $x_j$ is more uncertain than $x_i$ w.r.t. the expected uncertainty $\mathbb{E}[s(\cdot)]$, then the probability of the sample uncertain measure $s$ making a wrong decision is bounded. Based on such result, we establish the following error bounds for uncertainty metrics \textsc{mean} and \textsc{var}:
\begin{theorem}\label{thm:mean-vs-var-beta-finite}
Consider inputs $x_i$, $x_j$, with $y_i \sim \mathcal{B}(\alpha_i,\beta_i)$, $y_j \sim \mathcal{B}(\alpha_j,\beta_j)$, and $\alpha_i+\beta_i = \alpha_j+\beta_j$. Let $\Delta_{ij}(s) := \mathbb{E}[s(x_j)-s(x_i)]$ where $s(\cdot)$ denotes an uncertainty score estimated from $n$ \emph{i.i.d.} ensemble learners. If $\alpha_i < \alpha_j \leq \beta_j$, then $\Delta_{ij}(s_\textsc{mean}) > \Delta_{ij}(s_\textsc{var}) > 0$. Furthermore, it holds that $P\left(s_\textsc{mean}(x_i) > s_\textsc{mean}(x_j)\right) = \mathcal{O}\left(\frac{1}{n \Delta_{ij}^2(s_\textsc{mean})}\right)$ and $P\left(s_\text{Var}(x_i) > s_\text{Var}(x_j)\right) = \mathcal{O}\left(\frac{1}{n \Delta_{ij}^2(s_\textsc{var})}\right)$.  
\end{theorem}

The proof is given in Section~\ref{sec:mean-vs-var-beta-finite} of the supplementary materials. 
Theorem~\ref{thm:mean-vs-var-beta-finite} implies that if $x_i$ is more likely to be positive than $x_j$, then for ensemble learners of fixed size, the upper bound on the probability of $s_\textsc{mean}$ making a wrong decision is lower. In other words, $s_\textsc{mean}$ is likely to be a more robust measure than $s_\textsc{var}$.

A direct corollary of the above theorem states that under infinite ensemble size,  using either \textsc{mean} or \textsc{var} as the uncertainty metric does not make a difference. 
\begin{corollary}\label{thm:mean-vs-var-beta}
If sample size is infinite, then under the conditions of Theorem~\ref{thm:mean-vs-var-beta-finite}, we have $s_\textsc{mean}(x_i)<s_\textsc{mean}(x_j) \Leftrightarrow s_\text{Var}(x_i)<s_\text{Var}(x_j)$.
\end{corollary}

\section{Experimental Results}\label{sec:results}

\begin{figure}[t]
  \centering
  \begin{subfigure}[t]{0.19\linewidth}
    \centering
    \includegraphics[height=2cm]{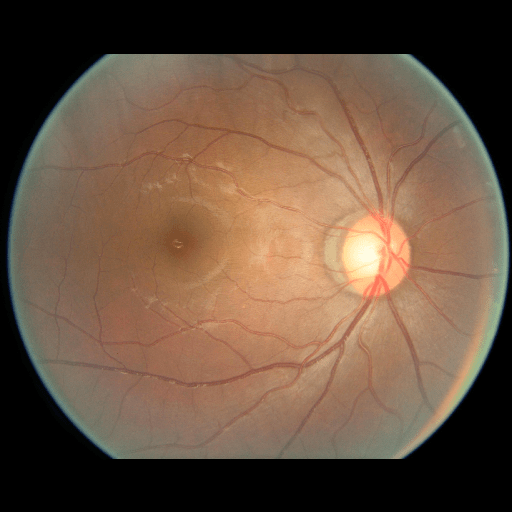}
    \includegraphics[trim=0.3cm 0.9cm 0cm 0.85cm,clip, height=2cm]{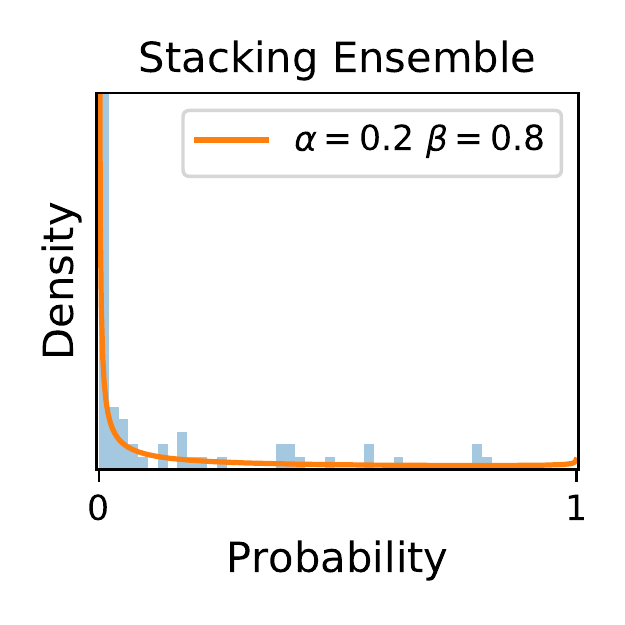}
    \includegraphics[trim=0.3cm 0.9cm 0cm 0.85cm,clip, height=2cm]{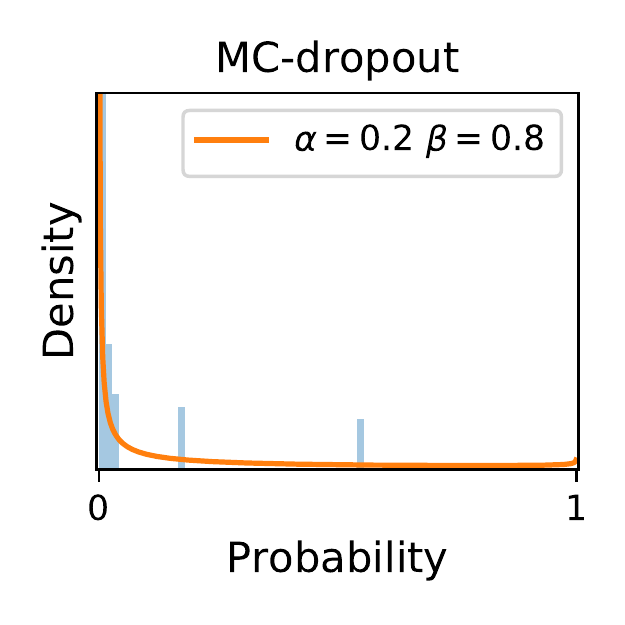}
    \includegraphics[trim=0.3cm 0cm 0cm 0.85cm,clip, height=2.45cm]{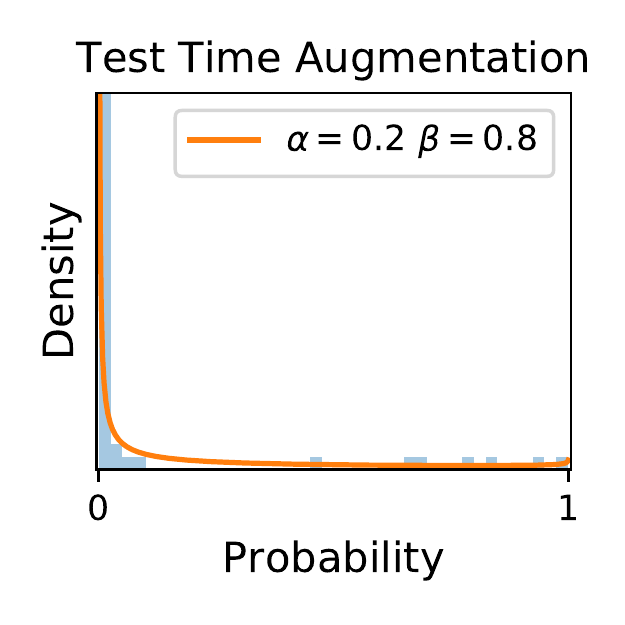}
    \caption{SL0 (No-DR)}
  \end{subfigure}
  \begin{subfigure}[t]{0.19\linewidth}
    \centering
    \includegraphics[height=2cm]{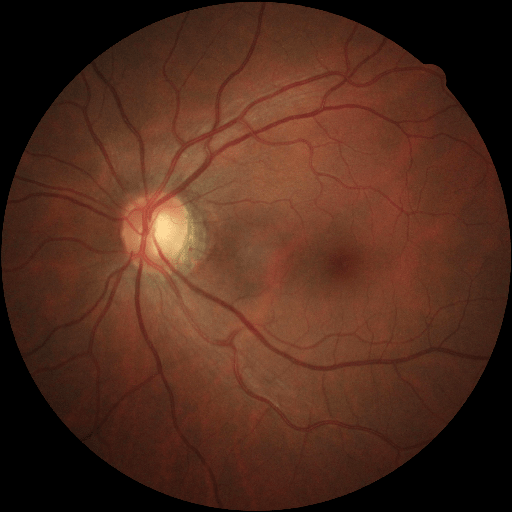}
    \includegraphics[trim=0.85cm 0.9cm 0cm 0.85cm,clip, height=2cm]{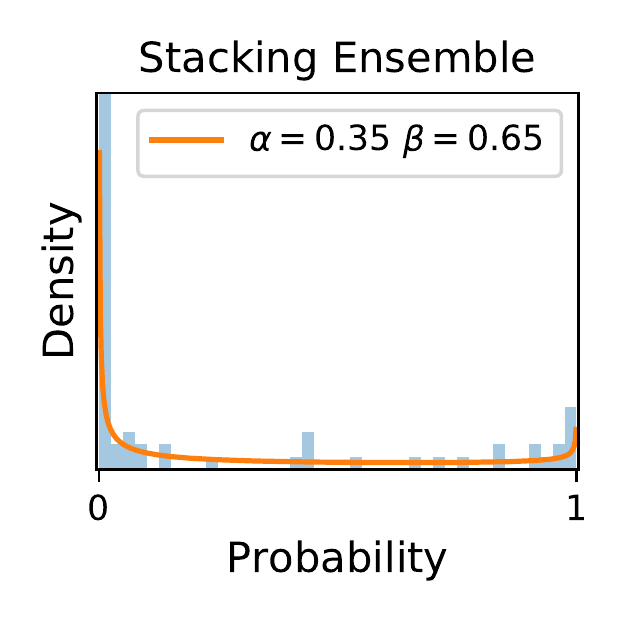}
    \includegraphics[trim=0.85cm 0.9cm 0cm 0.85cm,clip, height=2cm]{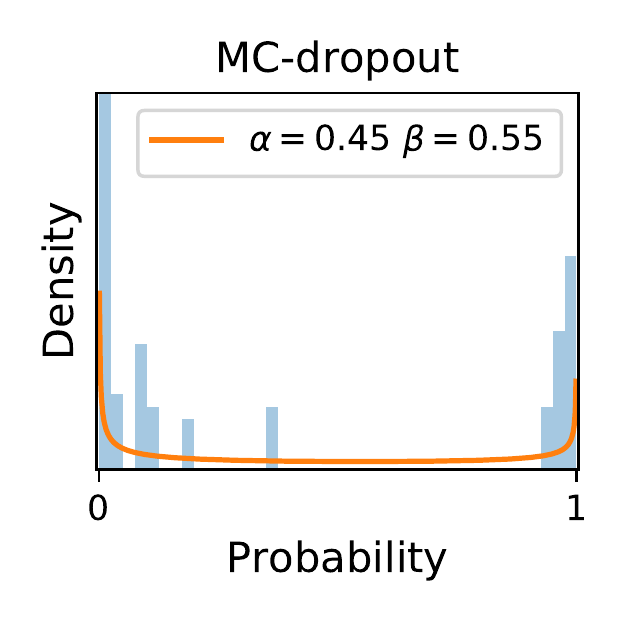}
    \includegraphics[trim=0.85cm 0cm 0cm 0.85cm,clip, height=2.45cm]{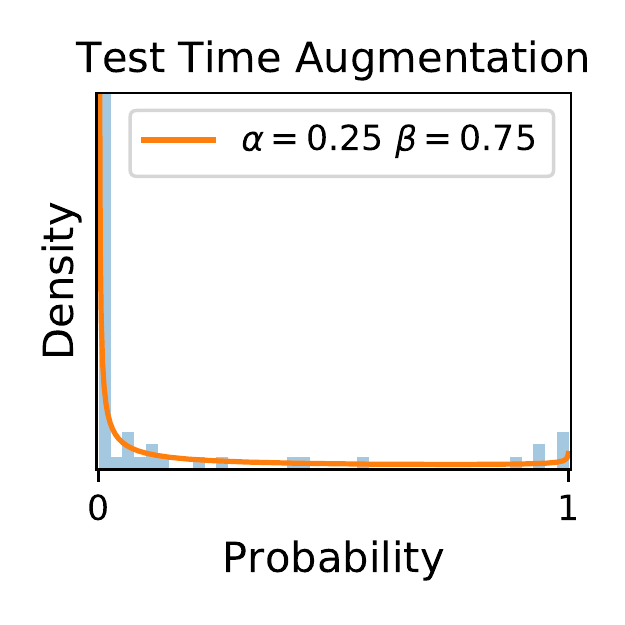}
    \caption{{SL1 (Mild)}}
  \end{subfigure}
  \begin{subfigure}[t]{0.19\linewidth}
    \centering
    \includegraphics[height=2cm]{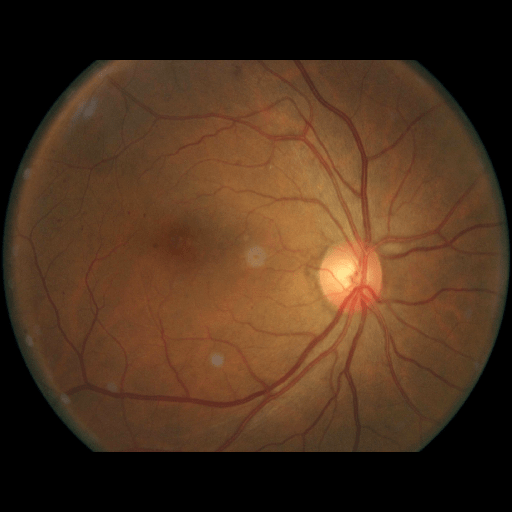}
    \includegraphics[trim=0.85cm 0.9cm 0cm 0.85cm,clip, height=2cm]{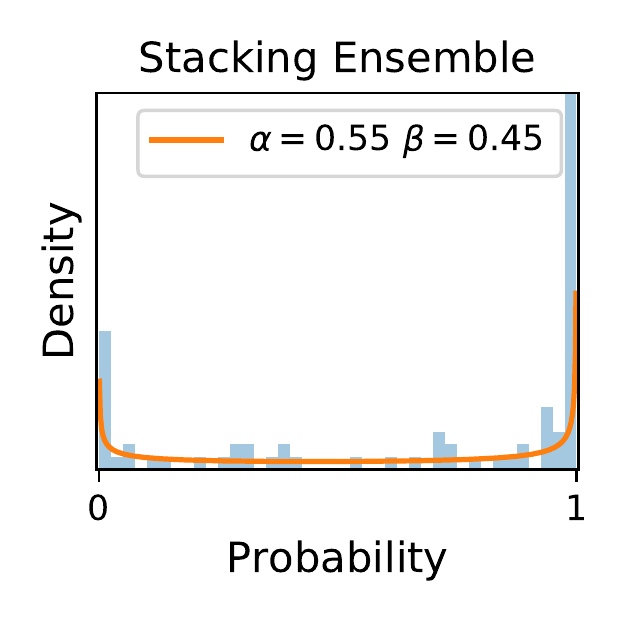}
    \includegraphics[trim=0.85cm 0.9cm 0cm 0.85cm,clip, height=2cm]{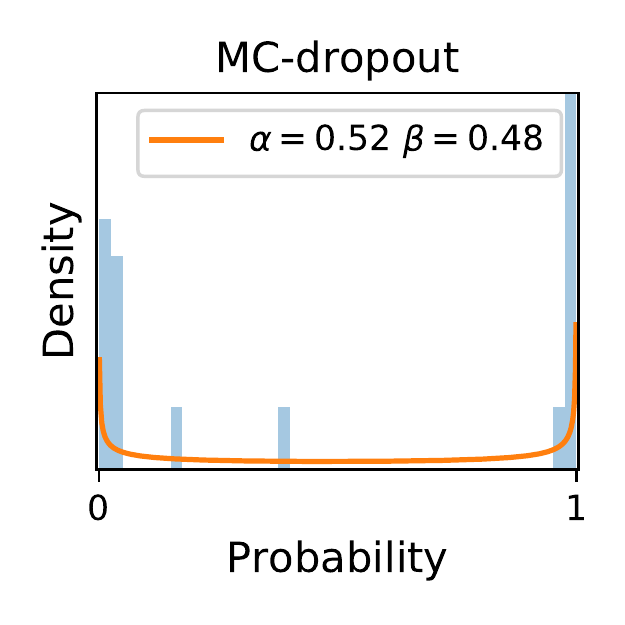}
    \includegraphics[trim=0.85cm 0cm 0cm 0.85cm,clip, height=2.45cm]{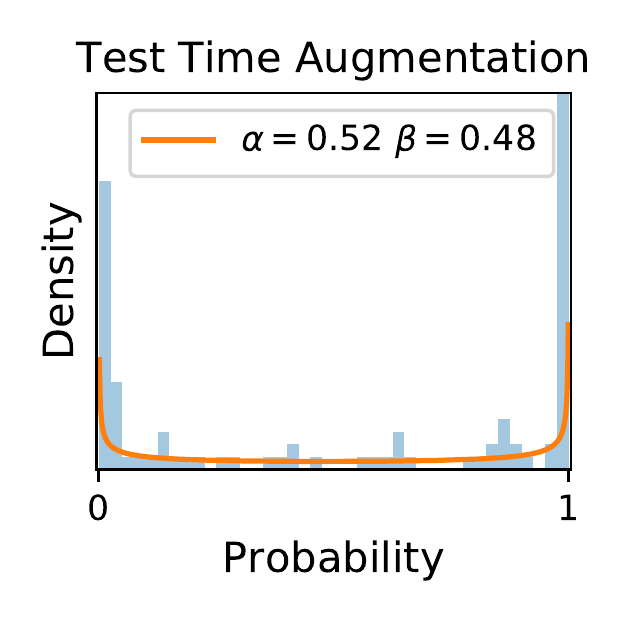}
    \caption{{SL2 (Moderate)}}
  \end{subfigure}
  \begin{subfigure}[t]{0.19\linewidth}
    \centering
    \includegraphics[height=2cm]{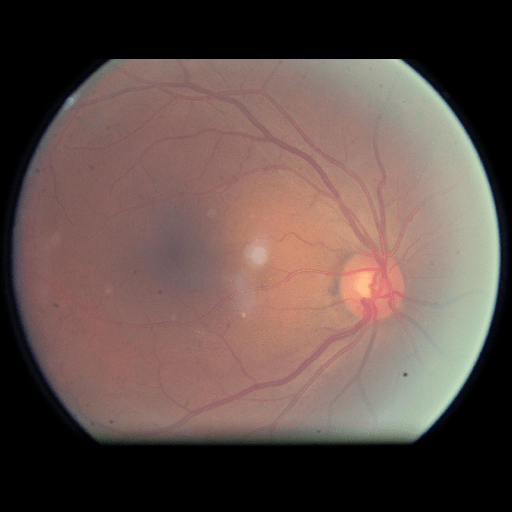}
    \includegraphics[trim=0.85cm 0.9cm 0cm 0.85cm,clip, height=2cm]{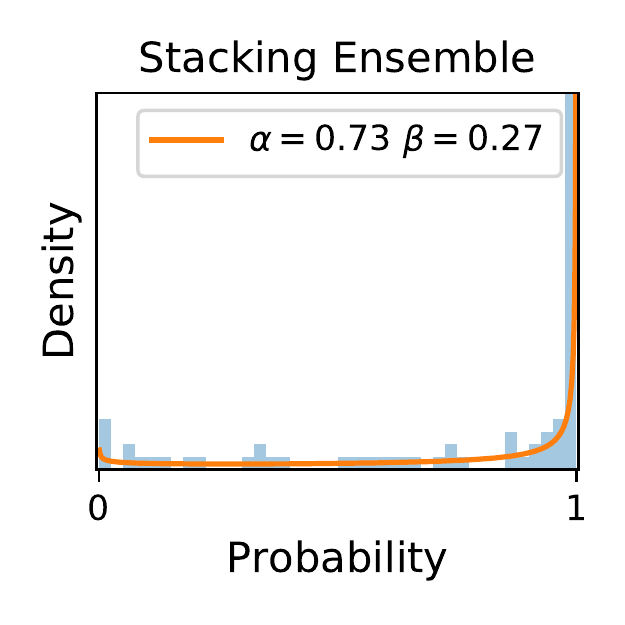}
    \includegraphics[trim=0.85cm 0.9cm 0cm 0.85cm,clip, height=2cm]{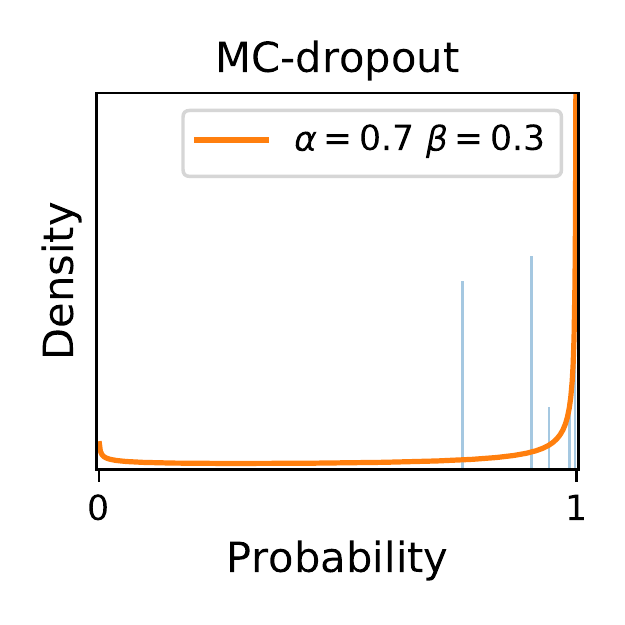}
    \includegraphics[trim=0.85cm 0cm 0cm 0.85cm,clip, height=2.45cm]{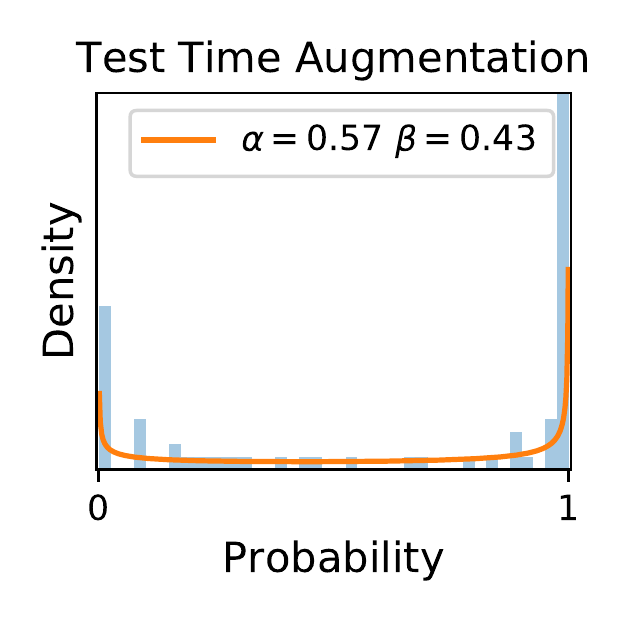}
    \caption{{SL3 (Severe)}}
  \end{subfigure}
  \begin{subfigure}[t]{0.19\linewidth}
    \centering
    \includegraphics[height=2cm]{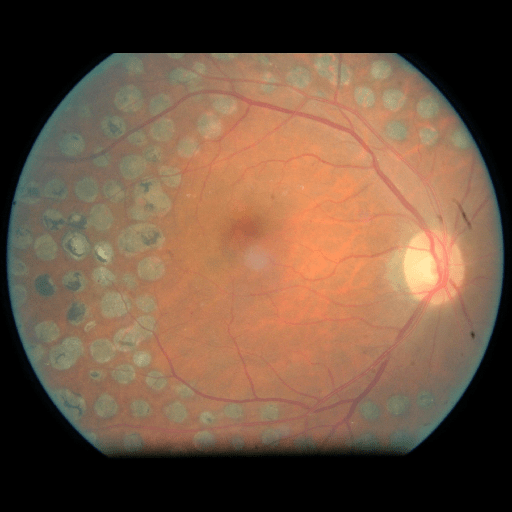}
    \includegraphics[trim=0.85cm 0.9cm 0cm 0.85cm,clip, height=2cm]{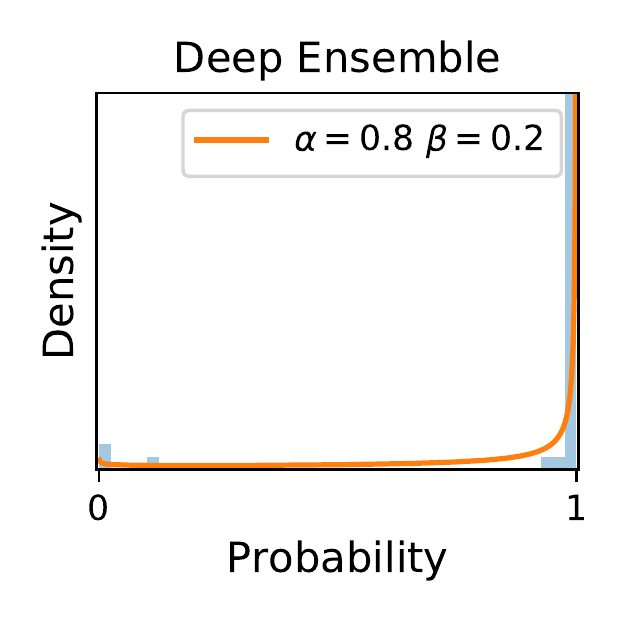}
    \includegraphics[trim=0.85cm 0.9cm 0cm 0.85cm,clip, height=2cm]{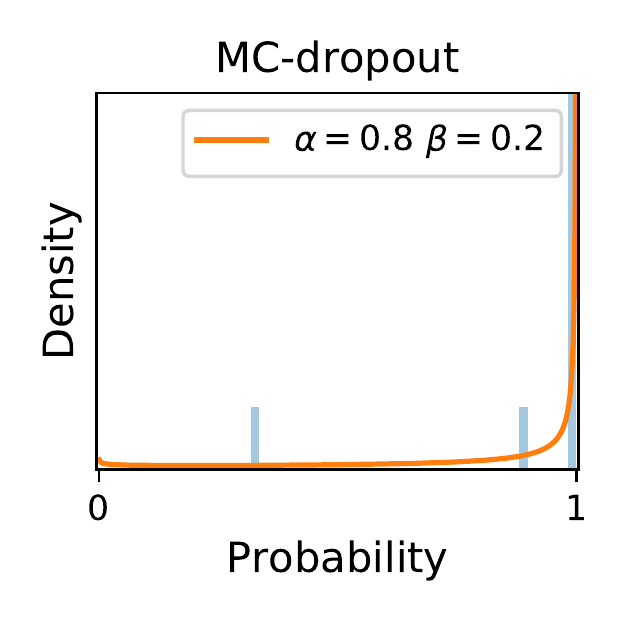}
    \includegraphics[trim=0.85cm 0cm 0cm 0.85cm,clip, height=2.45cm]{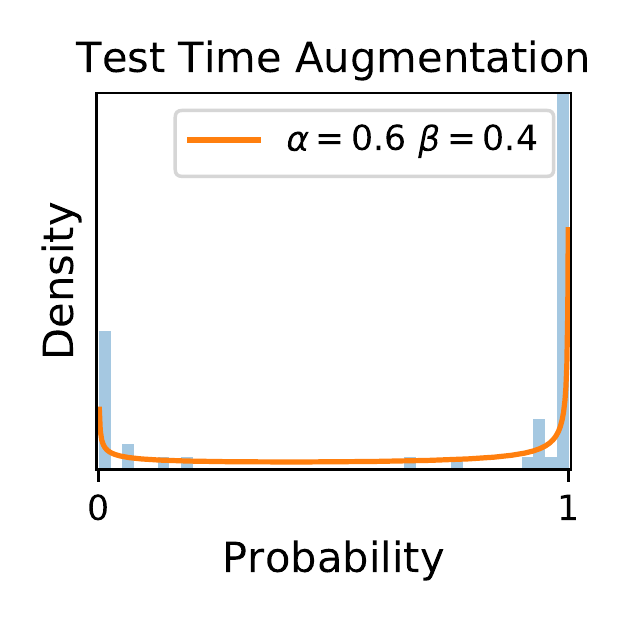}
    \caption{{SL4 (Proliferate)}}
  \end{subfigure}
\caption{Fundus images (top panel) of the five \acp{SL} of diabetic retinopathy disease and the distributions (shown as histograms) of their corresponding classifier predictions under stacking ensemble (second panel), MC-dropout (third panel) and \ac{TTA} (fourth panel). A beta distribution displayed by the orange curve is fitted to each distribution. Additional examples can be found in Figure~\ref{fig:beta-dist-five-SL-plus} in the supplementary materials.}
  \label{fig:beta-dist-five-SL}
\end{figure}

We conducted a case study on diagnosing diabetic retinopathy with ensembles of \ac{DL} models. 
For benchmarking the performance of our ensemble-based solutions under the scheme described in Sec.\ref{sec:Human-AI-scheme}, we used two popular collections of diabetic retinopathy image data, the Kaggle Diabetic Retinopathy dataset~\cite{cuadros2009eyepacs} (hereafter referred to as ``Kaggle-DR'') and the Messidor-2 dataset~\cite{decenciere_feedback_2014}, each respectively consisting of $88702$ and $1748$ high resolution images. Diabetic retinopathy is graded into five \acp{SL}, as displayed in Figure~\ref{fig:beta-dist-five-SL}. 
Following the problem setup used in previous papers~\cite{gulshan2016development}, we trained models to distinguish the referable (SL2-4) cases from the non-referable ones (SL0\,\&\,SL1) (see Section~\ref{sec:in-dist-datasets} for more detailed descriptions). We also tested our trained ensemble models on two \ac{o.o.d.} image datasets (ImageNet~\cite{deng2009imagenet} and CIFAR-10~\cite{cifar10dataset}) to examine their capabilities of identifying \ac{o.o.d.} inputs (see Section~\ref{sec:datasets} in the supplementary materials).

Three types of ensemble methods (stacking ensemble, MC-dropout and \ac{TTA}) described in Section~\ref{sec:ensemble-method} were evaluated in our experiments. We did not adopt the deep ensemble approach~\cite{lakshminarayanan2017simple} in our case study because we did not manage to train the networks from scratch (random initialization). Our solution was to train all the models by reusing the weights from a pretrained ImageNet model, which proved to work well but also prevented us from implementing the original deep ensemble approach~\cite{lakshminarayanan2017simple}.

Our stacking ensemble models consisted of networks of two architectures (resnet34~\cite{he2016deep} and VGG16~\cite{simonyan2014very}). To induce diversity, we trained the networks using different subset of images (as in bagging~\cite{breiman1996bagging}), images of different resolutions, training-time data augmentation strategies, and number of training epochs (as in snapshot ensembles~\cite{huang2017snapshot}). MC-dropout ensembles and \ac{TTA} ensembles are created during test time by repeatedly sampling the trained networks.

To obtain more statistically-convincing results, we created multiple instances for each model setting in our comparison study and report their average performance. Additional details of our experimental setup as well as our code are given in the supplementary materials. Next we analyze the findings from our experiment results.

\subsection{Distribution of Uncertainty Scores}\label{sec:uncertainty-scores}

\noindent\textbf{Distribution of Uncertainty Scores Across Different Severity Levels}
As explained in Section~\ref{sec:methodology}, each uncertainty metric essentially defines an order/ranking among the data points. We conducted an analysis to better understand what data will be assigned high uncertainty under a particular uncertainty metric $U$. Picking out the $\theta\%$ highest ranked data points ($\theta=1,2,5,10$), we calculated the ratio of data points from each \ac{SL}. Figure~\ref{fig:table2} summarizes the results as box plots for the Kaggle-DR and the Messidor-2 datasets; additional detailed statistics can be found in Table~\ref{tab:severity-ratio} in the supplementary materials. From the plot and table, SL1\,\&\,SL2 examples account for a higher proportion among the top-ranked uncertain examples across the three ensemble methods. This finding matches our intuition that incipient disease examples (SL1\,\&\,SL2) are more likely to be considered uncertain by ensemble methods due to their ambiguity.

\noindent\textbf{Uncertainty Scores on Out-of-Distribution Datasets}
As an additional experiment, we also tested the performance of the ensemble models on \ac{o.o.d.} data inputs, which is a classic application of prediction uncertainties~\cite{lakshminarayanan2017simple}. For this task, we produced distribution plots similar to those in Figure~\ref{fig:table2} for the previous experiment in our supplementary materials. The visualizations for the two \ac{o.o.d.} image datasets can be found in Figure~\ref{fig:histogram_imagenet} for ImageNet and in Figure~\ref{fig:histogram_cifar10} for CIFAR-10. The results showed that the majority of \ac{o.o.d.} data received higher uncertainty scores than in-distribution data for all three ensemble methods, suggesting that these ensemble methods would indeed perform well on \ac{o.o.d.} detection tasks.

\begin{figure}[t]
  \centering
  \includegraphics[width=0.99\textwidth]{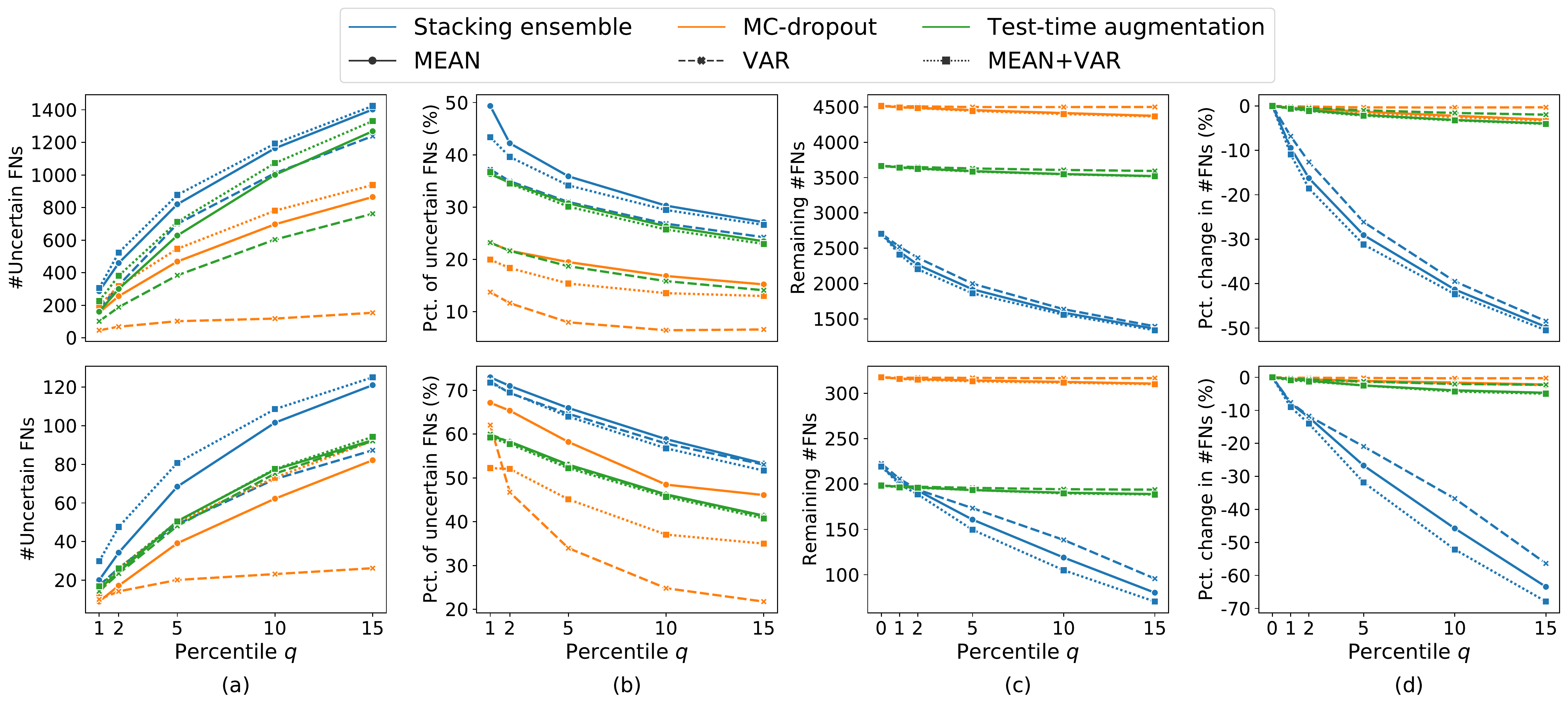}
  \caption{Experimental results on the Kaggle-DR dataset (top panel) and Messidor-2 dataset (bottom panel) after applying uncertainty-informed human-AI diagnostic schemes, with $\rho=0.2$. An ensemble size of 10 is used for the displayed stacking ensembles, and an ensemble size of 50 for MC-dropout and \ac{TTA} methods. We show in (a) the number of false negatives among the identified uncertain negatives and in (b) the \ac{FNP} under different ensemble methods and uncertainty metrics. The number of remaining false negative decisions after human efforts are involved is showed in (c), and the percentage reduction (change) in number of false negatives is displayed in (d). Additional detailed results can be found in Table~\ref{tab:fnp-kaggle} and Table~\ref{tab:fnp-messidor} in the supplementary materials.}
  \label{fig:experiment-results}
\end{figure}

\subsection{Identifying False Negative Decisions Using Prediction Uncertainties}\label{sec:fnp}

\noindent\textbf{False Negatives vs. Uncertain Negatives}
As discussed in Section~\ref{sec:formulation}, the \ac{FNP} measures the probability of the identified uncertain negatives being actual false negatives. The higher the \ac{FNP}, the more wrong predictions an uncertainty-informed scheme can potentially correct. In Figures~\ref{fig:experiment-results}b, we report the \ac{FNP} values under different $q$ settings from different combinations of ensemble methods and uncertainty metrics. The results in uncertain false negatives and their ratios to the number of uncertain negatives are illustrated in Figures~\ref{fig:experiment-results}a\,\&\,\ref{fig:experiment-results}b, respectively. \textsc{mean} gives higher false negative precision values than \textsc{var}, which validates Theorem~\ref{thm:mean-vs-var-beta-finite}.

\noindent\textbf{A Hypothetical Uncertainty Metric \textsc{mean}+\textsc{var}}
We can observe from the results in Figures~\ref{fig:experiment-results}\,\&\,\ref{fig:table2} that the performance indices given by \textsc{mean} and \textsc{var} are sometimes close. Hence,   a natural question to ask is how much overlap is there between the uncertain examples identified by the two uncertainty metrics. To answer this question, we created a hypothetical metric \textsc{mean+var} in our analysis in addition to \textsc{mean} and \textsc{var}. The uncertain examples identified by \textsc{mean+var} are the union of the two sets of uncertain examples identified by \textsc{mean} and by \textsc{var}, not subject to the constraint imposed by $q$. Therefore, it is at least as good as \textsc{mean} or \textsc{var}. If \textsc{mean} and \textsc{var} do not have much overlapping, \textsc{mean+var} will identify many more false negatives than either of them alone; however, we can see from Figure~\ref{fig:experiment-results} that this is not the case. The results given by \textsc{mean+var} do not have much improvement over those given by \textsc{mean}, indicating that many of the false negatives identified by \textsc{var} are also captured by \textsc{mean}, matching the expectation of Theorem~\ref{thm:mean-vs-var-beta-finite}.

\noindent\textbf{Reduction in the Number of False Negative Decisions}
Finally, we report the performance of the uncertainty-informed human-AI diagnostic scheme, and compare that to the baseline scheme in which decision uncertainties are not used to inform decision-making. 
We show in Figure~\ref{fig:experiment-results}c the number of remaining false negatives at different values of $q$, and in Figure~\ref{fig:experiment-results}d the percentage changes w.r.t. the baseline ($q=0$). We can see from the two plots that \textsc{mean} gives modest improvement compared to \textsc{var} with all three ensemble methods, as asserted by Theorem~\ref{thm:mean-vs-var-beta-finite}. When $q=5$, the stacking ensemble method achieved a $29.08\%$ decrease (1918 people) in the number of false negatives under \textsc{mean} and a $26.14\%$ decrease (1997 people) under \textsc{var}. We include more detailed results in Tables~\ref{tab:num-false-negatives-DR}\,\&\,\ref{tab:num-false-negatives-messidor} in the supplementary materials.

\vspace{-3mm}
\subsection{Comparing the Ensemble Methods}\label{sec:comparison-uncertainty}

\begin{figure}[t]
  \centering
  \includegraphics[width=0.99\textwidth]{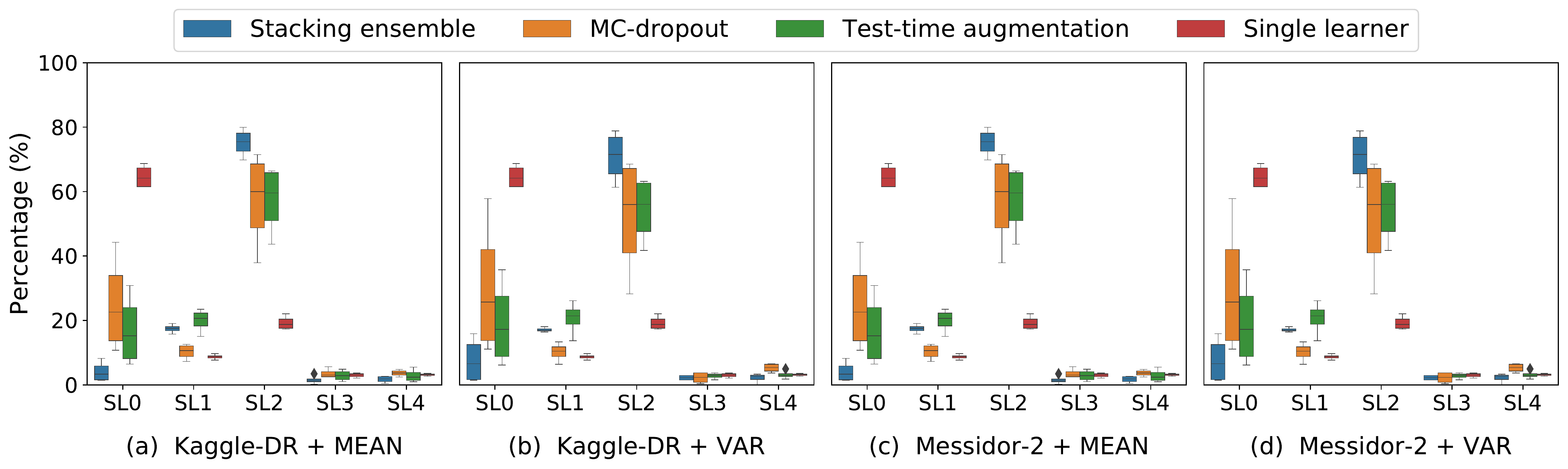}
  \caption{Distributions of uncertainty scores across different severity levels, where the variations represented by the boxes are due to varying $\theta$ values. A comparison of the three ensemble methods used in our case study and the single learner case is presented. The same set of ensemble models that we analyzed in Figure~\ref{fig:experiment-results} are used in this figure.  More detailed results are given in Table~\ref{tab:severity-ratio} in the supplementary materials.}
  \label{fig:table2}
\end{figure}

Comparing the three ensemble methods in Figure~\ref{fig:table2}, the stacking ensemble method has the highest ratios of SL1\,\&\,SL2 data among the high-uncertainty examples it identified under both \textsc{mean} and \textsc{var}. \ac{TTA} showed slightly better performance than MC-dropout but still falls behind the stacking ensemble method. Considering the fact that SL0 examples accounted for the majority of the dataset, the stacking ensemble method was much more precise (specific) in selecting truly ambiguous data points that were difficult to classify. From Figure~\ref{fig:experiment-results}, we can also see that the stacking ensemble method greatly outperformed the other two methods in finding false negatives under both \textsc{mean} and \textsc{var} uncertainty metrics.

In contrast, the MC-dropout method showed the worst overall performance among the three, as it can be seen from the high ratios of SL0 examples among the uncertain negatives in Figure~\ref{fig:table2}. The histograms in Figure~\ref{fig:beta-dist-five-SL} provides another perspective to look into the phenomenon, where a decent proportion of MC-dropout model's predictions on SL0 inputs entailed low confidence (far from 0 or 1), which from another angle explained why MC-dropout was less specific in terms of lower \ac{FNP}; many no-DR inputs (i.e. SL0) were erroneously assigned high uncertainty by MC-dropout models.

It is still an open question why the evaluated MC-dropout networks signaled relatively high uncertainty on SL0\,\&\,SL3\,\&\,SL4 data that are less likely to be ambiguous. We conjecture that much of the ``uncertainty'' indicated by disagreement among test-time dropout samples actually reflects the stochastic nature of dropout networks rather than the real decision uncertainty associated with the data. It is worth noting that the MC-dropout model we evaluated was not weak \textit{per se}; they all achieved above $0.98$ \ac{AUC} scores on test sets. The weakness of individual test-time samples (which explains their low-confidence predictions on SL0\,\&\,SL3\,\&\,SL4) might have been hidden when they are aggregated into an ensemble---a well-known advantage of ensemble learning. Our results suggested that the uncertainty information given by implicit ensemble methods such as MC-dropout and \ac{TTA} might not be as reliable as that from explicit ensemble approaches (e.g., stacking ensembles). Similar findings on MC-dropout can  be found in some previous papers~\cite{lakshminarayanan2017simple}.

\vspace{-3mm}
\section{Discussion and Conclusion}\label{sec:conclusion}

Ensemble learning is a powerful and widely-applied method for improving decision-making models. In addition, the uncertainty information from ensemble learners can be exploited to indicate \ac{o.o.d.} inputs or decisions that are potentially wrong. Despite a number of recent papers on this topic, an in-depth comparison and analysis on these methods is still lacking. In this paper, we studied the choices of ensemble methods and uncertainty metrics when designing an ensemble-based diagnostic system. Through theoretical analysis and extensive experimentation, we answered the question why the \textsc{mean} metric is preferable to the \textsc{var} metric. Our results also showed that the stacking ensemble method outperforms MC-dropout and \ac{TTA}. Our future work includes extending the current investigation into multiclass classification scenarios and designing more effective type-2 metrics that combine the advantages of \textsc{mean} and \textsc{var}. 



\clearpage
\section*{Broader Impact}

\acf{DL} have shown great promise in medical diagnosis applications due to their ability to classify data accurately~\cite{de2018clinically,cheng2016computer,leibig2017leveraging}; however, the risks of misclassification made by AI can never be overlooked, as mistakes are often costly in healthcare. Our research aims to tackle the mistakes made by AI, and to seek ways to mitigate the consequences.

\paragraph{Impact on Related \ac{ML} Fields}
Our work contributes to the theoretical discussion of several important sub-fields of \ac{ML} and AI, including active learning~\cite{chen14active,javdani14near}, anomaly detection\cite{jin2019detecting, jin2019encoder}, and ensemble learning~\cite{zhou2012ensemble}. Our discussed research opens up a new path towards gaining deeper understanding of the interactions between ensemble learning and uncertainty estimation. Prior works in this area are mostly empirical. Although appealing improvement has been demonstrated by various ensemble methods and on a diverse set of tasks, gaining a systemic understanding is very important which can guide us in designing more effective algorithms and avoid unnecessary trial-and-errors. 

The applications of the proposed ensemble methodology can go beyond healthcare AI. For example, the \ac{FDD} of industrial machines~\cite{li2016fault,jin2019detecting,jin2019one} share a very similar problem structure as disease diagnosis problems, and as a result the proposed methodology can also be applied to detecting incipient faults~\cite{jin2019detecting}, another important and challenging topic.

\paragraph{Impact on Healthcare AI}
One direct impact of our researched method on healthcare AI is the accuracy improvement. By selecting the uncertain negative examples to send for human diagnosticians, our method helps reduce the number false negatives. On top of that, the uncertainty-informed diagnostic scheme is critical to the adoption of AI-based diagnostic schemes, since it can offer an assessment of decision risks in addition to high accuracy, which is important both to healthcare providers and to patients.

\paragraph{Potential Issues and Opportunities for Further Research}
Apart from the aforementioned contributions of our research, we would also like to emphasize the potential issues that could arise from our research. As already mentioned in our paper, the uncertainty estimation process can be thought of as another classifier that sits on top of the main classifier (the one that tells diseased cases from healthy ones), which can be used to predict 1) whether the predictions of an ensemble model are correct or wrong, or 2) whether or not an input is \ac{o.o.d.}. Uncertainty estimation, like any other classifier, can also have bias in 
how they make decisions (in our task, decide which examples are more uncertain). It is still unknown whether the uncertainty estimation techniques helps mitigate or strengthen the biases of the main classifier. As of now, these questions are largely unanswered, and therefore we need further research and investigations to better understand them, which will constitute an essential part of our future research.

\clearpage
\bibliographystyle{ieeetr}
\bibliography{refs}

\begin{thebibliography}{10}

\bibitem{lakshminarayanan2017simple}
B.~Lakshminarayanan, A.~Pritzel, and C.~Blundell, ``Simple and scalable
  predictive uncertainty estimation using deep ensembles,'' in {\em Advances in
  neural information processing systems}, pp.~6402--6413, 2017.

\bibitem{gal2016dropout}
Y.~Gal and Z.~Ghahramani, ``Dropout as a {B}ayesian approximation: Representing
  model uncertainty in deep learning,'' in {\em international conference on
  machine learning}, pp.~1050--1059, 2016.

\bibitem{jin2019augmenting}
B.~Jin, Y.~Tan, Y.~Chen, and A.~Sangiovanni-Vincentelli, ``Augmenting monte
  carlo dropout classification models with unsupervised learning tasks for
  detecting and diagnosing out-of-distribution faults,'' {\em arXiv preprint
  arXiv:1909.04202}, 2019.

\bibitem{ayhan2018test}
M.~S. Ayhan and P.~Berens, ``Test-time data augmentation for estimation of
  heteroscedastic aleatoric uncertainty in deep neural networks,'' 2018.

\bibitem{wang2018automatic}
G.~Wang, W.~Li, S.~Ourselin, and T.~Vercauteren, ``Automatic brain tumor
  segmentation using convolutional neural networks with test-time
  augmentation,'' in {\em International MICCAI Brainlesion Workshop},
  pp.~61--72, Springer, 2018.

\bibitem{jin2020ensemble}
B.~Jin, Y.~Tan, Y.~Chen, and V.~A.~S. Poolla, Kameshwar, ``Are ensemble
  classifiers powerful enough for the detection and diagnosis of
  intermediate-severity faults?,'' {\em Proceedings of the 25th ACM SIGKDD
  International Conference on Knowledge Discovery \& Data Mining (submitted)},
  2020.

\bibitem{blennow2003csf}
K.~Blennow and H.~Hampel, ``Csf markers for incipient alzheimer's disease,''
  {\em The Lancet Neurology}, vol.~2, no.~10, pp.~605--613, 2003.

\bibitem{fort2019deep}
S.~Fort, H.~Hu, and B.~Lakshminarayanan, ``Deep ensembles: A loss landscape
  perspective,'' {\em arXiv preprint arXiv:1912.02757}, 2019.

\bibitem{wolpert1992stacked}
D.~H. Wolpert, ``Stacked generalization,'' {\em Neural networks}, vol.~5,
  no.~2, pp.~241--259, 1992.

\bibitem{sikora2015modified}
R.~Sikora {\em et~al.}, ``A modified stacking ensemble machine learning
  algorithm using genetic algorithms,'' in {\em Handbook of Research on
  Organizational Transformations through Big Data Analytics}, pp.~43--53, IGi
  Global, 2015.

\bibitem{huang2017snapshot}
G.~Huang, Y.~Li, G.~Pleiss, Z.~Liu, J.~E. Hopcroft, and K.~Q. Weinberger,
  ``Snapshot ensembles: Train 1, get {M} for free,'' {\em arXiv preprint
  arXiv:1704.00109}, 2017.

\bibitem{srivastava2014dropout}
N.~Srivastava, G.~Hinton, A.~Krizhevsky, I.~Sutskever, and R.~Salakhutdinov,
  ``Dropout: a simple way to prevent neural networks from overfitting,'' {\em
  The Journal of Machine Learning Research}, vol.~15, no.~1, pp.~1929--1958,
  2014.

\bibitem{gal2016uncertainty}
Y.~Gal, ``Uncertainty in deep learning,'' {\em University of Cambridge}, 2016.

\bibitem{zhou2012ensemble}
Z.-H. Zhou, {\em Ensemble methods: foundations and algorithms}.
\newblock Chapman and Hall/CRC, 2012.

\bibitem{zoph2016neural}
B.~Zoph and Q.~V. Le, ``Neural architecture search with reinforcement
  learning,'' {\em arXiv preprint arXiv:1611.01578}, 2016.

\bibitem{bardenet2013collaborative}
R.~Bardenet, M.~Brendel, B.~K{\'e}gl, and M.~Sebag, ``Collaborative
  hyperparameter tuning,'' in {\em International conference on machine
  learning}, pp.~199--207, 2013.

\bibitem{wong2016understanding}
S.~C. Wong, A.~Gatt, V.~Stamatescu, and M.~D. McDonnell, ``Understanding data
  augmentation for classification: when to warp?,'' in {\em 2016 international
  conference on digital image computing: techniques and applications (DICTA)},
  pp.~1--6, IEEE, 2016.

\bibitem{rodriguez2009sensitivity}
J.~D. Rodriguez, A.~Perez, and J.~A. Lozano, ``Sensitivity analysis of k-fold
  cross validation in prediction error estimation,'' {\em IEEE transactions on
  pattern analysis and machine intelligence}, vol.~32, no.~3, pp.~569--575,
  2009.

\bibitem{leibig2017leveraging}
C.~Leibig, V.~Allken, M.~S. Ayhan, P.~Berens, and S.~Wahl, ``Leveraging
  uncertainty information from deep neural networks for disease detection,''
  {\em Scientific reports}, vol.~7, no.~1, p.~17816, 2017.

\bibitem{jin2019detecting}
B.~Jin, D.~Li, S.~Srinivasan, S.-K. Ng, K.~Poolla, and
  A.~Sangiovanni-Vincentelli, ``Detecting and diagnosing incipient building
  faults using uncertainty information from deep neural networks,'' in {\em
  2019 IEEE International Conference on Prognostics and Health Management
  (ICPHM)}, pp.~1--8, IEEE, 2019.

\bibitem{goldberger2003efficient}
J.~Goldberger, S.~Gordon, and H.~Greenspan, ``An efficient image similarity
  measure based on approximations of kl-divergence between two gaussian
  mixtures,'' in {\em null}, p.~487, IEEE, 2003.

\bibitem{cuadros2009eyepacs}
J.~Cuadros and G.~Bresnick, ``Eyepacs: an adaptable telemedicine system for
  diabetic retinopathy screening,'' {\em Journal of diabetes science and
  technology}, vol.~3, no.~3, pp.~509--516, 2009.

\bibitem{decenciere_feedback_2014}
E.~Decencière, X.~Zhang, G.~Cazuguel, B.~Lay, B.~Cochener, C.~Trone, P.~Gain,
  R.~Ordonez, P.~Massin, A.~Erginay, B.~Charton, and J.-C. Klein, ``Feedback on
  a publicly distributed database: the messidor database,'' {\em Image Analysis
  \& Stereology}, vol.~33, pp.~231--234, Aug. 2014.

\bibitem{gulshan2016development}
V.~Gulshan, L.~Peng, M.~Coram, M.~C. Stumpe, D.~Wu, A.~Narayanaswamy,
  S.~Venugopalan, K.~Widner, T.~Madams, J.~Cuadros, {\em et~al.}, ``Development
  and validation of a deep learning algorithm for detection of diabetic
  retinopathy in retinal fundus photographs,'' {\em Jama}, vol.~316, no.~22,
  pp.~2402--2410, 2016.

\bibitem{deng2009imagenet}
J.~Deng, W.~Dong, R.~Socher, L.-J. Li, K.~Li, and L.~Fei-Fei, ``Imagenet: A
  large-scale hierarchical image database,'' in {\em 2009 IEEE conference on
  computer vision and pattern recognition}, pp.~248--255, IEEE, 2009.

\bibitem{cifar10dataset}
A.~Krizhevsky, ``Learning multiple layers of features from tiny images,'' {\em
  University of Toronto}, 05 2012.

\bibitem{he2016deep}
K.~He, X.~Zhang, S.~Ren, and J.~Sun, ``Deep residual learning for image
  recognition,'' in {\em Proceedings of the IEEE conference on computer vision
  and pattern recognition}, pp.~770--778, 2016.

\bibitem{simonyan2014very}
K.~Simonyan and A.~Zisserman, ``Very deep convolutional networks for
  large-scale image recognition,'' {\em arXiv preprint arXiv:1409.1556}, 2014.

\bibitem{breiman1996bagging}
L.~Breiman, ``Bagging predictors,'' {\em Machine learning}, vol.~24, no.~2,
  pp.~123--140, 1996.

\bibitem{de2018clinically}
J.~De~Fauw, J.~R. Ledsam, B.~Romera-Paredes, S.~Nikolov, N.~Tomasev,
  S.~Blackwell, H.~Askham, X.~Glorot, B.~O’Donoghue, D.~Visentin, {\em
  et~al.}, ``Clinically applicable deep learning for diagnosis and referral in
  retinal disease,'' {\em Nature medicine}, vol.~24, no.~9, pp.~1342--1350,
  2018.

\bibitem{cheng2016computer}
J.-Z. Cheng, D.~Ni, Y.-H. Chou, J.~Qin, C.-M. Tiu, Y.-C. Chang, C.-S. Huang,
  D.~Shen, and C.-M. Chen, ``Computer-aided diagnosis with deep learning
  architecture: applications to breast lesions in us images and pulmonary
  nodules in ct scans,'' {\em Scientific reports}, vol.~6, no.~1, pp.~1--13,
  2016.

\bibitem{chen14active}
Y.~Chen, H.~Shioi, C.~F. Montesinos, L.~P. Koh, S.~Wich, and A.~Krause,
  ``Active detection via adaptive submodularity,'' in {\em Proc. International
  Conference on Machine Learning (ICML)}, June 2014.

\bibitem{javdani14near}
S.~Javdani, Y.~Chen, A.~Karbasi, A.~Krause, J.~A. Bagnell, and S.~Srinivasa,
  ``Near-optimal bayesian active learning for decision making,'' in {\em In
  Proc. International Conference on Artificial Intelligence and Statistics
  (AISTATS)}, April 2014.

\bibitem{jin2019encoder}
B.~Jin, Y.~Tan, A.~Nettekoven, Y.~Chen, U.~Topcu, Y.~Yue, and A.~S.
  Vincentelli, ``An encoder-decoder based approach for anomaly detection with
  application in additive manufacturing,'' {\em arXiv preprint
  arXiv:1907.11778}, 2019.

\bibitem{li2016fault}
D.~Li, Y.~Zhou, G.~Hu, and C.~J. Spanos, ``Fault detection and diagnosis for
  building cooling system with a tree-structured learning method,'' {\em Energy
  and Buildings}, vol.~127, pp.~540--551, 2016.

\bibitem{jin2019one}
B.~Jin, Y.~Chen, D.~Li, K.~Poolla, and A.~Sangiovanni-Vincentelli, ``A
  one-class support vector machine calibration method for time series change
  point detection,'' {\em arXiv preprint arXiv:1902.06361}, 2019.

\bibitem{cho2008variance}
E.~Cho and M.~J. Cho, ``Variance of sample variance,'' {\em Section on Survey
  Research Methods--JSM}, vol.~2, pp.~1291--1293, 2008.

\bibitem{google_2018}
Kaggle, ``Messidor-2 {DR} grades,'' Jul 2018.

\bibitem{wang2019aleatoric}
G.~Wang, W.~Li, M.~Aertsen, J.~Deprest, S.~Ourselin, and T.~Vercauteren,
  ``Aleatoric uncertainty estimation with test-time augmentation for medical
  image segmentation with convolutional neural networks,'' {\em
  Neurocomputing}, vol.~338, pp.~34--45, 2019.

\end{thebibliography}

\appendix
\renewcommand\thefigure{S.\arabic{figure}}
\renewcommand\thetable{S.\arabic{table}}
\clearpage

\section*{\Large Supplementary Material}
\section{Proofs of Theoretical Results}

\subsection*{Errata}
There were three typos in the submitted version of the main paper for Lemma~\ref{lm:sample-uncertainty} and Theorem~\ref{thm:mean-vs-var-beta-finite}: the directions of inequality signs in the statements were mistakenly flipped.  To avoid confusion to the reviewers, we give the correct versions below; the changes have been highlighted in red. The narrative around our theoretical results in the main paper remains correct and unchanged.

\setcounter{lemma}{0}
\setcounter{theorem}{0}
\begin{lemma}
Consider two inputs $x_i, x_j$ with uncertainty score $s(x_i)$ and $s(x_j)$ estimated from $n$ \emph{i.i.d.} ensemble learners, and denote by $\Delta_{ij}(s) := \mathbb{E}[s(x_j)-s(x_i)]$ the difference of expected uncertainty score.
If $\Delta_{ij}(s) > 0$, then $P\left(\red{s(x_i)>s(x_j)}\right) = \mathcal{O}\left(\frac{\text{Var}(s(x_i)+\text{Var}(s(x_j))}{\Delta_{ij}^2(s)}\right)$.
\end{lemma}

\begin{theorem}
Consider inputs $x_i$, $x_j$, with $y_i \sim \mathcal{B}(\alpha_i,\beta_i)$, $y_j \sim \mathcal{B}(\alpha_j,\beta_j)$, and $\alpha_i+\beta_i = \alpha_j+\beta_j$. Let $\Delta_{ij}(s) := \mathbb{E}[s(x_j)-s(x_i)]$ where $s(\cdot)$ denotes an uncertainty score estimated from $n$ \emph{i.i.d.} ensemble learners. If $\alpha_i < \alpha_j \leq \beta_j$, then $\Delta_{ij}(s_\textsc{mean}) > \Delta_{ij}(s_\textsc{var}) > 0$. Furthermore, it holds that $P\left(\red{s_\textsc{mean}(x_i) > s_\textsc{mean}(x_j)}\right) = \mathcal{O}\left(\frac{1}{n \Delta_{ij}^2(s_\textsc{mean})}\right)$ and $P\left(\red{s_\textsc{var}(x_i) > s_\textsc{var}(x_j)}\right) = \mathcal{O}\left(\frac{1}{n \Delta_{ij}^2(s_\textsc{var})}\right)$.  
\end{theorem}

\subsection{Proof of Lemma~\ref{thm:mean-vs-var-beta-finite}}\label{sec:lemma}
\begin{proof}
Let $\rvdiff=s(x_j)-s(x_i)$ be a random variable, where $s(x_i)$ and $s(x_j)$ denote the uncertainty score of $x_i$ and $x_j$ estimated from $n$ i.i.d. emsemble learners. Clearly $\Delta_{ij}(s) = \expct{\rvdiff} > 0$.

By Chebyshev's inequality, we obtain
\begin{align*}
P\left( \abs{\rvdiff - \expct{\rvdiff}} \geq \Delta_{ij}(s) \right) \leq \frac{\text{Var}(\rvdiff)}{\Delta_{ij}^2(s)}
\end{align*}
which implies that
\begin{align*}
P\left( \rvdiff - \expct{\rvdiff} \leq -\Delta_{ij}(s) \right) &= 
P\left( \rvdiff - \Delta_{ij}(s) \leq -\Delta_{ij}(s) \right) \\
&= P\left( s(x_j)-s(x_i) \leq 0 \right) \\
&= P\left( s(x_i) \geq s(x_j) \right) \\
&\leq \frac{\text{Var}(\rvdiff)}{\Delta_{ij}^2(s)} 
\end{align*}
Further note that $\text{Var}(\rvdiff) = \text{Var}(s(x_j)-s(x_i)) = \text{Var}(s(x_j)) + \text{Var}(s(x_i))$; we hence conclude that
\begin{align*}
P\left(s(x_i) > s(x_j)\right) = \mathcal{O}\left(\frac{\text{Var}(s(x_i)+\text{Var}(s(x_j))}{\Delta_{ij}^2(s)}\right)  
\end{align*}
which completes the proof.
\end{proof}

\subsection{Proof of Theorem~\ref{thm:mean-vs-var-beta-finite}}\label{sec:mean-vs-var-beta-finite}
\begin{proof}
To prove the first statement, i.e. $\Delta_{ij}(s_\textsc{mean}) > \Delta_{ij}(s_\textsc{var}) > 0$, we consider the following properties of the beta distribution:
\begin{align*}
\mu(x_i) = \frac{\alpha_i}{\alpha_i + \beta_i}, \qquad
\sigma(x_i) = \frac{\alpha_i \beta_i}{(\alpha_i + \beta_i)^2 (1+\alpha_i+\beta_i)} = \frac{\mu(x_i)(1-\mu(x_i))}{1+\alpha_i+\beta_i}
\end{align*}
where $\mu(x_i)$ and $\sigma(x_i)$ respectively represent the mean and variance of the beta distribution $\mathcal{B}(\alpha_i, \beta_i)$.

Let $\alpha_i + \beta_i = \alpha_j + \beta_j = c$. Since $\alpha_i < \alpha_j \leq \beta_j$, we know 
\begin{align*}
    \mu(x_i) &= \frac{\alpha_i}{\alpha_i + \beta_i} < \frac{\alpha_j}{\alpha_j + \beta_j} = \mu(x_j) \leq \frac{1}{2}\\
    \sigma(x_i) &= \frac{\mu(x_i)(1-\mu(x_i))}{1+\alpha_i+\beta_i} < \frac{\mu(x_j)(1-\mu(x_j))}{1+\alpha_j+\beta_j} = \sigma(x_j)
\end{align*}
Therefore, $\Delta_{ij}(s_\textsc{mean}) = \expct{s_\textsc{mean}(x_j) - s_\textsc{mean}(x_i)} = \mu(x_j) - \mu(x_i)> 0$, and $\Delta_{ij}(s_\textsc{var}) = \expct{s_\text{Var}(x_j) - s_\text{Var}(x_i)} = \sigma(x_j) - \sigma(x_i) > 0$.
Furthermore, notice that
\begin{align*}
    \Delta_{ij}(s_\textsc{var}) 
    &=  \frac{\mu(x_j)(1-\mu(x_j)) - \mu(x_i)(1-\mu(x_i))}{1+c} \\
    &< \mu(x_j)(1-\mu(x_j)) - \mu(x_i)(1-\mu(x_i)) \\
    &= \mu(x_j)-\mu(x_i) - (\mu^2(x_j) - \mu^2(x_i)) \\
    &< \Delta_{ij}(s_\textsc{mean}),
\end{align*}
which proves the first statement.

To prove the second half of Theorem~\ref{thm:mean-vs-var-beta-finite}, i.e., to show the upper bounds on the error of $s_\textsc{mean}$ and $s_\textsc{var}$, we plug in the definition of $s_\textsc{mean}$ and $s_\textsc{var}$ to Lemma~\ref{thm:mean-vs-var-beta-finite}:
\begin{align*}
    P\left(s_\textsc{mean}(x_i) > s_\textsc{mean}(x_j)\right) &=
    \mathcal{O}\left(\frac{\text{Var}(s_\textsc{mean}(x_i)+\text{Var}(s_\textsc{mean}(x_j))}{\Delta_{ij}^2(s_\textsc{mean})}\right)\\
    &\stackrel{(a)}{=}\mathcal{O}\left(\frac{\sigma(x_j)+\sigma(x_i)}{n \Delta_{ij}^2(s_\textsc{mean})}\right) \\
    &=\mathcal{O}\left(\frac{1}{n \Delta_{ij}^2(s_\textsc{mean})}\right)
\end{align*}
where step (a) is due to $\text{Var}(s_\textsc{mean}(x_i)) = \sigma_i/n$. Similarly,
\begin{align*}
    P\left(s_\text{Var}(x_i) > s_\text{Var}(x_j)\right) &=
    \mathcal{O}\left(\frac{\text{Var}(s_\text{var}(x_i)+\text{Var}(s_\text{var}(x_j))}{\Delta_{ij}^2(s_\textsc{var})}\right)\\
    &\stackrel{(b)}{=}\mathcal{O}\left(\frac{1}{n \Delta_{ij}^2(s_\textsc{var})}\right)
\end{align*}
Here, step (b) is due to the variance of sample variance $\text{Var}(s_\text{Var}(x_i)) = \frac{1}{n} (\mu_4 - \sigma^2(x_i)) + \mathcal{O}(n^{-2}) = \mathcal{O}\left(\frac1n\right)$ where $\mu_4$ is the Kurtosis of the beta distribution $\mathcal{B}(\alpha_i,\beta_i)$~\cite{cho2008variance}.
\end{proof}

\section{Datasets Used in Our Study}\label{sec:datasets}

\subsection{In-distribution Datasets: Kaggle-DR and Messidor-2}\label{sec:in-dist-datasets}

The Kaggle-DR dataset comprises $88,702$ high resolution images. The presence of diabetic retinopathy is rated into five different \acp{SL}: no-DR ({SL0}), mild ({SL1}), moderate ({SL2}), severe ({SL3}), and proliferate ({SL4}), as illustrated in Figure~\ref{fig:beta-dist-five-SL}. We divided the Kaggle-DR dataset into a development set and a test set, which respectively consisted of $58,125$ and $30,577$ images. The data in the development set were used to train and validate our \acf{DL} models. The Messidor-2 dataset~\cite{google_2018} that consisted of $1,748$ images was also used in our experiment as an additional dataset to test the true generalization performance of the models trained on the Kaggle-DR dataset. Images in Messidor-2 dataset were graded into the five \acp{SL} as in the Kaggle-DR dataset. Figure~\ref{fig:dataset} provides an illustration of the datasets used in our experiments.

\subsubsection{Development Set of Kaggle-DR}\label{sec:development-kaggle-dr}

\paragraph{Ratio of Incipient Disease Data $\rho$}
SL1 and SL2 are the two incipient disease conditions with low \ac{SL}. Due to their close resemblance, the SL1\,\&\,SL2 are presumably more difficult to differentiate in our binary classification setting that attempts to identify the referable cases (SL2 and above) from the non-referable ones (SL0\,\&\,SL1). 

\setcounter{figure}{0}
\setcounter{table}{0}
\begin{figure}[b]
    \centering
    \includegraphics[width=\textwidth]{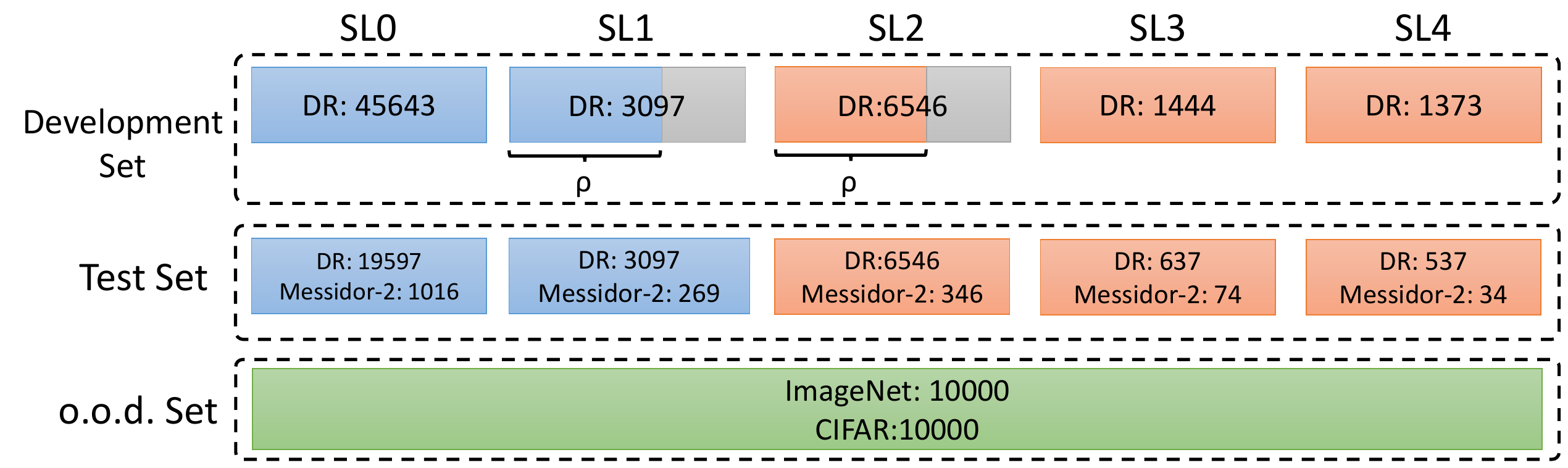}
    \caption{An illustration of the datasets used in our experimental study.}
    \label{fig:dataset}
\end{figure}

Since the incipient disease data are closer to the decision boundary than data of other \acp{SL}, it can be expected that the amount of available (accurately labeled) incipient disease data in the training set will greatly influence the final performance of the learned classifiers. To capture this effect, we conducted a set of experiments; in each experiment, only a proportion $\rho\in\{0.2, 0.4, 0.6, 0.8, 1.0\}$ of the SL1\,\&\,SL2 data in the development set were retained, and the rest were left out. Note that the same set of images were retained for a given $\rho$ value to avoid data leakage when constructing the ensemble models; in other words, each single learner in an ensemble was exposed to the same set of training images during training.

\noindent\textbf{Induce Diversity Using Bagging}
Each base learner in our training set was trained using $80\%$ of the retained development set data (drawn by random sampling with replacement), as in the bagging strategy~\cite{breiman1996bagging}. We notice that some previous works~\cite{lakshminarayanan2017simple} chose not to use the bagging strategy in training \ac{DL} models because bagging in those cases does not give improvement and can sometimes hurt performance. While it requires future work to understand better how the bagging strategy gets along with \ac{DL} models especially in medical image domains, we remark that it is still necessary to hold out a fraction of the development set data as a validation set for training each base learner.

\subsection{Out-of-distribution Test Datasets: ImageNet \& CIFAR-10}\label{sec:ood-datasets}
For comparing the performance of the evaluated ensemble methods in detecting \ac{o.o.d.} data, we used ImageNet\cite{deng2009imagenet} and CIFAR-10\cite{cifar10dataset} as additional test sets in our experiments. Due to the large sizes of these datasets, we used subsampled versions of the two image datasets by randomly selecting $10000$ images from each dataset.

\section{Experimental Details}\label{sec:experiment-details}

\subsection{Data Preprocessing}
\begin{figure}[htb]
  \centering
  \begin{subfigure}[t]{0.90\linewidth}
    \centering
    \includegraphics[height=3.5cm]{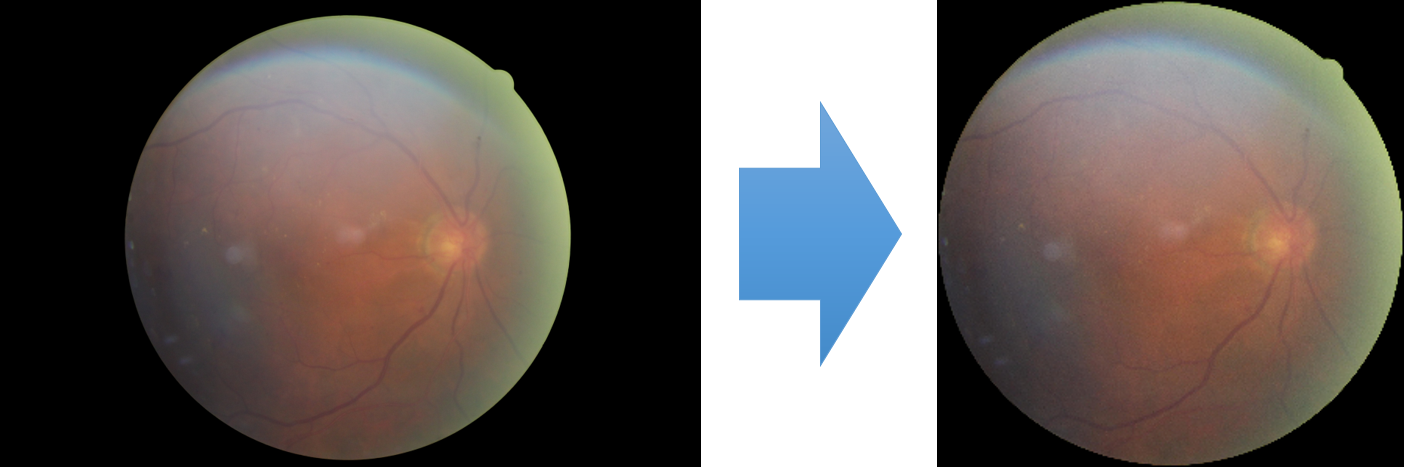}
    \caption{Cropping}
    \label{fig:fundus-form-1}
  \end{subfigure}
  \\
  \begin{subfigure}[t]{0.90\linewidth}
    \centering
    \includegraphics[height=3.5cm]{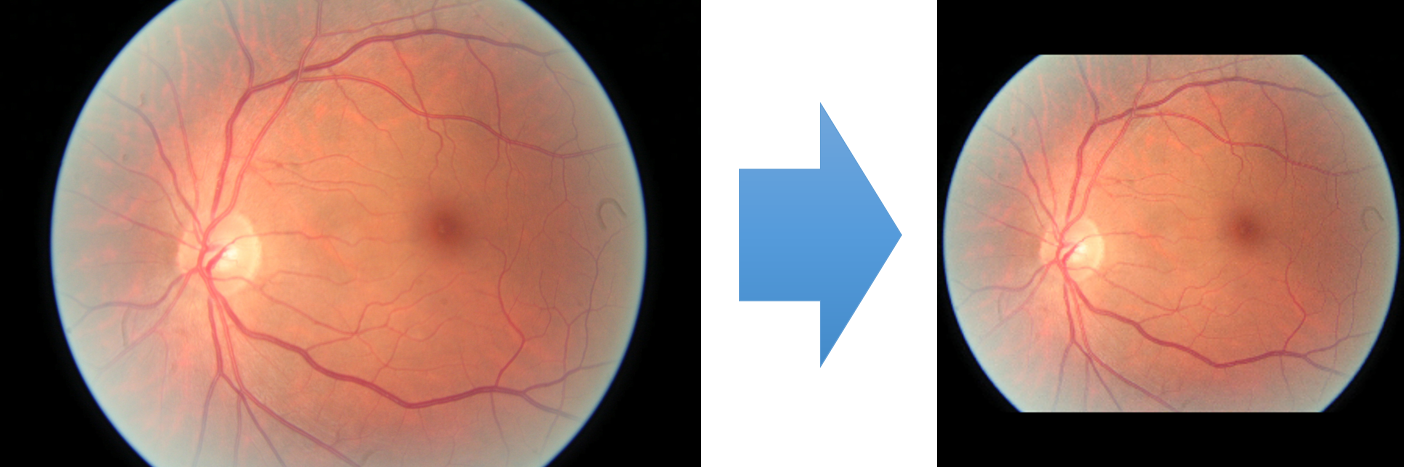}
    \caption{Blank padding}
    \label{fig:fundus-form-2}
  \end{subfigure}
  \caption{Preprocessing of fundus image data from Kaggle-DR and Messidor-2 datasets.}
  \label{fig:preprocessing}
\end{figure}

The image data used in our experiment were all unified into square-shaped images with resolutions $224\times 224$ or $384\times 384$ in our preprocessing procedures. For training each neural network model, only images of the same resolution were used. The original images came with either of the two forms as exemplified in Figure~\ref{fig:preprocessing}. In the first form (Figure~\ref{fig:fundus-form-1}), the entire fundus was visible in the image. We cropped the image such that the fundus was tightly fitted inside the square. In the second form of input images shown in Figure~\ref{fig:fundus-form-2}, part of the fundus was not visible. We padded blank strips to make the image square-shaped and in a unified resolution. See the provided code for further details.

\subsection{Training Ensembles of Deep Learning Models}

We implemented our neural networks under the PyTorch framework and then trained them on Linux workstations. The code is provided as part of the supplementary materials. 

\subsubsection{Data Augmentation}\label{sec:data-aug-train}
Data augmentation has proved to be an important technique for training \acf{DL} models that can prevent overfitting and can enhance model's generalization ability. It is also a key part of the \acf{TTA} technique~\cite{ayhan2018test}, one ensemble method evaluated in our experimental study and is reviewed earlier in Sec.~\ref{sec:ensemble-method}.

We utilized several different types of data augmentation operations at training time and also at test time for our \ac{TTA} models, using modules from the \texttt{torchvision} package. The operations include \texttt{RandomResizedCrop}, \texttt{adjust\_brightness}, \texttt{adjust\_saturation} and \texttt{adjust\_contrast} that randomly adjust the aspect ratio, the brightness, the saturation and the contrast respectively. The degree (strength) of data augmentation in our experiments was controlled by a multiplier $\gamma\in\{0.1,0.3,0.5,0.7\}$; see the provided code for further details.

\subsection{Constructing Ensemble Models}

Next we give the details how we train ensemble models using the three methods evaluated in this paper.

\paragraph{Stacking Ensembles}

The \ac{DL} models used to construct our stacking ensembles vary in their architecture, image data resolution, training set selection, the number of training epochs and data augmentation strengths. Two different \ac{CNN} architectures, Resnet34~\cite{he2016deep} and VGG16~\cite{simonyan2014very}, were used in our experiments. We used binary-crossentropy as the loss function and Adam as the optimizer. All parameters were initialized with the weights from pretrained  models trained provided by \texttt{torchvision} for the ImageNet classification task.

Since our experiments involved scanning different $\rho$ values, to reduce the total training effort, we first trained our models with non-incipient disease data (only SL0\,\&\,SL3\,\&\,SL4) for $130$ epochs, and then continued to train the resulting networks with all training data (SL0-4) till convergence. 

Most models reached an \ac{AUC} above $0.98$ on both the training and the validation sets. We discarded the bad performing models and put the rest into a pool. The retained models in the pool were then used as base learners for constructing stacking ensembles. To create an ensemble model instance, we randomly picked $K$ single learners from the pool. In our experiment, we evaluated $K=5$ and $K=10$, two ensemble sizes used in previous works~\cite{lakshminarayanan2017simple,gulshan2016development}. Their individual predictions were then combined and grouped for later analysis. 

To construct MC-dropout ensembles, we used the dropout models with dropout rate $0.25$ from the pool for conducting MC-dropout experiments. The MC-dropout models were sampled for $K\in\{5,10,20,50\}$ times, and the results were then combined and grouped for later analysis. For \ac{TTA}~\cite{ayhan2018test,wang2019aleatoric} ensembles, the diversity comes from the stochasticity injected to the inputs at test time. Uncertainty estimation for \textsc{mean} and \textsc{var} can then be obtained by repeatedly sampling the same network with stochastic inputs, as with MC-dropout models. We tested \ac{TTA} models with various degrees of test-time data augmentation $\gamma_\text{test}\in\{0.1,0.3,0.5,0.7\}$, and with different sample size (ensemble size) $K\in\{5,10,20,50\}$.

\subsection{Distribution of Uncertainty Scores}
\begin{figure}[b]
  \centering
  \begin{subfigure}[t]{0.32\linewidth}
    \centering
    \includegraphics[height=2.88cm]{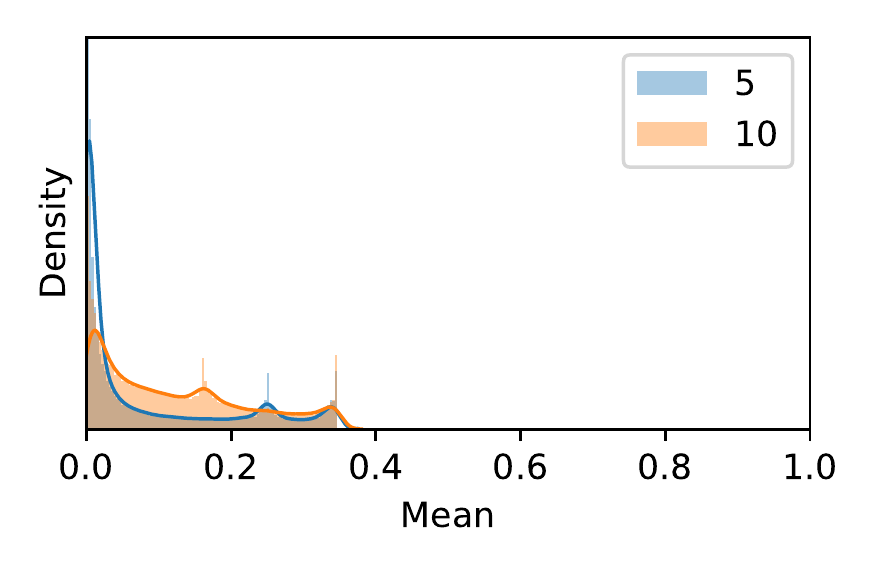}
    \caption{Stacking ensembles}
  \end{subfigure}
  \begin{subfigure}[t]{0.32\linewidth}
    \centering
    \includegraphics[height=2.88cm]{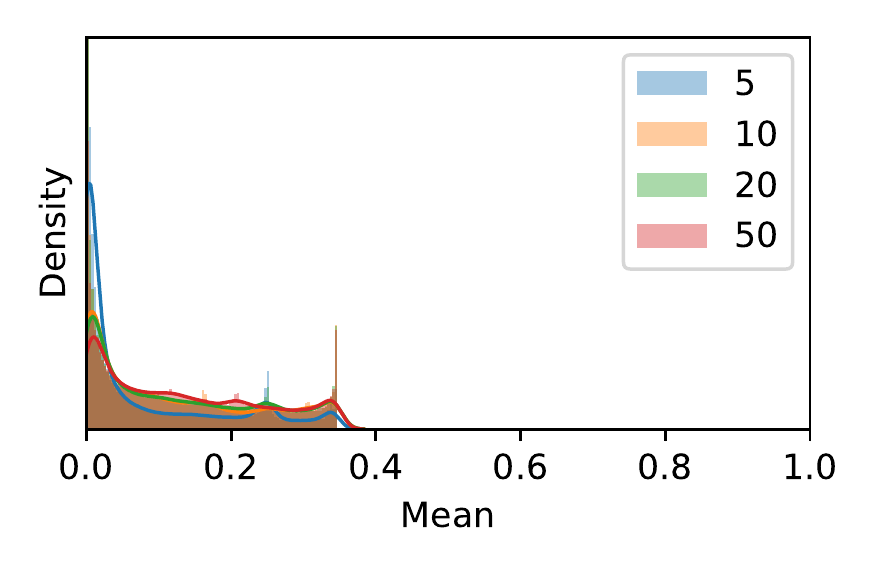}
    \caption{MC-dropout}
  \end{subfigure}
  \begin{subfigure}[t]{0.32\linewidth}
    \centering
    \includegraphics[height=2.88cm]{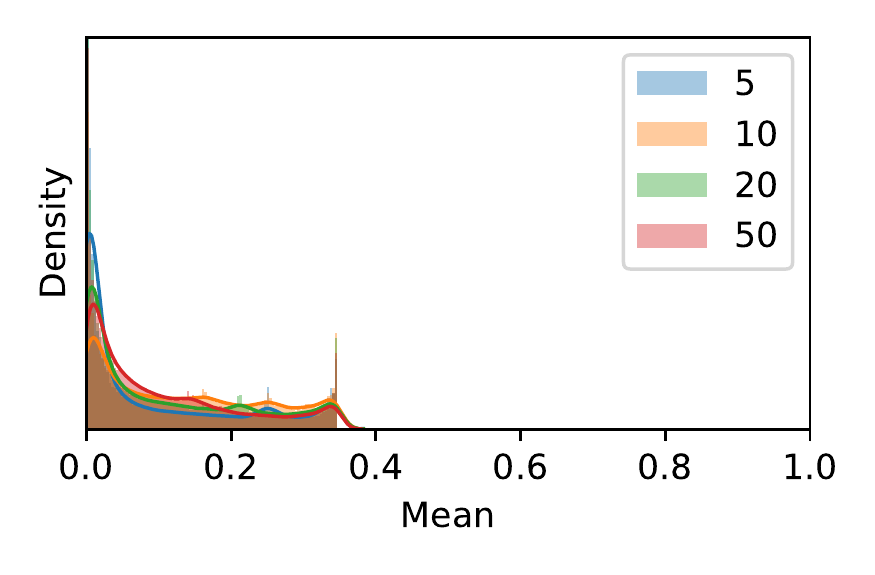}
    \caption{Test-time augmentation}
  \end{subfigure}

  \begin{subfigure}[t]{0.32\linewidth}
    \centering
    \includegraphics[height=2.88cm]{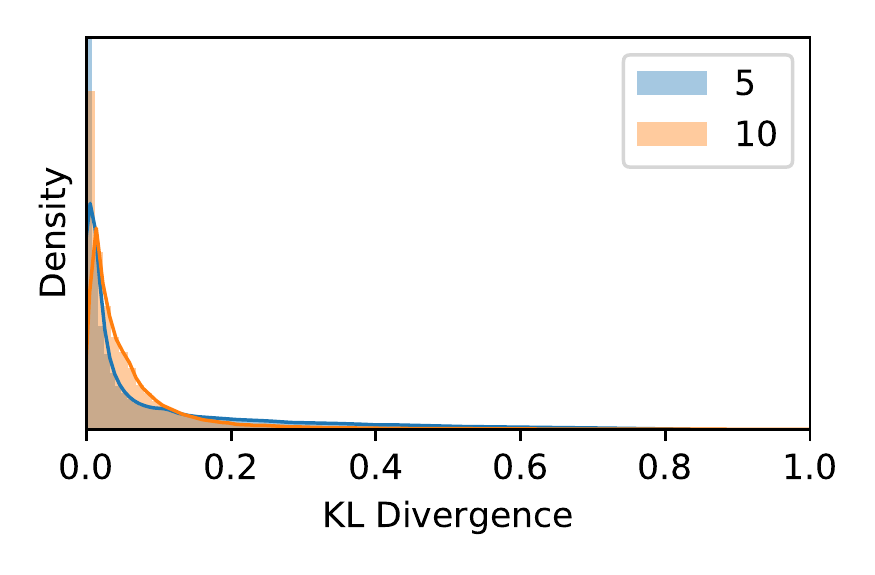}
    \caption{Stacking ensembles}
  \end{subfigure}
  \begin{subfigure}[t]{0.32\linewidth}
    \centering
    \includegraphics[height=2.88cm]{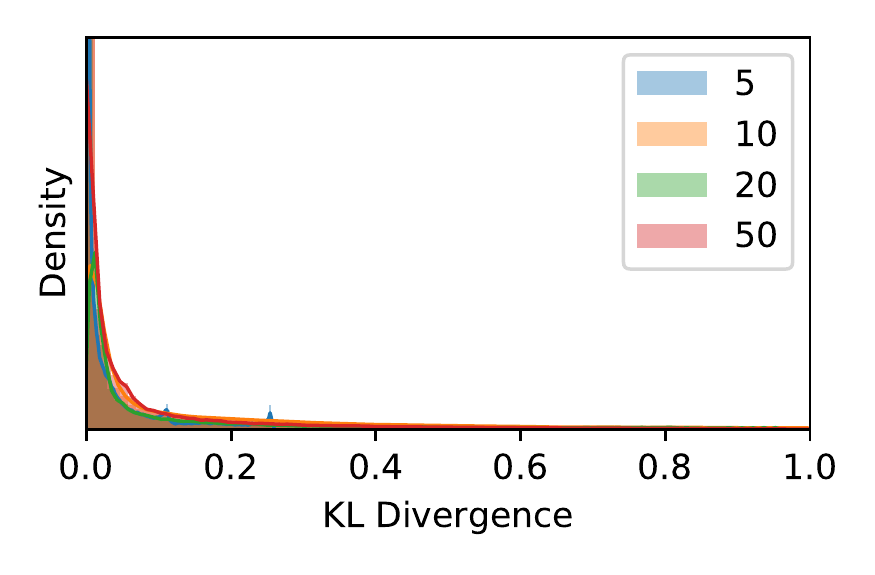}
    \caption{MC-dropout}
  \end{subfigure}
  \begin{subfigure}[t]{0.32\linewidth}
    \centering
    \includegraphics[height=2.88cm]{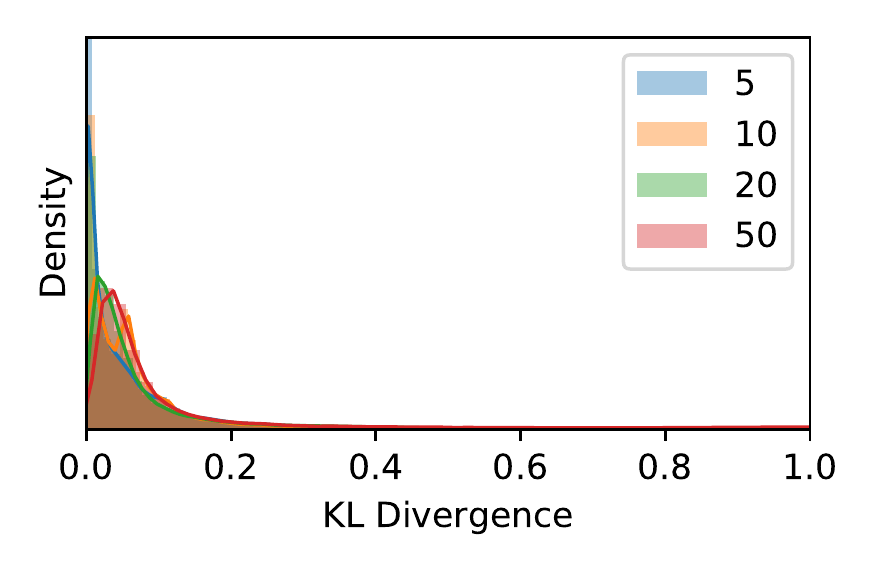}
    \caption{Test-time augmentation}
  \end{subfigure} 
  
  \begin{subfigure}[t]{0.32\linewidth}
    \centering
    \includegraphics[height=2.88cm]{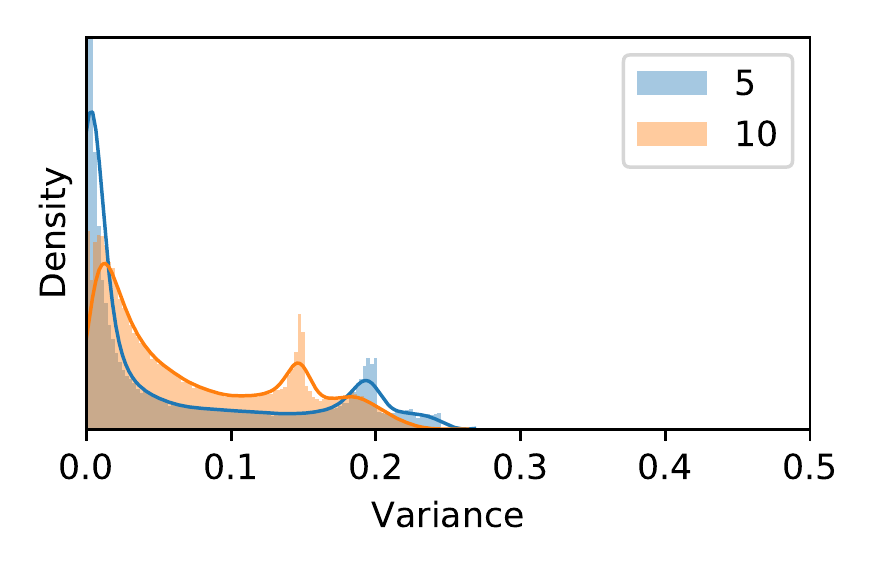}
    \caption{Stacking ensembles}
  \end{subfigure}
  \begin{subfigure}[t]{0.32\linewidth}
    \centering
    \includegraphics[height=2.88cm]{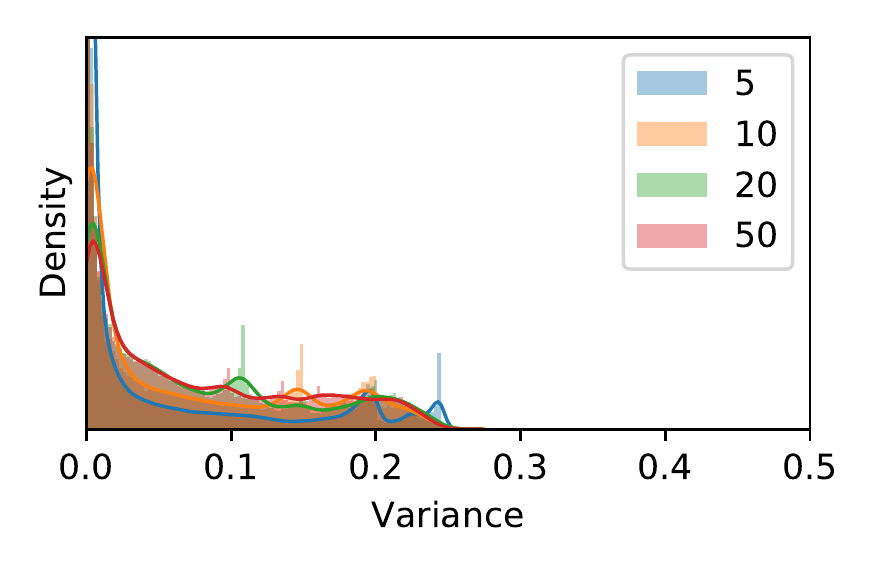}
    \caption{MC-dropout}
  \end{subfigure}
  \begin{subfigure}[t]{0.32\linewidth}
    \centering
    \includegraphics[height=2.88cm]{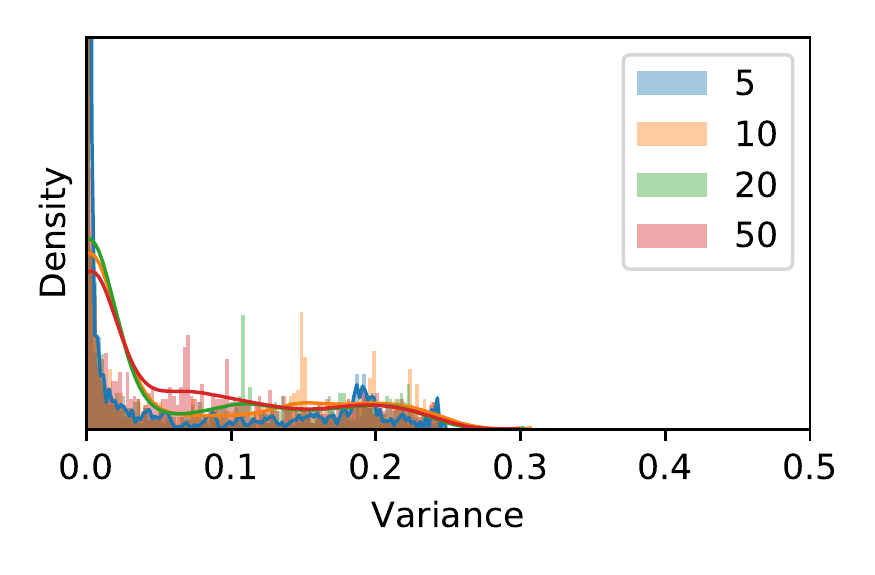}
    \caption{Test-time augmentation}
  \end{subfigure}
  \caption{Distributions of uncertainty scores for the Kaggle-{DR} dataset.}
  \label{fig:histogram_kaggle}
\end{figure}

\begin{figure}[b]
  \centering
  \begin{subfigure}[t]{0.32\linewidth}
    \centering
    \includegraphics[height=2.88cm]{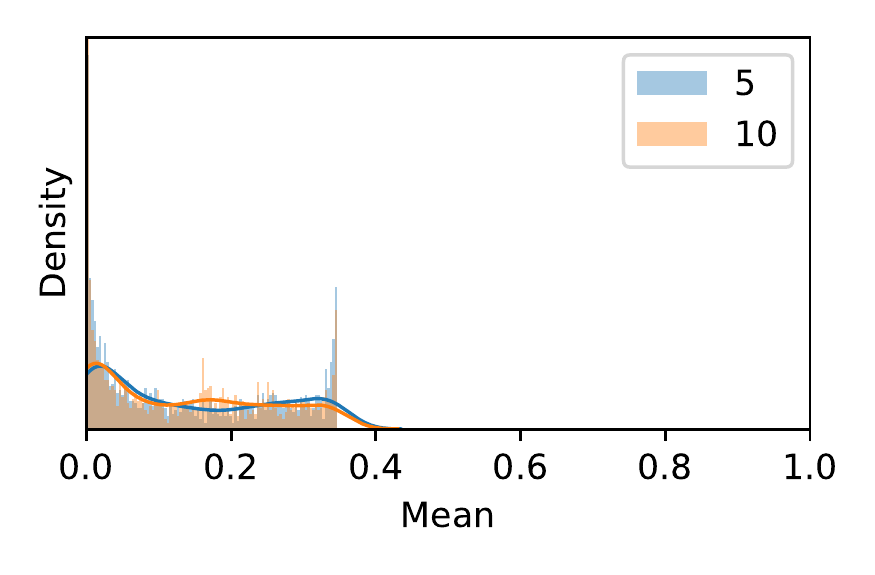}
    \caption{Stacking ensembles}
  \end{subfigure}
  \begin{subfigure}[t]{0.32\linewidth}
    \centering
    \includegraphics[height=2.88cm]{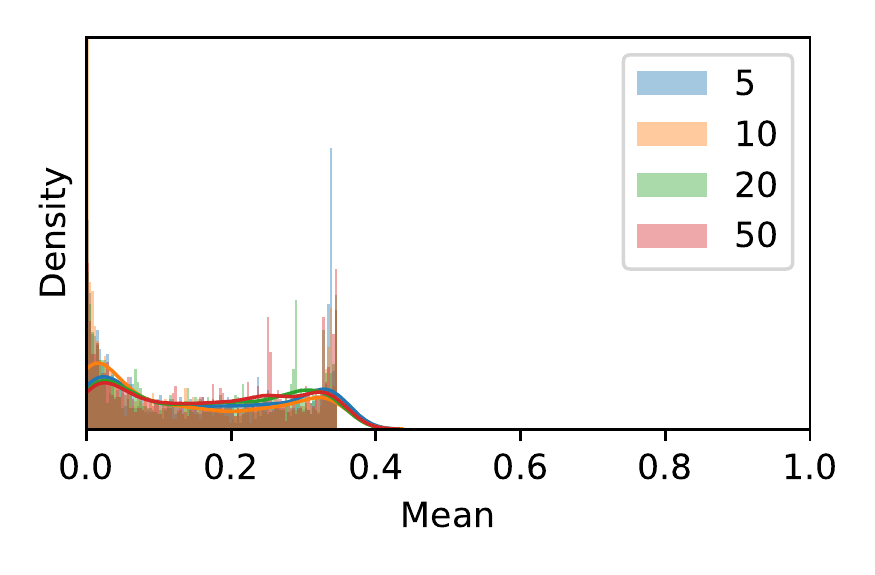}
    \caption{MC-dropout}
  \end{subfigure}
  \begin{subfigure}[t]{0.32\linewidth}
    \centering
    \includegraphics[height=2.88cm]{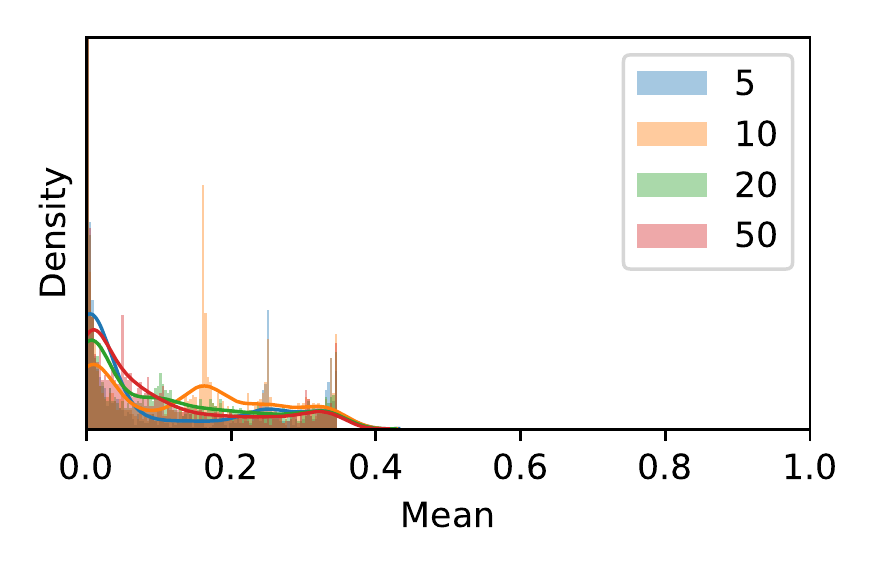}
    \caption{Test-time augmentation}
  \end{subfigure}

  \begin{subfigure}[t]{0.32\linewidth}
    \centering
    \includegraphics[height=2.88cm]{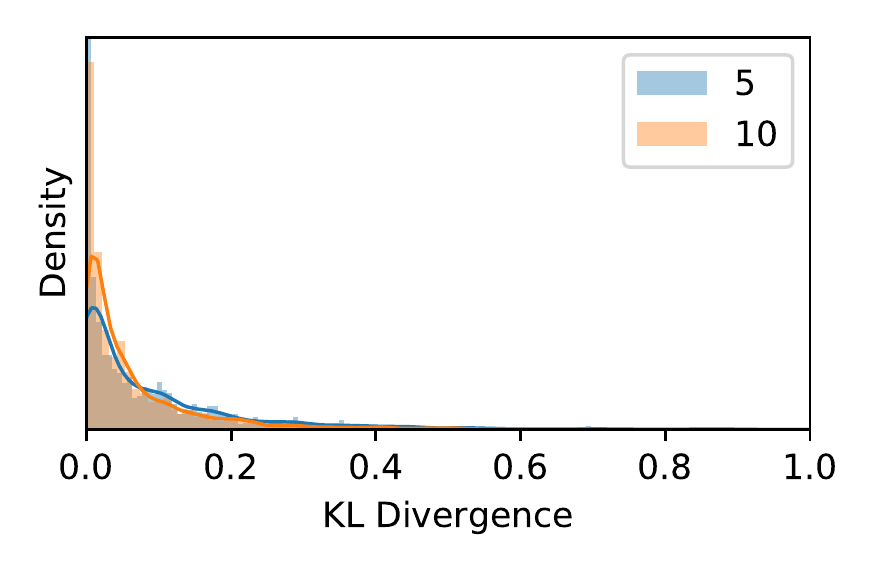}
    \caption{Stacking ensembles}
  \end{subfigure}
  \begin{subfigure}[t]{0.32\linewidth}
    \centering
    \includegraphics[height=2.88cm]{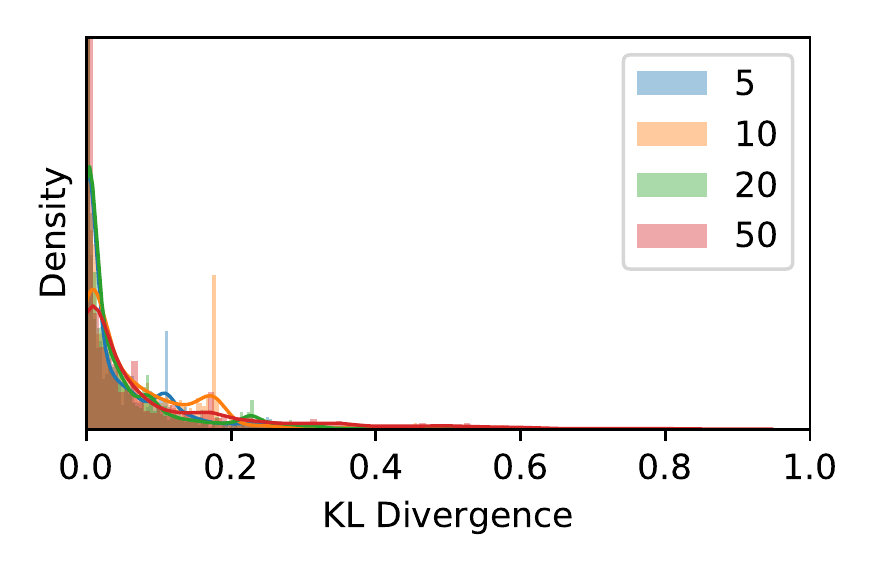}
    \caption{MC-dropout}
  \end{subfigure}
  \begin{subfigure}[t]{0.32\linewidth}
    \centering
    \includegraphics[height=2.88cm]{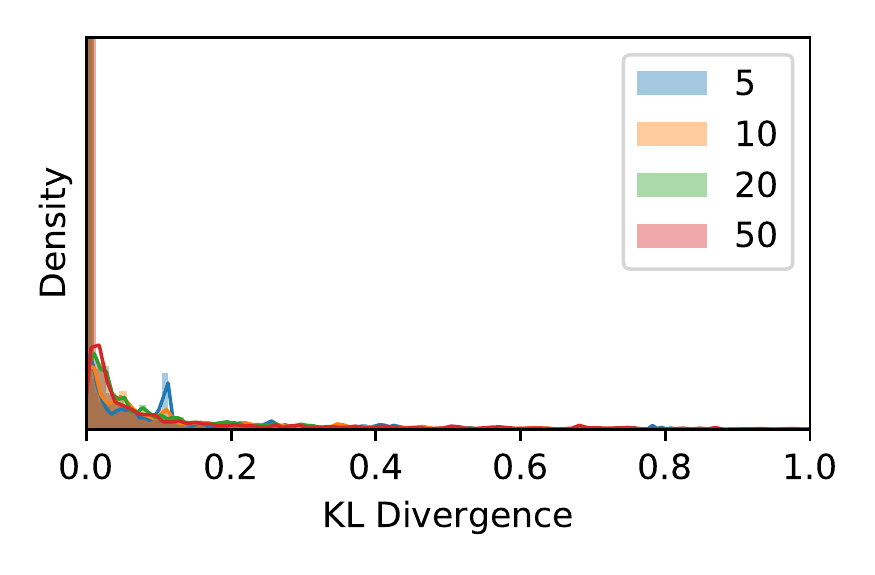}
    \caption{Test-time augmentation}
  \end{subfigure} 
  
  \begin{subfigure}[t]{0.32\linewidth}
    \centering
    \includegraphics[height=2.88cm]{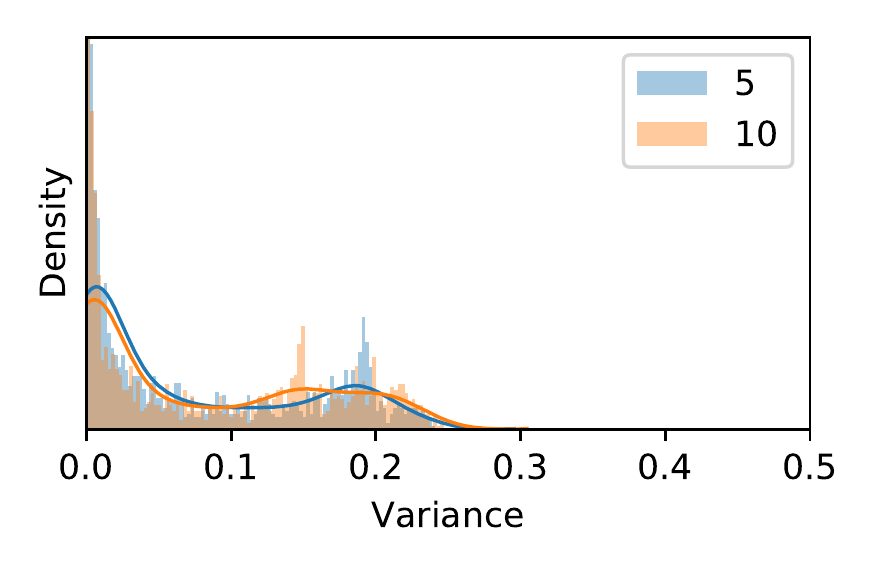}
    \caption{Stacking ensembles}
  \end{subfigure}
  \begin{subfigure}[t]{0.32\linewidth}
    \centering
    \includegraphics[height=2.88cm]{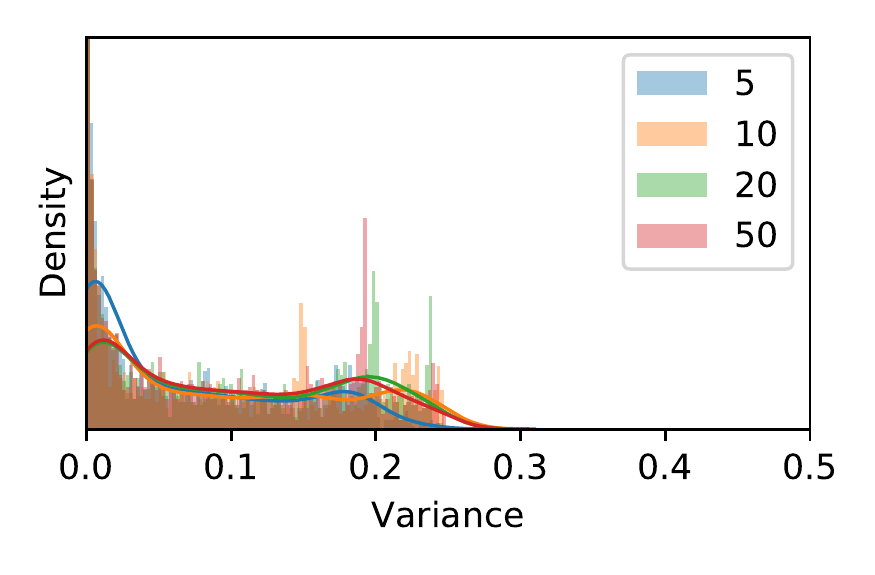}
    \caption{MC-dropout}
  \end{subfigure}
  \begin{subfigure}[t]{0.32\linewidth}
    \centering
    \includegraphics[height=2.88cm]{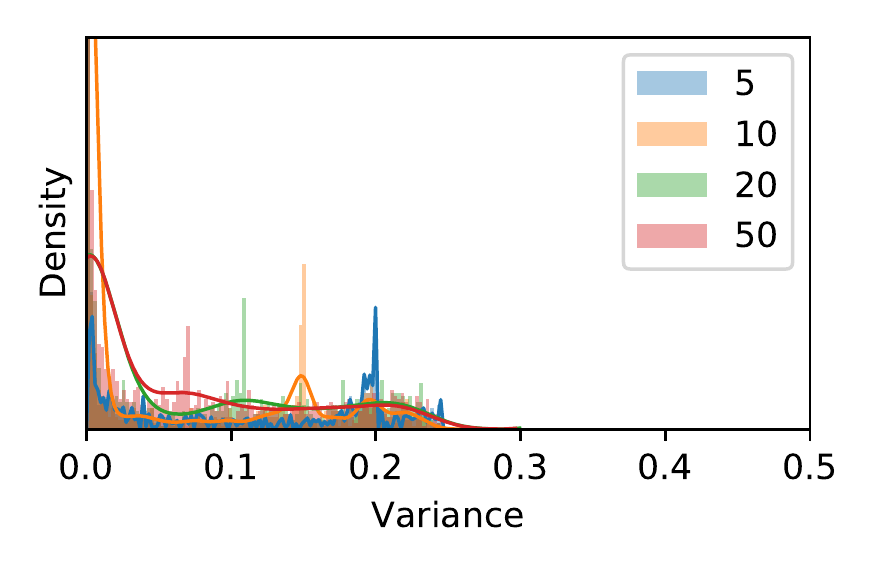}
    \caption{Test-time augmentation}
  \end{subfigure}
  \caption{Distributions of uncertainty scores for the Messidor-2 dataset.}
  \label{fig:histogram_messidor}
\end{figure}

\begin{figure}[b]
  \begin{subfigure}[t]{0.32\linewidth}
    \centering
    \includegraphics[height=2.88cm]{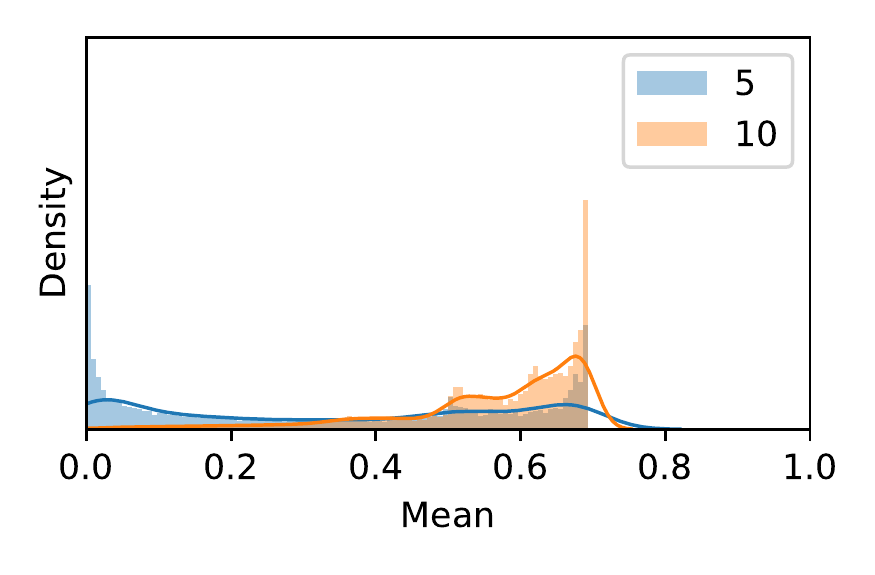}
    \caption{Stacking ensembles}
  \end{subfigure}
  \begin{subfigure}[t]{0.32\linewidth}
    \centering
    \includegraphics[height=2.88cm]{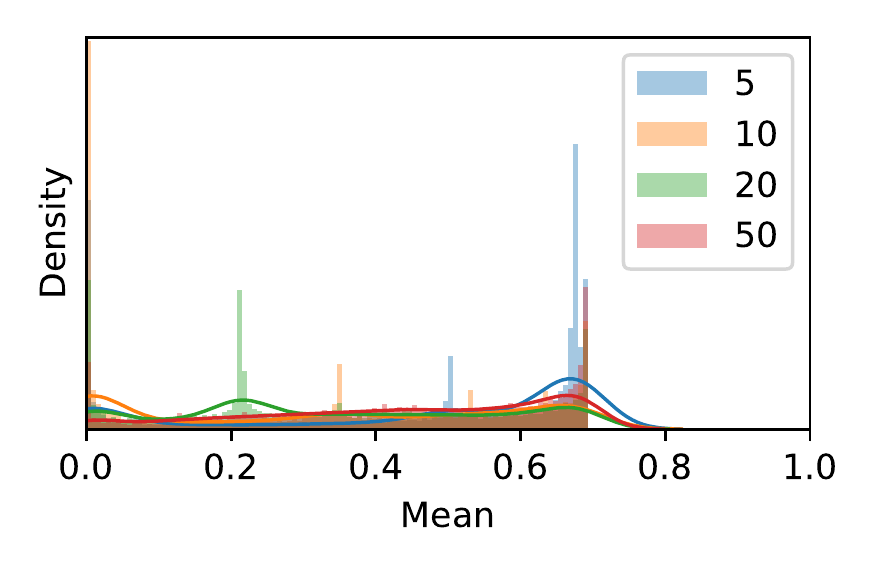}
    \caption{MC-dropout}
  \end{subfigure}
  \begin{subfigure}[t]{0.32\linewidth}
    \centering
    \includegraphics[height=2.88cm]{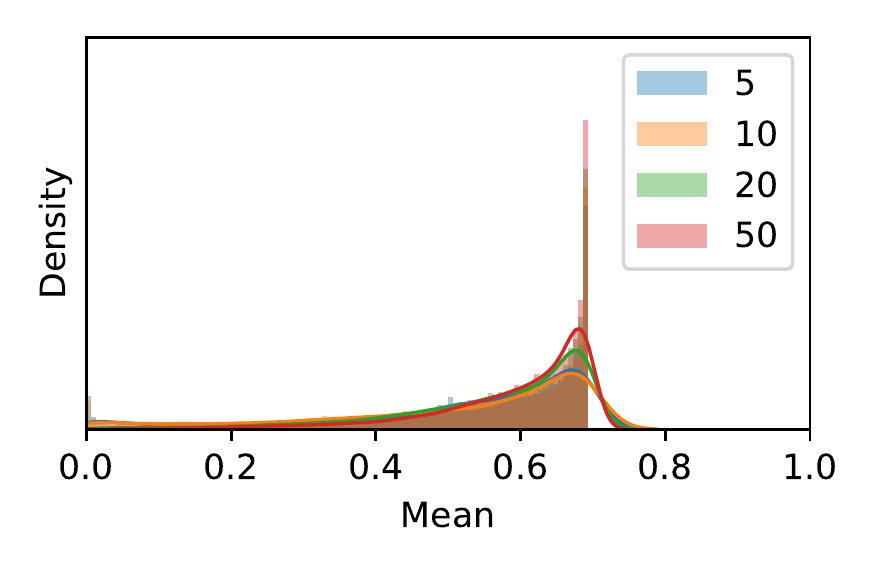}
    \caption{Test-time augmentation}
  \end{subfigure}
  
  \begin{subfigure}[t]{0.32\linewidth}
    \centering
    \includegraphics[height=2.88cm]{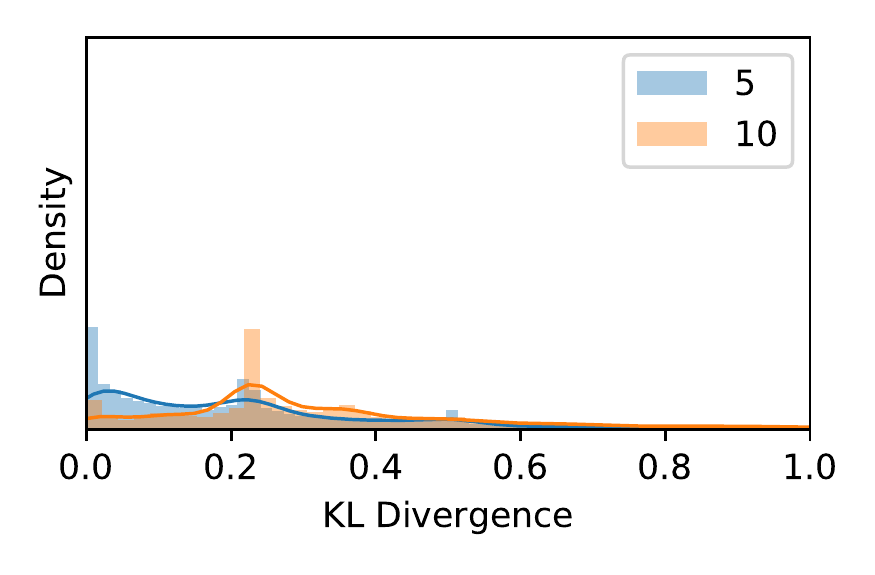}
    \caption{Stacking ensembles}
  \end{subfigure}
  \begin{subfigure}[t]{0.32\linewidth}
    \centering
    \includegraphics[height=2.88cm]{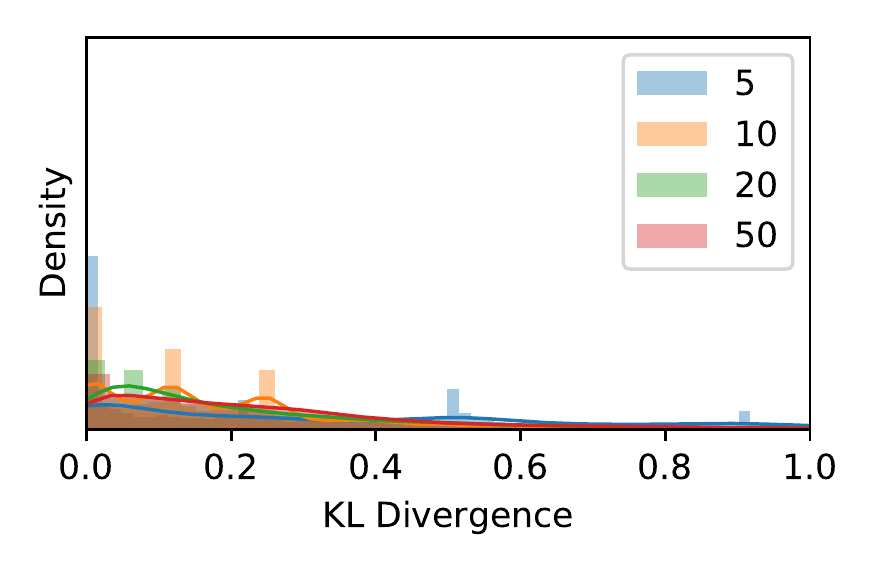}
    \caption{MC-dropout}
  \end{subfigure}
  \begin{subfigure}[t]{0.32\linewidth}
    \centering
    \includegraphics[height=2.88cm]{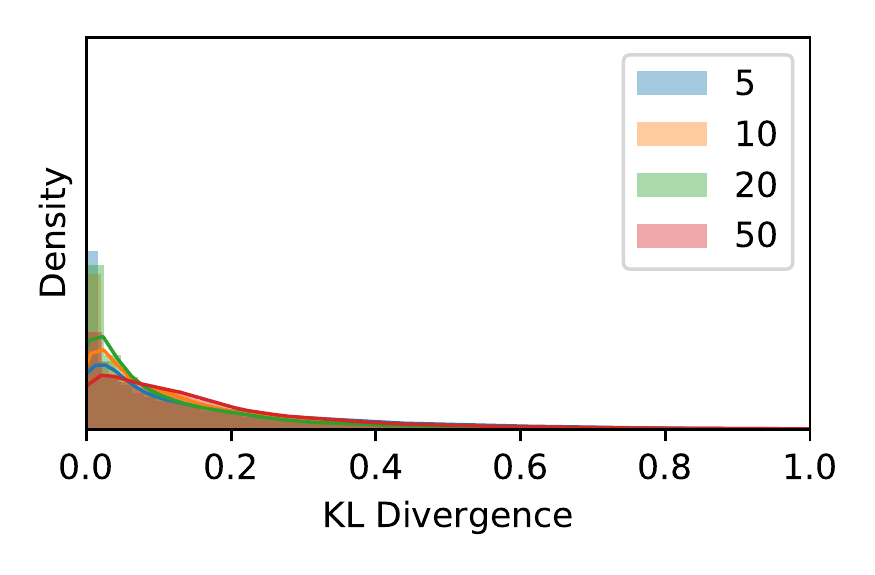}
    \caption{Test-time augmentation}
  \end{subfigure}
  
  \begin{subfigure}[t]{0.32\linewidth}
    \centering
    \includegraphics[height=2.88cm]{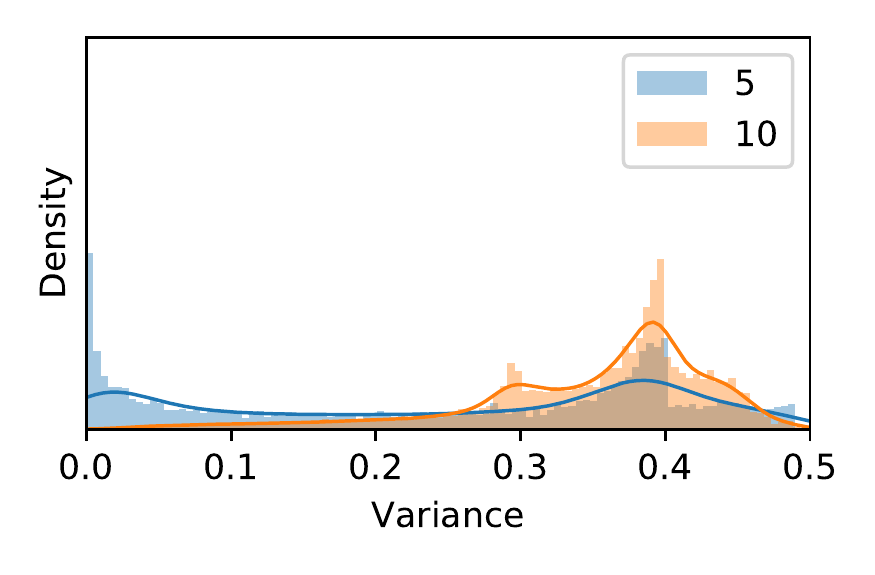}
    \caption{Stacking ensembles}
  \end{subfigure}
  \begin{subfigure}[t]{0.32\linewidth}
    \centering
    \includegraphics[height=2.88cm]{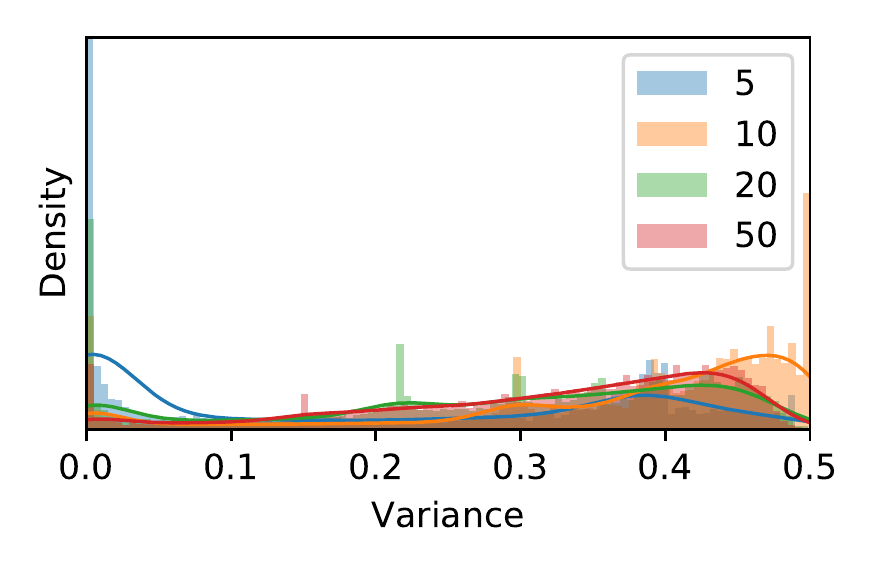}
    \caption{MC-dropout}
  \end{subfigure}
  \begin{subfigure}[t]{0.32\linewidth}
    \centering
    \includegraphics[height=2.88cm]{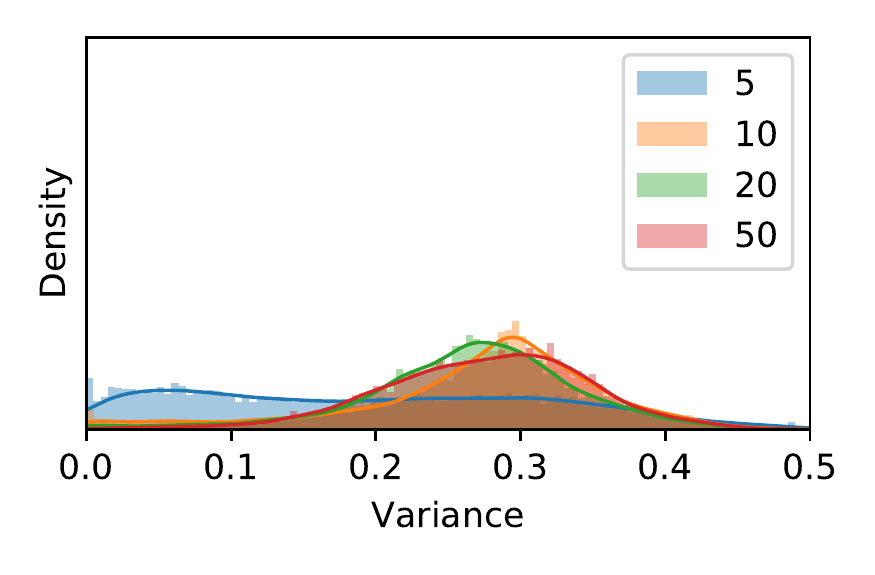}
    \caption{Test-time augmentation}
  \end{subfigure}
  \caption{Distributions of uncertainty scores for the ImageNet dataset.}
  \label{fig:histogram_imagenet}
\end{figure}

\begin{figure}[b]
  \begin{subfigure}[t]{0.32\linewidth}
    \centering
    \includegraphics[height=2.88cm]{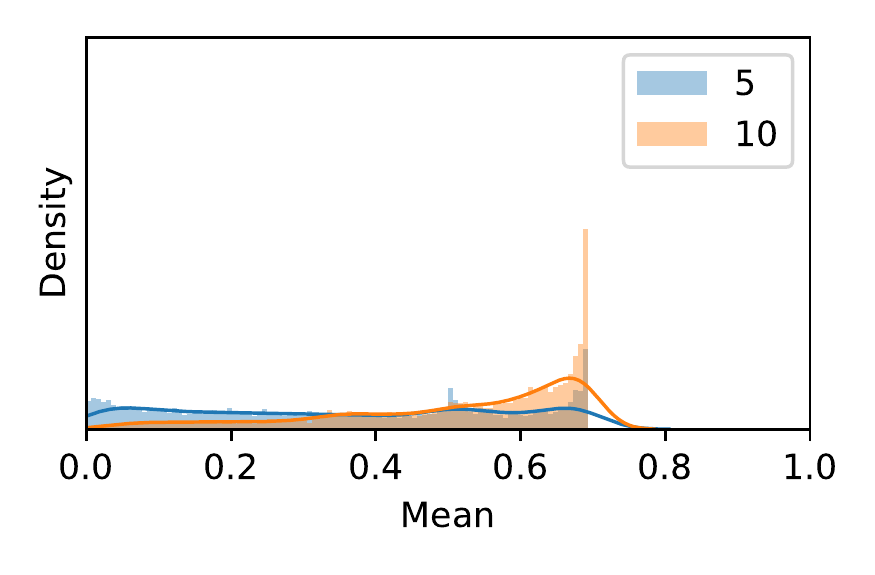}
    \caption{Stacking ensembles}
  \end{subfigure}
  \begin{subfigure}[t]{0.32\linewidth}
    \centering
    \includegraphics[height=2.88cm]{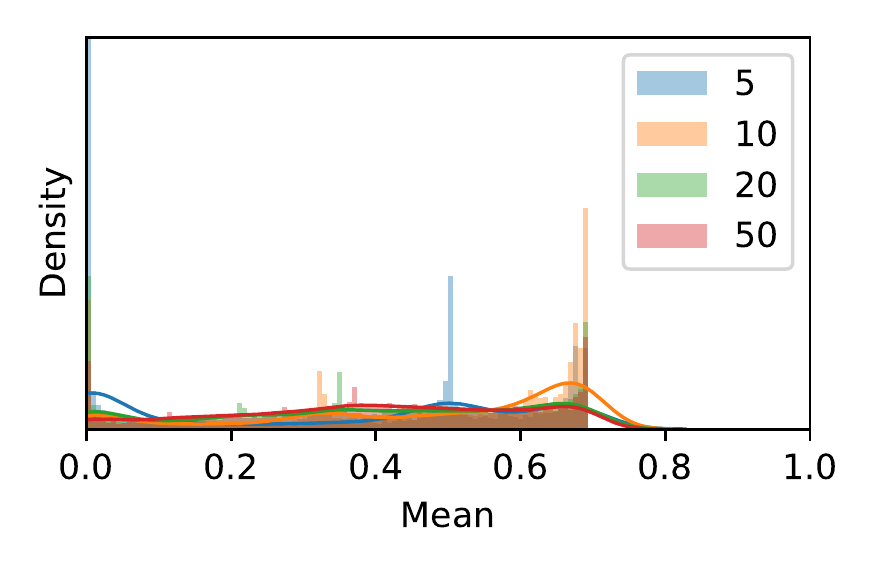}
    \caption{MC-dropout}
  \end{subfigure}
  \begin{subfigure}[t]{0.32\linewidth}
    \centering
    \includegraphics[height=2.88cm]{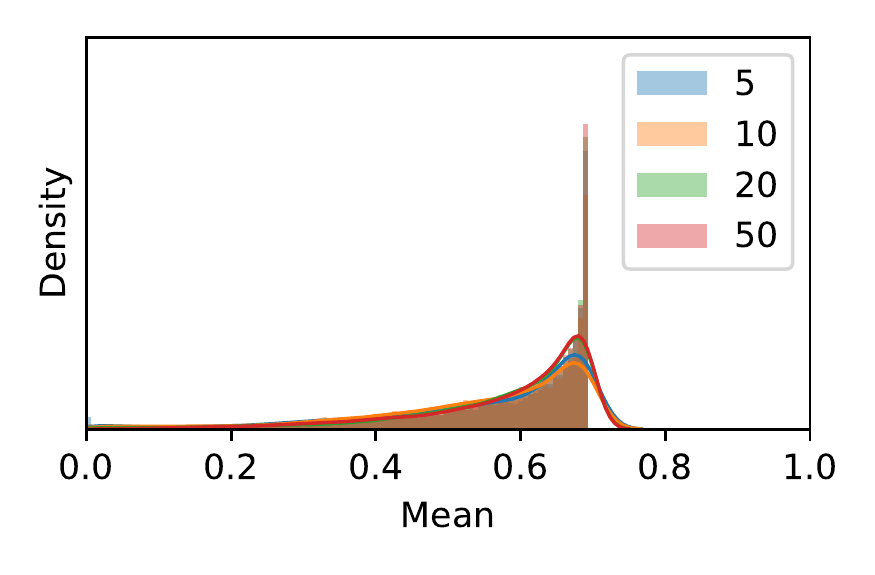}
    \caption{Test-time augmentation}
  \end{subfigure}
  
  \begin{subfigure}[t]{0.32\linewidth}
    \centering
    \includegraphics[height=2.88cm]{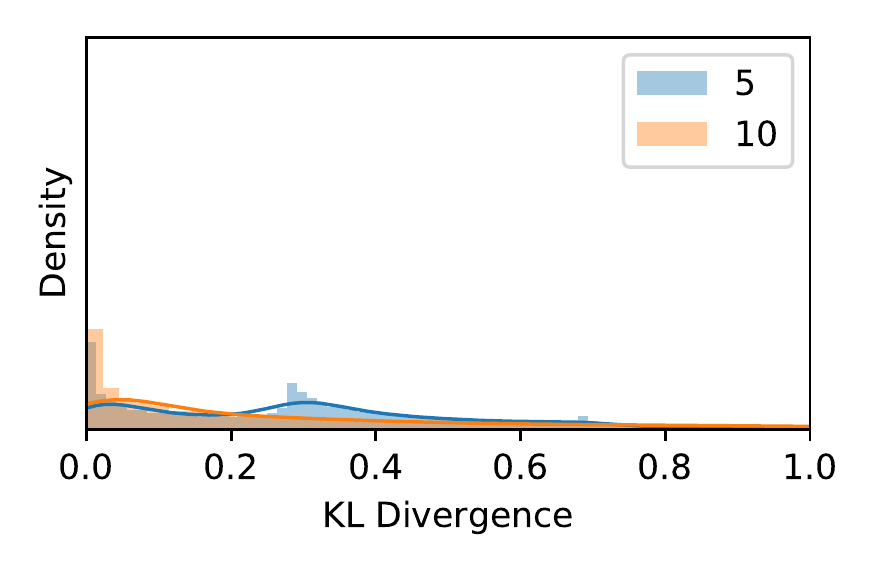}
    \caption{Stacking ensembles}
  \end{subfigure}
  \begin{subfigure}[t]{0.32\linewidth}
    \centering
    \includegraphics[height=2.88cm]{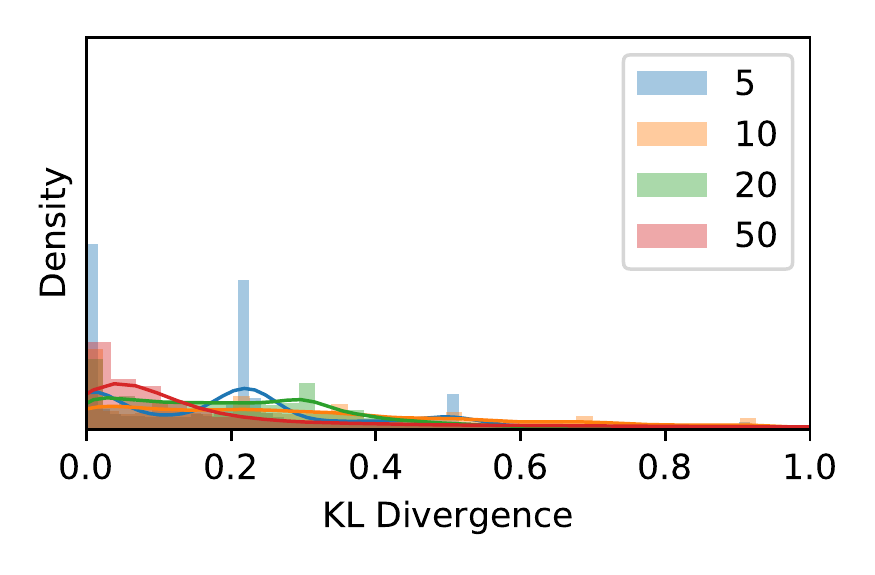}
    \caption{MC-dropout}
  \end{subfigure}
  \begin{subfigure}[t]{0.32\linewidth}
    \centering
    \includegraphics[height=2.88cm]{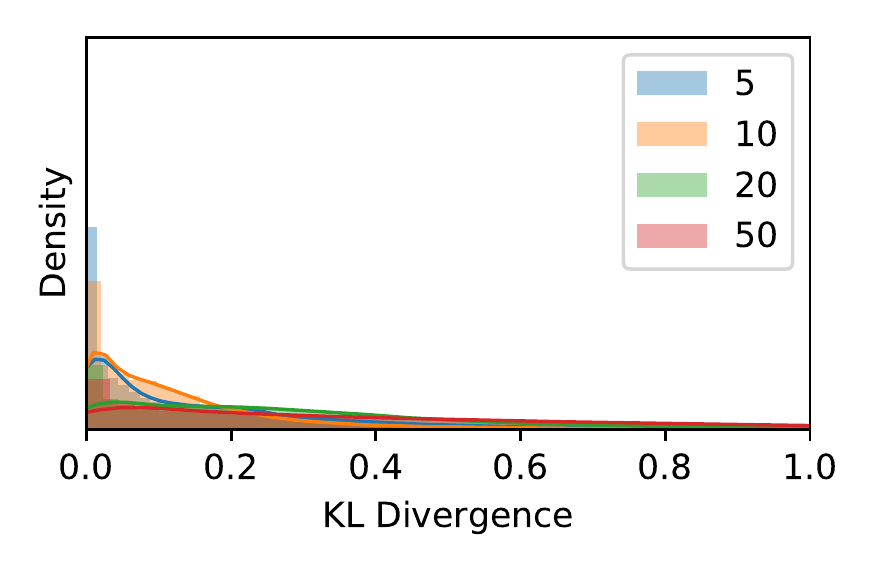}
    \caption{Test-time augmentation}
  \end{subfigure}
  
  \begin{subfigure}[t]{0.32\linewidth}
    \centering
    \includegraphics[height=2.88cm]{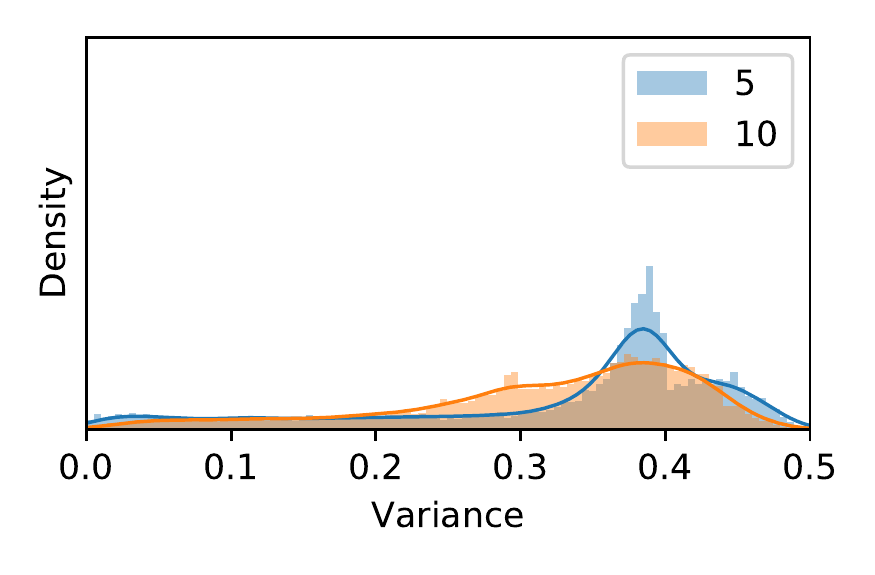}
    \caption{Stacking ensembles}
  \end{subfigure}
  \begin{subfigure}[t]{0.32\linewidth}
    \centering
    \includegraphics[height=2.88cm]{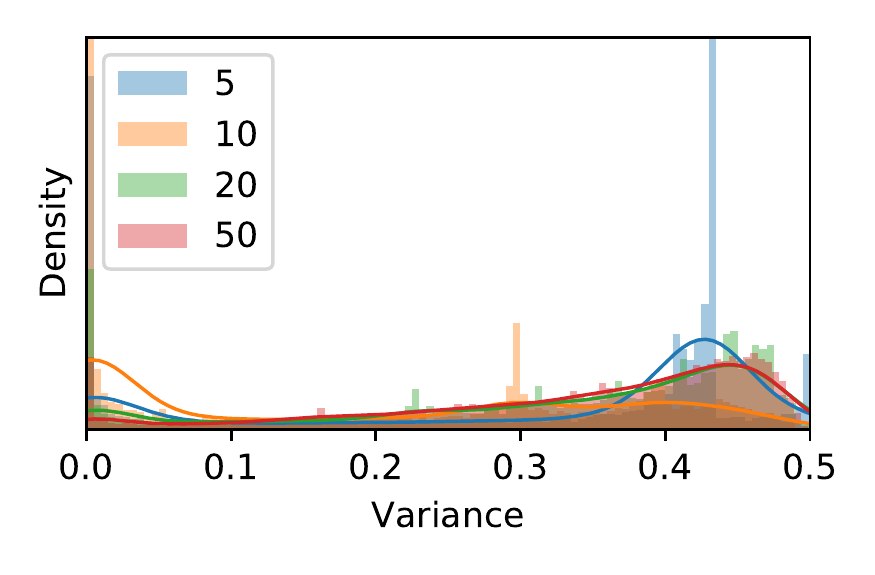}
    \caption{MC-dropout}
  \end{subfigure}
  \begin{subfigure}[t]{0.32\linewidth}
    \centering
    \includegraphics[height=2.88cm]{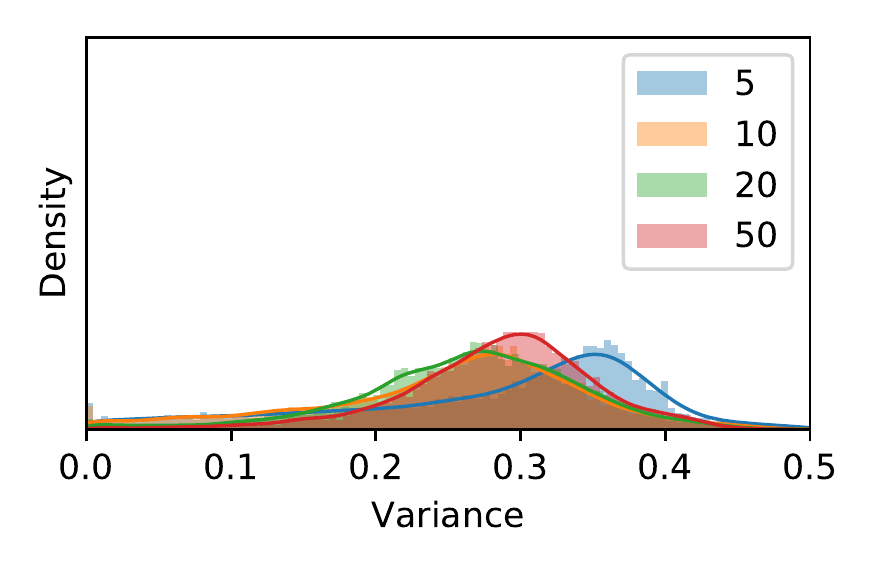}
    \caption{Test-time augmentation}
  \end{subfigure}
  \caption{Distributions of uncertainty scores for the CIFAR-10 dataset.}
  \label{fig:histogram_cifar10}
\end{figure}

As discussed in Section~\ref{sec:uncertainty-scores} and Section~\ref{sec:comparison-uncertainty} in the main paper, the \texttt{mean} metric and the stacking ensemble will have better performance in the precision (specificity) on the ambiguous data. Here, more detailed results are shown in Figures~\ref{fig:histogram_kaggle}\,\&\,\ref{fig:histogram_cifar10} and Table~\ref{tab:severity-ratio}. Figures~\ref{fig:histogram_kaggle}\,\&\,\ref{fig:histogram_messidor} show the histograms of the uncertainty score for Kaggle-DR and Messidor-2 datasets that are the \ac{i.d.} dataset in our experiment and Figure\ref{fig:histogram_imagenet} \&~\ref{fig:histogram_cifar10} show the histograms for ImageNet and CIFAR-10 datasets, which is the \ac{o.o.d.} datasets in our experiment. Each group of histograms contains results from the three evaluated ensemble methods (stacking ensemble, MC-dropout and \ac{TTA}) and the three uncertainty metrics (\textsc{mean}, \textsc{var} and \textsc{kl}). Additional detailed results not displayed in Figure~\ref{fig:table2} can be found in Table~\ref{tab:severity-ratio}, which shows the proportion of the data of different \acp{SL} varies across different $\theta$. For comparison, we also included in Table~\ref{tab:severity-ratio} the results from single learners, and the proportions of data of different \acp{SL} (before any selection was made).

\subsection{Identifying False Negative Decisions Using Prediction Uncertainties}
As reported in Section~\ref{sec:fnp} in the main paper, \textsc{mean} had better performance compared to \textsc{var}. Here we show the detailed results for both datasets. The \ac{FNP} values under different settings ($\rho$ and $q$) can be found in Tables~\ref{tab:fnp-kaggle}\,\&\,\ref{tab:fnp-messidor}. In addition to the \ac{FNP} percentage, we also report the numbers of uncertain negatives and the count of false negatives among uncertain negatives. The numbers of remaining false negatives after involving human diagnosticians are reported in Tables~\ref{tab:num-false-negatives-DR} \&~\ref{tab:num-false-negatives-messidor}.

\begin{table}[htb]
\centering
\caption{A breakdown of the high-uncertainty data points across different \acp{SL}. The five percentage numbers in each entry show the respective proportion of the data of the five severity levels, SL0 to SL4, among all high-uncertainty data points.}\label{tab:severity-ratio}
\resizebox{\textwidth}{!}{%
\begin{tabular}{ccccccc}
\hline
\textbf{Dataset} & \textbf{$\theta$ (\%)} & \textbf{Uncertainty metric} & \textbf{Stacking ensemble} & \textbf{MC-dropout} & \textbf{Test time augmentation} & \textbf{Single learner} \\ \hline
\multirow{18}{*}{cuadros2009eyepacs-DR} &  & Raw Data Distribution & \multicolumn{4}{c}{73.6, 7.0, 14.8, 2.3, 2.1} \\ \cline{3-7} 
 & \multicolumn{1}{l}{} & \multicolumn{1}{l}{} & \multicolumn{1}{l}{} & \multicolumn{1}{l}{} & \multicolumn{1}{l}{} & \multicolumn{1}{l}{} \\
 & \multirow{3}{*}{10} & \textsc{mean} & 8.24, 15.76, 69.82, 3.47, 2.71 & 44.27, 7.32, 37.94, 5.63, 4.85 & 30.83, 15.07, 43.69, 4.9, 5.51 & 66.9, 8.6, 17.3, 3.7, 3.6 \\
 &  & \textsc{var} & 15.9, 16.41, 61.37, 2.98, 3.34 & 57.8, 6.42, 28.27, 3.78, 3.73 & 35.73, 13.73, 41.75, 3.81, 4.98 & --- \\
 &  & \textsc{kl} & 15.67, 16.38, 61.21, 3.94, 2.8 & 49.08, 8.17, 33.62, 4.61, 4.52 & 34.2, 16.62, 38.28, 5.21, 5.7 & --- \\
 &  &  &  &  &  &  \\
 & \multirow{3}{*}{5} & \textsc{mean} & 5.09, 17.77, 73.41, 1.2, 2.53 & 30.57, 9.31, 52.33, 3.61, 4.18 & 21.75, 19.38, 53.41, 3.89, 1.57 & 61.5, 9.7, 22.1, 3.4, 3.3 \\
 &  & \textsc{var} & 11.44, 17.25, 66.91, 1.54, 2.87 & 36.83, 9.62, 45.25, 3.81, 4.49 & 24.84, 20.51, 49.61, 3.19, 1.85 & --- \\
 &  & \textsc{kl} & 11.19, 17.79, 67.17, 1.63, 2.22 & 33.91, 8.69, 42.83, 6.64, 7.93 & 28.5, 19.11, 44.32, 4.27, 3.81 & --- \\
 &  &  &  &  &  &  \\
 & \multirow{3}{*}{2} & \textsc{mean} & 1.65, 19.09, 77.52, 1.28, 0.47 & 14.68, 12.6, 67.66, 2.55, 2.51 & 8.74, 23.49, 65.73, 1.04, 1 & 68.7, 8.8, 17.7, 2.1, 2.7 \\
 &  & \textsc{var} & 1.73, 16.88, 76.19, 2.94, 2.26 & 14.67, 11.31, 66.7, 0.96, 6.36 & 9.7, 22.43, 62.46, 2.59, 2.82 & --- \\
 &  & \textsc{kl} & 1.8, 18.72, 76.69, 1.75, 1.03 & 16.25, 10.74, 64.87, 3.65, 4.49 & 11.16, 23.23, 61.16, 2.48, 1.97 & --- \\
 &  &  &  &  &  &  \\
 & \multirow{3}{*}{1} & \textsc{mean} & 1.4, 17.23, 79.95, 0.19, 1.23 & 10.77, 12.01, 71.45, 2.5, 3.26 & 6.49, 21.93, 66.38, 1.9, 3.31 & 61.5, 7.7, 19.9, 2.8, 3.0 \\
 &  & \textsc{var} & 1.48, 18.09, 78.83, 1.52, 0.08 & 11.12, 13.39, 68.53, 0.42, 6.54 & 6.18, 26.11, 63.16, 1.58, 2.97 & --- \\
 &  & \textsc{kl} & 1.03, 18.41, 78.6, 0.86, 1.1 & 13.9, 11.56, 67.62, 3.27, 3.65 & 8.77, 24.81, 62.89, 1.92, 1.61 & --- \\
 &  &  &  &  &  &  \\ \hline
\multicolumn{1}{l}{\multirow{17}{*}{Messidor-2}} & \multicolumn{1}{l}{} & \multicolumn{1}{l}{Raw Data Distribution} & \multicolumn{4}{c}{58.2, 15.4, 21.4, 4.3, 2.0} \\ \cline{3-7} 
\multicolumn{1}{l}{} & \multicolumn{1}{l}{} & \multicolumn{1}{l}{} & \multicolumn{4}{c}{} \\
\multicolumn{1}{l}{} & \multirow{3}{*}{10} & \textsc{mean} & 11.07, 17.15, 65.83, 2.98, 2.98 & 42.39, 13.82, 36.54, 4.08, 3.16 & 33.9, 26.72, 32.55, 4.66, 2.17 & 57.0, 16.3, 20.5, 4.0, 2.2 \\
\multicolumn{1}{l}{} &  & \textsc{var} & 20.26, 11.3, 63.58, 2.24, 2.61 & 50.74, 12.8, 26.82, 6.3, 3.35 & 42.83, 29.24, 20.85, 4.22, 2.86 & --- \\
\multicolumn{1}{l}{} &  & \textsc{kl} & 20.72, 12.04, 63.01, 2.33, 1.89 & 50.3, 13.64, 28.15, 4.52, 3.39 & 41.98, 28.63, 21.21, 5.81, 2.37 & --- \\
\multicolumn{1}{l}{} &  &  &  &  &  &  \\
\multicolumn{1}{l}{} & \multirow{3}{*}{5} & \textsc{mean} & 7.32, 20.39, 68.35, 1.67, 2.26 & 33.34, 14.9, 43.98, 3.91, 3.87 & 21.59, 35.06, 38, 2.98, 2.36 & 61.5, 16.0, 18.8, 1.8, 1.9 \\
\multicolumn{1}{l}{} &  & \textsc{var} & 9.87, 18.64, 66.33, 2.88, 2.28 & 36.41, 14.85, 42.14, 3.25, 3.34 & 29.47, 31.16, 32.48, 3.38, 3.51 & --- \\
\multicolumn{1}{l}{} &  & \textsc{kl} & 9.53, 18.8, 66.09, 2.78, 2.8 & 41.35, 14.61, 37.43, 2.9, 3.7 & 29.67, 32.73, 29.8, 4.23, 3.57 & --- \\
\multicolumn{1}{l}{} &  &  &  &  &  &  \\
\multicolumn{1}{l}{} & \multirow{3}{*}{2} & \textsc{mean} & 4.51, 23.23, 69.68, 1.58, 1 & 28.61, 14.08, 47.09, 5.28, 4.93 & 9.69, 38.29, 49.41, 1.24, 1.37 & 60.6, 19.7, 15.4, 2.4, 1.9 \\
\multicolumn{1}{l}{} &  & \textsc{var} & 6.05, 21.51, 67.69, 2.13, 2.62 & 30.48, 15.8, 44.04, 6.09, 3.59 & 14.64, 36.6, 44.19, 1.64, 2.93 & --- \\
\multicolumn{1}{l}{} &  & \textsc{kl} & 6.71, 22.07, 67.32, 2.12, 1.79 & 31.36, 15.56, 45.45, 4.58, 3.05 & 18.85, 36.28, 39.91, 2.06, 2.9 & --- \\
\multicolumn{1}{l}{} &  &  &  &  &  &  \\
\multicolumn{1}{l}{} & \multirow{3}{*}{1} & \textsc{mean} & 1.55, 25.31, 72.17, 0.53, 0.44 & 19.73, 16.3, 60.36, 1.08, 2.54 & 8.14, 38.56, 52.24, 1.03, 0.02 & 60.4, 15.5, 19.0, 3.0, 2.1 \\
\multicolumn{1}{l}{} &  & \textsc{var} & 1.28, 24, 72.15, 1.75, 0.83 & 25.39, 15.11, 51.8, 2.59, 5.11 & 10.6, 39.71, 47.79, 0.78, 1.13 & --- \\
\multicolumn{1}{l}{} &  & \textsc{kl} & 2.06, 23.15, 72.72, 2, 0.08 & 23.17, 14.03, 59.74, 1.81, 1.25 & 11.49, 40.22, 45.21, 1.92, 1.16 & --- \\
\multicolumn{1}{l}{} &  &  &  &  &  & \\ 
\hline
\end{tabular}
}
\end{table}

\begin{table}[p]
\centering
\caption{Performance in terms of false negative precision numbers for the Kaggle-DR dataset. The fraction in each entry shows the number of false negatives (the numerator) and the number of uncertain negatives (the denominator). The percentage numbers in the parentheses are the corresponding false negative precision values.}
\label{tab:fnp-kaggle}
\resizebox{\textwidth}{!}{%
\begin{tabular}{cccccccc}
\hline
\textbf{\begin{tabular}[c]{@{}c@{}}Ensemble\\ Method\end{tabular}} & \textbf{$\rho$} & \textbf{\begin{tabular}[c]{@{}c@{}}Uncertainty\\ Metric\end{tabular}} & \textbf{$q=1$} & \textbf{$q=2$} & \textbf{$q=5$} & \textbf{$q=10$} & \textbf{$q=15$} \\ \hline
\multirow{21}{*}{\textbf{\begin{tabular}[c]{@{}c@{}}Stacking\\ Ensemble\end{tabular}}} &  &  &  &  &  &  &  \\
 & \multirow{3}{*}{0.2} & \textsc{mean} & 286.45\,/\,580.59 (49.34\%) & 459.35\,/\,1088.36 (42.21\%) & 820.2\,/\,2285.32 (35.89\%) & 1163.8\,/\,3843.18 (30.28\%) & 1401.9\,/\,5169.09 (27.12\%) \\
 &  & \textsc{var} & 175.45\,/\,471.55 (37.21\%) & 322.91\,/\,923.86 (34.95\%) & 669.68\,/\,2161.73 (30.98\%) & 1011.09\,/\,3768.86 (26.83\%) & 1238.5\,/\,5106.95 (24.25\%) \\
 &  & \textsc{mean}+\textsc{var} & 305.85\,/\,704.86 (43.39\%) & 523.45\,/\,1321.68 (39.6\%) & 877.35\,/\,2569.23 (34.15\%) & 1192.6\,/\,4049.45 (29.45\%) & 1423.55\,/\,5352.45 (26.6\%) \\
 &  &  &  &  &  &  &  \\
 & \multirow{3}{*}{0.4} & \textsc{mean} & 257.70\,/\,560.09 (46.01\%) & 457.50\,/\,1073.45 (42.62\%) & 814.35\,/\,2340.55 (34.79\%) & 1131.65\,/\,3974.82 (28.47\%) & 1336.05\,/\,5335.91 (25.04\%) \\
 &  & \textsc{var} & 129.27\,/\,348.95 (37.05\%) & 276.09\,/\,795.59 (34.7\%) & 619.86\,/\,2088.91 (29.67\%) & 949.59\,/\,3784.23 (25.09\%) & 1165.23\,/\,5216 (22.34\%) \\
 &  & \textsc{mean}+\textsc{var} & 300.55\,/\,690.95 (43.5\%) & 511.70\,/\,1282.36 (39.9\%) & 867.00\,/\,2657.14 (32.63\%) & 1169.55\,/\,4292.95 (27.24\%) & 1361.6\,/\,5589.18 (24.36\%) \\
 &  &  &  &  &  &  &  \\
 & \multirow{3}{*}{0.6} & \textsc{mean} & 182.05\,/\,404.36 (45.13\%) & 337.18\,/\,779.09 (43.36\%) & 757.50\,/\,2101.32 (36.05\%) & 1119.10\,/\,3889.55 (28.77\%) & 1332.40\,/\,5316.82 (25.06\%) \\
 &  & \textsc{var} & 67.41\,/\,193.55 (34.83\%) & 165.50\,/\,489.64 (33.8\%) & 490.36\,/\,1705.36 (28.75\%) & 864.18\,/\,3510.23 (24.62\%) & 1112.32\,/\,5014.59 (22.18\%) \\
 &  & \textsc{mean}+\textsc{var} & 216.00\,/\,507.82 (42.53\%) & 391.30\,/\,967.32 (40.45\%) & 819.03\,/\,2490.36 (32.9\%) & 1180.25\,/\,4405.14 (26.79\%) & 1387.55\,/\,5840.77 (23.76\%) \\
 &  &  &  &  &  &  &  \\
 & \multirow{3}{*}{0.8} & \textsc{mean} & 184.75\,/\,455.68 (40.54\%) & 313.40\,/\,828.64 (37.82\%) & 601.00\,/\,1838.32 (32.69\%) & 912.45\,/\,3304.68 (27.61\%) & 1108.36\,/\,4570.09 (24.26\%) \\
 &  & \textsc{var} & 73.86\,/\,252.50 (29.25\%) & 156.45\,/\,557.55 (28.06\%) & 367.27\,/\,1460.82 (25.14\%) & 600.95\,/\,2802.82 (21.44\%) & 790.00\,/\,4085.05 (19.34\%) \\
 &  & \textsc{mean}+\textsc{var} & 215.75\,/\,588.45 (36.66\%) & 361.85\,/\,1053.82 (34.34\%) & 653.85\,/\,2184.50 (29.93\%) & 975.40\,/\,3932 (24.81\%) & 1161.70\,/\,5303.55 (21.9\%) \\
 &  &  &  &  &  &  &  \\
 & \multirow{3}{*}{1.0} & \textsc{mean} & 139.35\,/\,297.05 (46.91\%) & 257.12\,/\,582.59 (44.11\%) & 594.50\,/\,1572.32 (37.78\%) & 1052.10\,/\,3648.09 (28.84\%) & 1293.40\,/\,5302.5 (24.39\%) \\
 &  & \textsc{var} & 38.36\,/\,123.68 (31.02\%) & 89.09\,/\,290.91 (30.63\%) & 306.50\,/\,1159.18 (26.44\%) & 758.68\,/\,3215.27 (23.6\%) & 1064.59\,/\,5042.23 (21.11\%) \\
 &  & \textsc{mean}+\textsc{var} & 163.85\,/\,377.59 (43.39\%) & 301.38\,/\,743.32 (40.6\%) & 665.50\,/\,1993.09 (33.39\%) & 1111.80\,/\,4165.68 (26.69\%) & 1354.75\,/\,5911.32 (22.92\%) \\
 &  &  &  &  &  &  &  \\ \hline
\multirow{21}{*}{\textbf{\begin{tabular}[c]{@{}c@{}}MC\\ Dropout\end{tabular}}} &  &  &  &  &  &  &  \\
 & \multirow{3}{*}{0.2} & \textsc{mean} & 156.00\,/\,672.00 (23.21\%) & 259.00\,/\,1196 00(21.66\%) & 468.50\,/\,2404.00 (19.49\%) & 697.25\,/\,4143.50 (16.83\%) & 864.75\,/\,5682.00 (15.22\%) \\
 &  & \textsc{var} & 47.00\,/\,342.00 (13.74\%) & 68.50\,/\,589.00 (11.63\%) & 102.50\,/\,1284.50 (7.98\%) & 118.50\,/\,1833.50 (6.46\%) & 154.05\,/\,2332.98 (6.60\%) \\
 &  & \textsc{mean}+\textsc{var} & 198.75\,/\,996.00 (19.95\%) & 318.75\,/\,1737.00 (18.35\%) & 546.75\,/\,3553.00 (15.39\%) & 780.75\,/\,5763.00 (13.55\%) & 938.50\,/\,7220.08 (13.00\%) \\
 &  &  &  &  &  &  &  \\
 & \multirow{3}{*}{0.4} & \textsc{mean} & 127.75\,/\,442.00 (28.9\%) & 278.00\,/\,1013.50 (27.43\%) & 588.50\,/\,2593.00 (22.7\%) & 840.75\,/\,4461.00 (18.85\%) & 1015.53\,/\,6070.00 (16.73\%) \\
 &  & \textsc{var} & 26.00\,/\,128.00 (20.31\%) & 175.50\,/\,817.00 (21.48\%) & 234.05\,/\,1578.00 (14.86\%) & 302.75\,/\,2856.50 (10.60\%) & 336.50\,/\,3998.00 (8.42\%) \\
 &  & \textsc{mean}+\textsc{var} & 148.75\,/\,543.58 (27.37\%) & 369.5\,/\,1529.00 (24.17\%) & 687.75\,/\,3641.00 (18.89\%) & 975.00\,/\,6531.00 (14.93\%) & 1164.75\,/\,9097.50 (12.8\%) \\
 &  &  &  &  &  &  &  \\
 & \multirow{3}{*}{0.6} & \textsc{mean} & 99.25\,/\,368.00 (26.97\%) & 182.50\,/\,709.00 (25.74\%) & 386.50\,/\,1718.00 (22.50\%) & 651.50\,/\,3326.50 (19.59\%) & 846.00\,/\,4824.50 (17.54\%) \\
 &  & \textsc{var} & 80.00\,/\,310.00 (25.81\%) & 110.25\,/\,480.50 (22.94\%) & 213.00\,/\,1357.00 (15.7\%) & 293.50\,/\,2597.50 (11.3\%) & 366.00\,/\,4017.50 (9.11\%) \\
 &  & \textsc{mean}+\textsc{var} & 108.54\,/\,421.66 (26.97\%) & 213.75\,/\,877.50 (24.36\%) & 489.25\,/\,2598.50 (18.83\%) & 823.5\,/\,5382.00 (15.30\%) & 1060.25\,/\,8051.00 (13.17\%) \\
 &  &  &  &  &  &  &  \\
 & \multirow{3}{*}{0.8} & \textsc{mean} & 80.25\,/\,275.00 (29.18\%) & 166.25\,/\,580.50 (28.64\%) & 420.50\,/\,1639.50 (25.65\%) & 758.25\,/\,3742.00 (20.26\%) & 967.75\,/\,5573.00 (17.36\%) \\
 &  & \textsc{var} & 20.00\,/\,115.40 (17.33\%) & 102.50\,/\,360.50 (28.43\%) & 238.75\,/\,1286.50 (18.56\%) & 321.75\,/\,2755.00 (11.68\%) & 353.75\,/\,3894.50 (9.08\%) \\
 &  & \textsc{mean}+\textsc{var} & 89.14\,/\,317.89 (29.18\%) & 225.00\,/\,815.50 (27.59\%) & 496.00\,/\,2295.00 (21.61\%) & 867.75\,/\,5566.00 (15.59\%) & 1096.25\,/\,8399.50 (13.05\%) \\
 &  &  &  &  &  &  &  \\
 & \multirow{3}{*}{1.0} & \textsc{mean} & 97.50\,/\,356.50 (27.35\%) & 183.50\,/\,671.00 (27.35\%) & 390.75\,/\,1664.00 (23.48\%) & 649.25\,/\,3252.00 (19.96\%) & 846.41\,/\,4830.50 (17.51\%) \\
 &  & \textsc{var} & 51.75\,/\,211.00 (24.53\%) & 95.00\,/\,509.50 (18.65\%) & 164.00\,/\,1345.50 (12.19\%) & 212.00\,/\,2514.50 (8.43\%) & 242.25\,/\,3620.00 (6.69\%) \\
 &  & \textsc{mean}+\textsc{var} & 130.50\,/\,499.00 (26.15\%) & 248.75\,/\,1070.50 (23.24\%) & 507.75\,/\,2828.50 (17.95\%) & 778.25\,/\,5352.5 (14.54\%) & 984.50\,/\,7836.00 (12.56\%) \\
 & \multicolumn{1}{l}{} & \multicolumn{1}{l}{} &  &  &  &  &  \\ \hline
\multicolumn{1}{l}{} & \multicolumn{1}{l}{} & \multicolumn{1}{l}{} & \multicolumn{1}{l}{} & \multicolumn{1}{l}{} & \multicolumn{1}{l}{} & \multicolumn{1}{l}{} & \multicolumn{1}{l}{} \\
\multirow{19}{*}{\textbf{TTA}} & \multirow{3}{*}{0.2} & \textsc{mean} & 160.73\,/\,442.53 (36.32\%) & 300.73\,/\,867.40 (34.67\%) & 628.15\,/\,2043.80 (30.73\%) & 1001.05\,/\,3798.60 (26.35\%) & 1268.52\,/\,5405.95 (23.47\%) \\
 &  & \textsc{var} & 101.92\,/\,439.09 (23.21\%) & 188.45\,/\,871.03 (21.64\%) & 384.03\,/\,2053.84 (18.7\%) & 603.42\,/\,3807.60 (15.85\%) & 761.99\,/\,5403.30 (14.10\%) \\
 &  & \textsc{mean}+\textsc{var} & 225.89\,/\,615.50 (36.7\%) & 381.36\,/\,1105.61 (34.49\%) & 712.02\,/\,2366.18 (30.09\%) & 1072.93\,/\,4171.33 (25.72\%) & 1330.81\,/\,5796.02 (22.96\%) \\
 &  &  &  &  &  &  &  \\
 & \multirow{3}{*}{0.4} & \textsc{mean} & 145.64\,/\,446.73 (32.60\%) & 270.80\,/\,863.79 (31.35\%) & 556.17\,/\,2005.39 (27.73\%) & 865.48\,/\,3684.75 (23.49\%) & 1076.09\,/\,5198.03 (20.70\%) \\
 &  & \textsc{var} & 98.44\,/\,405.14 (24.30\%) & 183.54\,/\,800.00 (22.94\%) & 390.42\,/\,1915.01 (20.38\%) & 624.40\,/\,3583.47 (17.42\%) & 785.10\,/\,5113.27 (15.35\%) \\
 &  & \textsc{mean}+\textsc{var} & 216.05\,/\,672.65 (32.12\%) & 357.82\,/\,1174.99 (30.45\%) & 635.78\,/\,2379.68 (26.72\%) & 927.29\,/\,4097.41 (22.63\%) & 1127.04\,/\,5660.57 (19.91\%) \\
 &  &  &  &  &  &  &  \\
 & \multirow{3}{*}{0.6} & \textsc{mean} & 136.05\,/\,400.18 (34\%) & 254.99\,/\,790.19 (32.27\%) & 541.09\,/\,1888.11 (28.66\%) & 880.71\,/\,3601.50 (24.45\%) & 1123.41\,/\,5190.18 (21.64\%) \\
 &  & \textsc{var} & 94.30\,/\,402.53(23.43\%) & 173.96\,/\,776.89(22.39\%) & 374.77\,/\,1863.93(20.11\%) & 630.69\,/\,3574.91(17.64\%) & 828.19\,/\,5161.17(16.05\%) \\
 &  & \textsc{mean}+\textsc{var} & 204.16\,/\,649.81 (31.42\%) & 348.87\,/\,1168.24 (29.86\%) & 661.88\,/\,2488.02 (26.6\%) & 987.76\,/\,4306.54 (22.94\%) & 1208.71\,/\,5883.95 (20.54\%) \\
 &  &  &  &  &  &  &  \\
 & \multirow{3}{*}{0.8} & \textsc{mean} & 125.70\,/\,340.94 (36.87\%) & 237.19\,/\,671.14 (35.34\%) & 518.29\,/\,1631.14 (31.77\%) & 871.00\,/\,3115.51 (27.96\%) & 1149.79\,/\,4553.01 (25.25\%) \\
 &  & \textsc{var} & 55.64\,/\,282.56 (19.69\%) & 112.50\,/\,563.72 (19.96\%) & 281.28\,/\,1396.29 (20.14\%) & 562.35\,/\,2789.61 (20.16\%) & 852.66\,/\,4199.69 (20.30\%) \\
 &  & \textsc{mean}+\textsc{var} & 174.40\,/\,583.68 (29.88\%) & 323.77\,/\,1114.29 (29.06\%) & 654.62\,/\,2433.87 (26.90\%) & 1003.66\,/\,4082.39 (24.59\%) & 1264.61\,/\,5528.58 (22.87\%) \\
 &  &  &  &  &  &  &  \\
 & \multirow{3}{*}{1.0} & \textsc{mean} & 112.15\,/\,331.71 (33.81\%) & 214.10\,/\,655.97 (32.64\%) & 473.00\,/\,1595.31 (29.65\%) & 806.60\,/\,3091.35 (26.09\%) & 1076.09\,/\,4534.76 (23.73\%) \\
 &  & \textsc{var} & 57.26\,/\,300.19 (19.08\%) & 110.43\,/\,573.19 (19.27\%) & 265.93\,/\,1393.10 (19.09\%) & 512.16\,/\,2784.15 (18.4\%) & 750.04\,/\,4197.76 (17.87\%) \\
 &  & \textsc{mean}+\textsc{var} & 160.47\,/\,566.43 (28.33\%) & 287.13\,/\,1037.14 (27.68\%) & 572.94\,/\,2213.89 (25.88\%) & 908.01\,/\,3817.26 (23.79\%) & 1177.10\,/\,5311.63 (22.16\%) \\ 
 &  &  &  &  &  &  &  \\ \hline
\end{tabular}%
}
\end{table}

\begin{table}[p]
\centering
\caption{Performance in terms of \ac{FNP} numbers for the Messidor-2 dataset. The fraction in each entry shows the number of false negatives (the numerator) and the number of uncertain negatives (the denominator). The percentage numbers in the parentheses are the corresponding false negative precision values.}
\label{tab:fnp-messidor}
\resizebox{\textwidth}{!}{%
\begin{tabular}{cccccccc}
\hline
\textbf{\begin{tabular}[c]{@{}c@{}}Ensemble\\ Method\end{tabular}} & \textbf{$\rho$} & \textbf{\begin{tabular}[c]{@{}c@{}}Uncertainty\\ Metric\end{tabular}} & \textbf{$q=1$} & \textbf{$q=2$} & \textbf{$q=5$} & \textbf{$q=10$} & \textbf{$q=15$} \\ \hline
\multirow{21}{*}{\textbf{\begin{tabular}[c]{@{}c@{}}Stacking \\ Ensemble\end{tabular}}} &  &  &  &  &  &  &  \\
 & \multirow{3}{*}{0.2} & \textsc{mean} & 34.05\,/\,46.67 (72.96\%) & 34.25\,/\,48.24 (71\%) & 68.45\,/\,103.77 (65.96\%) & 101.61\,/\,172.67 (58.85\%) & 121.06\,/\,227.24 (53.28\%) \\
 &  & \textsc{var} & 29.17\,/\,40.49 (72.06\%) & 26.27\,/\,37.87 (69.37\%) & 49.01\,/\,75.83 (64.64\%) & 72.36\,/\,125.09 (57.85\%) & 87.29\,/\,164.57 (53.04\%) \\
 &  & \textsc{mean}+\textsc{var} & 37.06\,/\,51.65 (71.75\%) & 47.59\,/\,68.51 (69.47\%) & 80.77\,/\,126.23 (63.99\%) & 108.63\,/\,191.37 (56.76\%) & 125.1\,/\,242.12 (51.67\%) \\
 &  &  &  &  &  &  &  \\
 & \multirow{3}{*}{0.4} & \textsc{mean} & 22.48\,/\,32.66 (68.83\%) & 37.4\,/\,54.7 (68.37\%) & 70.75\,/\,113.8 (62.17\%) & 97.45\,/\,180.65 (53.94\%) & 112.8\,/\,228.75 (49.31\%) \\
 &  & \textsc{var} & 20.14\,/\,30.63 (65.74\%) & 33.3\,/\,51.75 (64.35\%) & 51.95\,/\,85.10 (60.62\%) & 89\,/\,166.25 (53.53\%) & 105.05\,/\,213.45 (49.22\%) \\
 &  & \textsc{mean}+\textsc{var} & 22.88\,/\,34.86 (65.82\%) & 51.85\,/\,79.8 (64.97\%) & 80.55\,/\,134 (60.11\%) & 101.65\,/\,194.05 (52.38\%) & 114.7\,/\,238.8 (48.03\%) \\
 &  &  &  &  &  &  &  \\
  & \multirow{3}{*}{0.6} & \textsc{mean} & 20.5\,/\,26.26 (78.06\%) & 28.7\,/\,37.9 (76.56\%) & 63.9\,/\,86.9 (73.53\%) & 90.75\,/\,135.10(67.17\%) & 104.75\,/\,173.26 (60.46\%) \\
 &  & \textsc{var} & 20.14\,/\,26.31 (76.54\%) & 28.5\,/\,37.87 (74.25\%) & 58.1\,/\,84.17 (69.02\%) & 85.1\,/\,135.92 (62.61\%) & 140.45\,/\,173.81 (57.79\%) \\
 &  & \textsc{mean}+\textsc{var} & 24.19\,/\,31.54 (76.69\%) & 34.8\,/\,46.26 (75.22\%) & 74.35\,/\,105.76 (70.3\%) & 100.3\,/\,158.68 (63.21\%) & 142.65\,/\,252.3 (56.54\%) \\
 &  &  &  &  &  &  &  \\
 & \multirow{3}{*}{0.8} & \textsc{mean} & 17.1\,/\,22.1 (77.37\%) & 27.35\,/\,37.51 (72.95\%) & 60.15\,/\,88.23 (68.17\%) & 100.65\,/\,170.5 (59.03\%) & 117.35\,/\,222.05 (52.85\%) \\
 &  & \textsc{var} & 14.77\,/\,19.05 (77.55\%) & 25.15\,/\,34.45 (73\%) & 53.1\,/\,78.39 (67.74\%) & 96.25\,/\,161.6 (59.56\%) & 118.05\,/\,221.45 (53.31\%) \\
 &  & \textsc{mean}+\textsc{var} & 18.02\,/\,23.72 (75.97\%) & 30.15\,/\,42.38 (71.15\%) & 63.5\,/\,96.7 (65.67\%) & 108.45\,/\,190.05 (57.06\%) & 123.9\,/\,242.15 (51.17\%) \\
 &  &  &  &  &  &  &  \\
 & \multirow{3}{*}{1.0} & \textsc{mean} & 17.52\,/\,22.02 (79.55\%) & 35.4\,/\,45.97 (77.01\%) & 88.1\,/\,125.05 (70.45\%) & 123.75\,/\,194.8 (63.53\%) & 93.7\,/\,252.05 (57.01\%) \\
 &  & \textsc{var} & 17.52\,/\,21.34 (76.27\%) & 30.75\,/\,42.36 (72.6\%) & 86.8\,/\,131.2 (66.16\%) & 121.3\,/\,201.2 (60.29\%) & 140.8\,/\,253.4 (55.56\%) \\
 &  & \textsc{mean}+\textsc{var} & 20.27\,/\,22.66 (76.22\%) & 40.9\,/\,56.15 (72.88\%) & 112.45\,/\,170.7 (65.88\%) & 135.95\,/\,226.95 (59.9\%) & 149.6\,/\,274.45 (54.51\%) \\
 & \multicolumn{1}{l}{} & \multicolumn{1}{l}{} &  &  &  &  &  \\ \hline
\multicolumn{1}{l}{\multirow{21}{*}{\textbf{MC-dropout}}} & \multicolumn{1}{l}{} & \multicolumn{1}{l}{} & \multicolumn{1}{l}{} & \multicolumn{1}{l}{} & \multicolumn{1}{l}{} & \multicolumn{1}{l}{} & \multicolumn{1}{l}{} \\
\multicolumn{1}{l}{} & \multirow{3}{*}{0.2} & \textsc{mean} & 8.89\,/\,13.24 (67.15\%) & 17.14\,/\,26.23 (65.35\%) & 39.12\,/\,67.2 (58.21\%) & 62.22\,/\,128.38 (48.47\%) & 82.14\,/\,178.22 (46.09\%) \\
\multicolumn{1}{l}{} &  & \textsc{var} & 10.07\,/\,16.22 (62.05\%) & 14.18\,/\,30.34 (46.73\%) & 20.11\,/\,59.2( 33.98\%) & 23.15\,/\,93.23 (24.83\%) & 26.2\,/\,120.28 (21.78\%) \\
\multicolumn{1}{l}{} &  & \textsc{mean}+\textsc{var} & 13.19\,/\,25.24 (52.25\%) & 26.18\,/\,50.3 (52.05\%) & 50.22\,/\,111.3 (45.12\%) & 73.12\,/\,197.23 (37.07\%) & 92.21\,/\,263.28 (35.02\%) \\
\multicolumn{1}{l}{} &  &  &  &  &  &  &  \\
\multicolumn{1}{l}{} & \multirow{3}{*}{0.4} & \textsc{mean} & 15.66\,/\,32.24 (48.58\%) & 33.19\,/\,54.33 (61.08\%) & 57.17\,/\,97.32 (58.74\%) & 78.13\,/\,152.21 (51.33\%) & 89.15\,/\,200.23 (44.52\%) \\
\multicolumn{1}{l}{} &  & \textsc{var} & 5.17\,/\,7.22 (41.66\%) & 5.21\,/\,7.29 (71.52\%) & 29.05\,/\,54.19 (53.6\%) & 30.16\,/\,82.29 (36.65\%) & 34.17\,/\,109.19 (31.3\%) \\
\multicolumn{1}{l}{} &  & \textsc{mean}+\textsc{var} & 16.38\,/\,32.26 (50.79\%) & 33.71\,/\,55.16 (61.12\%) & 61.19\,/\,110.31 (55.47\%) & 82.18\,/\,191.24 (42.97\%) & 95.15\,/\,258.3 (36.83\%) \\
\multicolumn{1}{l}{} &  &  &  &  &  &  &  \\
\multicolumn{1}{l}{} & \multirow{3}{*}{0.6} & \textsc{mean} & 15.04\,/\,22.15 (67.88\%) & 23.07\,/\,43.16 (53.44\%) & 27.19\,/\,53.74(50.6\%) & 33.13\,/\,48.44 (45.69\%) & 56.21\,/\,134.44 (41.81\%) \\
\multicolumn{1}{l}{} &  & \textsc{var} & 1.18\,/\,1.77 (66.69\%) & 4.09\,/\,8.17 (50.08\%) & 13.12\,/\,46.22 (28.38\%) & 14.18\,/\,49.5 (28.59\%) & 24.9\,/\,97.25 (25.6\%) \\
\multicolumn{1}{l}{} &  & \textsc{mean}+\textsc{var} & 15.1\,/\,22.17  (68.15\%) & 24.14\,/\,45.25 (53.34\%) & 30.23\,/\,71.79 (42.11\%) & 37.18\,/\,83.38 (44.59\%) & 66.41\,/\,207.33 (32.03\%) \\
\multicolumn{1}{l}{} &  &  &  &  &  &  &  \\
\multicolumn{1}{l}{} & \multirow{3}{*}{0.8} & \textsc{mean} & 7.16\,/\,13.28 (53.92\%) & 9.13\,/\,16.21 (56.3\%) & 18.06\,/\,35.16 (51.36\%) & 33.25\,/\,69.28 (47.99\%) & 64.18\,/\,130.34 (49.24\%) \\
\multicolumn{1}{l}{} &  & \textsc{var} & 4.25\,/\,6.9 (38.76\%) & 6.18\,/\,9.25 (66.76\%) & 18.13\,/\,40.28 (45\%) & 37.24\,/\,93.3 (39.92\%) & 50.16\,/\,126.22 (39.74\%) \\
\multicolumn{1}{l}{} &  & \textsc{mean}+\textsc{var} & 7.2\,/\,13.34 (53.96\%) & 15.15\,/\,25.28 (59.93\%) & 36.19\,/\,75.24 (48.1\%) & 68.26\,/\,158.38 (43.1\%) & 108.22\,/\,247.38 (43.75\%) \\
\multicolumn{1}{l}{} &  &  &  &  &  &  &  \\
\multicolumn{1}{l}{} & \multirow{3}{*}{1.0} & \textsc{mean} & 5.14\,/\,12.18 (42.19\%) & 6.11\,/\,17.15 (35.61\%) & 37.16\,/\,75.2 (49.42\%) & 58.17\,/\,128.22 (45.37\%) & 69.16\,/\,168.22 (41.11\%) \\
\multicolumn{1}{l}{} &  & \textsc{var} & 4.15\,/\,12.26 (33.85\%) & 4.25\,/\,16.32 (26.03\%) & 6.15\,/\,26.22 (23.47\%) & 7.13\,/\,42.23 (16.88\%) & 9.26\,/\,73.32 (12.63\%) \\
\multicolumn{1}{l}{} &  & \textsc{mean}+\textsc{var} & 8.09\,/\,21.19 (38.16\%) & 9.2\,/\,30.34 (30.33\%) & 39.13\,/\,90.2 (43.38\%) & 61.15\,/\,156.3 (39.13\%) & 73.17\,/\,223.28 (32.77\%) \\
\multicolumn{1}{l}{} &  &  &  &  &  &  &  \\ \hline
\multirow{20}{*}{\textbf{TTA}} &  &  &  &  &  &  &  \\
 & \multirow{3}{*}{0.2} & \textsc{mean} & 15.42\,/\,25.84 (59.6\%) & 24.28\,/\,41.65 (58.29\%) & 49.56\,/\,93.46 (53.03\%) & 75.78\,/\,163.84 (46.25\%) & 92.52\,/\,223.52 (41.39\%) \\
 &  & \textsc{var} & 14.27\,/\,23.79 (59.99\%) & 23.43\,/\,40.3 (58.13\%) & 48.1\,/\,91.33 (52.66\%) & 75.23\,/\,163.15 (46.11\%) & 92.13\,/\,223.53 (41.21\%) \\
 &  & \textsc{mean}+\textsc{var} & 16.79\,/\,28.33 (59.26\%) & 26\,/\,45.05 (57.72\%) & 50.36\,/\,96.42 (52.23\%) & 77.63\,/\,169.9 (45.69\%) & 94.22\,/\,231.09 (40.77\%) \\
 &  &  &  &  &  &  &  \\
 & \multirow{3}{*}{0.4} & \textsc{mean} & 12.27\,/\,22.39 (54.8\%) & 20.26\,/\,37.93 (53.42\%) & 42.14\,/\,82.11 (51.32\%) & 73.4\,/\,153.43 (47.84\%) & 93.67\,/\,212.11 (44.16\%) \\
 &  & \textsc{var} & 7.02\,/\,14.36 (48.87\%) & 11.47\,/\,24.61 (46.61\%) & 25.91\,/\,60.49 (42.84\%) & 48.88\,/\,122.81 (39.8\%) & 70.06\,/\,182.98 (38.29\%) \\
 &  & \textsc{mean}+\textsc{var} & 17.27\,/\,33.19 (52.04\%) & 29.33\,/\,58.16 (50.42\%) & 61.22\,/\,128.63 (47.6\%) & 89.44\,/\,204.85 (43.66\%) & 101.75\,/\,248.21 (40.99\%) \\
 &  &  &  &  &  &  &  \\
 & \multirow{3}{*}{0.6} & \textsc{mean} & 11.48\,/\,22.35 (51.37\%) & 19.57\,/\,38.61 (50.68\%) & 38.65\,/\,78.44 (49.28\%) & 69.91\,/\,153.38 (45.58\%) & 94.44\,/\,224.03 (42.16\%) \\
 &  & \textsc{var} & 6.95\,/\,13.58 (51.21\%) & 12.68\,/\,25.33 (50.08\%) & 35.43\,/\,71.75 (49.38\%) & 69.64\,/\,152.8 (45.58\%) & 94.13\,/\,224 (42.02\%) \\
 &  & \textsc{mean}+\textsc{var} & 16.51\,/\,32.3 (51.11\%) & 28.31\,/\,56.46 (50.14\%) & 51.71\,/\,105.96 (48.8\%) & 75.71\,/\,168.65 (44.89\%) & 97.85\,/\,235.1 (41.62\%) \\
 &  &  &  &  &  &  &  \\
 & \multirow{3}{*}{0.8} & \textsc{mean} & 10.81\,/\,20.18 (53.57\%) & 17.55\,/\,33.84 (51.86\%) & 33.78\,/\,72.11 (46.84\%) & 55.53\,/\,142.75 (38.9\%) & 69.61\,/\,199.8 (34.84\%) \\
 &  & \textsc{var} & 12.14\,/\,24.09 (50.42\%) & 19.05\,/\,38.48 (49.51\%) & 33.84\,/\,73 (46.35\%) & 54.5\,/\,137.78 (39.56\%) & 68.35\,/\,194.13 (35.21\%) \\
 &  & \textsc{mean}+\textsc{var} & 16.04\,/\,31.03 (51.7\%) & 23.57\,/\,47.18 (49.96\%) & 37.97\,/\,83.46 (45.49\%) & 57.9\,/\,151.68 (38.18\%) & 71.66\,/\,208.88 (34.31\%) \\
 &  &  &  &  &  &  &  \\
 & \multirow{3}{*}{1.0} & \textsc{mean} & 7.02\,/\,13.89 (50.54\%) & 11.82\,/\,22.86 (51.69\%) & 28.39\,/\,57.41 (49.46\%) & 54.8\,/\,122.8 (44.62\%) & 77.41\,/\,183.14(42.27\%) \\
 &  & \textsc{var} & 8.69\,/\,21.26 (40.88\%) & 14.39\,/\,34.35 (41.9\%) & 28.24\,/\,70.05 (40.31\%) & 51.41\,/\,130.38 (39.44\%) & 72.1\,/\,185.7(38.82\%) \\
 &  & \textsc{mean}+\textsc{var} & 14.08\,/\,31 (45.41\%) & 24.58\,/\,53.06 (46.32\%) & 54.95\,/\,123.21 (44.6\%) & 95.78\,/\,230.74 (41.51\%) & 117.39\,/\,292.46(40.14\%) \\
 &  &  &  &  &  &  &  \\ \hline
\end{tabular}%
}
\end{table}

\begin{table}[p]
\centering
\caption{Number of the remaining false negative predictions from uncertainty-informed diagnosis schemes for the Kaggle-DR dataset. The reduction from the baseline (no uncertainty information is exploited) is shown as percentage numbers in the parentheses.}
\label{tab:num-false-negatives-DR}
\resizebox{\textwidth}{!}{%
\begin{tabular}{cccccccc}
\hline
\textbf{\begin{tabular}[c]{@{}c@{}}Ensemble\\ Method\end{tabular}} & \textbf{$\rho$} & \textbf{\begin{tabular}[c]{@{}c@{}}Uncertainty\\ Metric\end{tabular}} & \textbf{$q=1$} & \textbf{$q=2$} & \textbf{$q=5$} & \textbf{$q=10$} & \textbf{$q=15$} \\ \hline
\multirow{21}{*}{\textbf{\begin{tabular}[c]{@{}c@{}}Stacking\\ Ensemble\end{tabular}}} &  &  &  &  &  &  &  \\
 & \multirow{3}{*}{0.2} & \textsc{mean} & 2449.65 (-9.43\%) & 2265 (-16.26\%) & 1918.15 (-29.08\%) & 1586.8 (-41.33\%) & 1359.2 (-49.75\%) \\
 &  & \textsc{var} & 2519.95 (-6.83\%) & 2363.6 (-12.61\%) & 1997.65 (-26.14\%) & 1636.15 (-39.51\%) & 1394.1 (-48.46\%) \\
 &  & \textsc{mean}+\textsc{var} & 2410.05 (-10.89\%) & 2201.8 (-18.59\%) & 1860.95 (-31.2\%) & 1558.85 (-42.37\%) & 1338.7 (-50.5\%) \\
 &  &  &  &  &  &  &  \\
 & \multirow{3}{*}{0.4} & \textsc{mean} & 2120.85 (-10.42\%) & 1929.25 (-18.51\%) & 1587.1 (-32.97\%) & 1281.25 (-45.88\%) & 1086.95 (-54.09\%) \\
 &  & \textsc{var} & 2230.95 (-5.77\%) & 2076.65 (-12.29\%) & 1713.45 (-27.63\%) & 1365.75 (-42.32\%) & 1134.75 (-52.07\%) \\
 &  & \textsc{mean}+\textsc{var} & 2079.25 (-12.18\%) & 1875.35 (-20.79\%) & 1533.6 (-35.23\%) & 1244.5 (-47.44\%) & 1057.85 (-55.32\%) \\
 &  &  &  &  &  &  &  \\
 & \multirow{3}{*}{0.6} & \textsc{mean} & 2217.6 (-7.27\%) & 2068.15 (-13.52\%) & 1663.85 (-30.43\%) & 1316.5 (-44.95\%) & 1113.05 (-53.46\%) \\
 &  & \textsc{var} & 2320.5 (-2.97\%) & 2215.9 (-7.34\%) & 1874.6 (-21.61\%) & 1477.1 (-38.24\%) & 1215.15 (-49.19\%) \\
 &  & \textsc{mean}+\textsc{var} & 2183.55 (-8.7\%) & 2014.7 (-15.76\%) & 1605.05 (-32.89\%) & 1259.6 (-47.33\%) & 1060.5 (-55.66\%) \\
 &  &  &  &  &  &  &  \\
 & \multirow{3}{*}{0.8} & \textsc{mean} & 2074.05 (-7.84\%) & 1949.45 (-13.38\%) & 1674.25 (-25.61\%) & 1373.55 (-38.97\%) & 1188.55 (-47.19\%) \\
 &  & \textsc{var} & 2172.35 (-3.47\%) & 2085.7 (-7.32\%) & 1864.25 (-17.16\%) & 1615.75 (-28.21\%) & 1412.45 (-37.24\%) \\
 &  & \textsc{mean}+\textsc{var} & 2042.7 (-9.24\%) & 1904.35 (-15.38\%) & 1626.45 (-27.73\%) & 1316.6 (-41.5\%) & 1132.65 (-49.67\%) \\
 &  &  &  &  &  &  &  \\
 & \multirow{3}{*}{1.0} & \textsc{mean} & 2066.75 (-6.06\%) & 1952.4 (-11.26\%) & 1629.55 (-25.93\%) & 1189.95 (-45.91\%) & 958.75 (-56.42\%) \\
 &  & \textsc{var} & 2159.55 (-1.84\%) & 2106.7 (-4.25\%) & 1875.7 (-14.74\%) & 1398.5 (-36.43\%) & 1076.2 (-51.08\%) \\
 &  & \textsc{mean}+\textsc{var} & 2042.2 (-7.18\%) & 1909.2 (-13.22\%) & 1561.6 (-29.02\%) & 1129.95 (-48.64\%) & 896.45 (-59.25\%) \\
 &  &  &  &  &  &  &  \\ \hline
\multirow{21}{*}{\textbf{\begin{tabular}[c]{@{}c@{}}MC\\ Dropout\end{tabular}}} &  &  &  &  &  &  &  \\
 & \multirow{3}{*}{0.2} & \textsc{mean} & 4496.1 (-0.39\%) & 4486.9 (-0.59\%) & 4454.9 (-1.3\%) & 4413.1 (-2.22\%) & 4374.3 (-3.08\%) \\
 &  & \textsc{var} & 4509.5 (-0.09\%) & 4507.7 (-0.13\%) & 4498.5 (-0.33\%) & 4498.3 (-0.34\%) & 4499.5 (-0.31\%) \\
 &  & \textsc{mean}+\textsc{var} & 4493.7 (-0.44\%) & 4482.7 (-0.68\%) & 4443.7 (-1.55\%) & 4399.1 (-2.53\%) & 4364.5 (-3.3\%) \\
 &  &  &  &  &  &  &  \\
 & \multirow{3}{*}{0.4} & \textsc{mean} & 4035.2 (-0.27\%) & 4026 (-0.49\%) & 3993.2 (-1.3\%) & 3980.2 (-1.63\%) & 3965.5 (-1.99\%) \\
 &  & \textsc{var} & 4043.9 (-0.05\%) & 4041.5 (-0.11\%) & 4036.8 (-0.23\%) & 4039.3 (-0.17\%) & 4038.5 (-0.19\%) \\
 &  & \textsc{mean}+\textsc{var} & 4033.7 (-0.3\%) & 4023.9 (-0.55\%) & 3989.5 (-1.4\%) & 3972.3 (-1.82\%) & 3956 (-2.22\%) \\
 &  &  &  &  &  &  &  \\
 & \multirow{3}{*}{0.6} & \textsc{mean} & 3678.4 (-0.35\%) & 3671.8 (-0.53\%) & 3657 (-0.93\%) & 3640.9 (-1.37\%) & 3622.1 (-1.88\%) \\
 &  & \textsc{var} & 3687.3 (-0.11\%) & 3687 (-0.12\%) & 3685.2 (-0.17\%) & 3682.2 (-0.25\%) & 3683.2 (-0.22\%) \\
 &  & \textsc{mean}+\textsc{var} & 3674.1 (-0.47\%) & 3665.1 (-0.72\%) & 3648 (-1.18\%) & 3632.3 (-1.6\%) & 3615.8 (-2.05\%) \\
 &  &  &  &  &  &  &  \\
 & \multirow{3}{*}{0.8} & \textsc{mean} & 3794.5 (-0.17\%) & 3785.5 (-0.41\%) & 3765 (-0.95\%) & 3741.5 (-1.57\%) & 3721.8 (-2.08\%) \\
 &  & \textsc{var} & 3798.8 (-0.06\%) & 3797.8 (-0.08\%) & 3797.1 (-0.1\%) & 3795.1 (-0.16\%) & 3796 (-0.13\%) \\
 &  & \textsc{mean}+\textsc{var} & 3791.1 (-0.26\%) & 3783.8 (-0.45\%) & 3763.1 (-1\%) & 3733.7 (-1.77\%) & 3725.1 (-2\%) \\
 &  &  &  &  &  &  &  \\
 & \multirow{3}{*}{1.0} & \textsc{mean} & 3517.8 (-0.2\%) & 3511.8 (-0.37\%) & 3490.6 (-0.98\%) & 3469.3 (-1.58\%) & 3455.4 (-1.97\%) \\
 &  & \textsc{var} & 3522.2 (-0.08\%) & 3519.7 (-0.15\%) & 3516.7 (-0.24\%) & 3518.1 (-0.2\%) & 3519.5 (-0.16\%) \\
 &  & \textsc{mean}+\textsc{var} & 3513.2 (-0.33\%) & 3507.4 (-0.5\%) & 3488.1 (-1.05\%) & 3465.3 (-1.69\%) & 3450 (-2.13\%) \\
 &  &  &  &  &  &  &  \\ \hline
\multirow{20}{*}{\textbf{TTA}} &  &  &  &  &  &  &  \\
 & \multirow{3}{*}{0.2} & \textsc{mean} & 3645.64 (-0.54\%) & 3629.61 (-0.98\%) & 3590.94 (-2.03\%) & 3549.78 (-3.15\%) & 3520.88 (-3.94\%) \\
 &  & \textsc{var} & 3655.58 (-0.27\%) & 3647.67 (-0.48\%) & 3629.36 (-0.98\%) & 3607.78 (-1.57\%) & 3593.35 (-1.96\%) \\
 &  & \textsc{mean}+\textsc{var} & 3641.16 (-0.66\%) & 3624.16 (-1.12\%) & 3585.15 (-2.19\%) & 3545.99 (-3.26\%) & 3517.06 (-4.05\%) \\
 &  &  &  & \textbf{} & \textbf{} & \textbf{} & \textbf{} \\
 & \multirow{3}{*}{0.4} & \textsc{mean} & 3618.64 (-0.43\%) & 3606.51 (-0.77\%) & 3574.31 (-1.65\%) & 3539.63 (-2.61\%) & 3511.74 (-3.37\%) \\
 &  & \textsc{var} & 3629.76 (-0.13\%) & 3624.83 (-0.26\%) & 3612.01 (-0.62\%) & 3594.3 (-1.1\%) & 3580.51 (-1.48\%) \\
 &  & \textsc{mean}+\textsc{var} & 3613.65 (-0.57\%) & 3596.8 (-1.03\%) & 3560.31 (-2.04\%) & 3521.91 (-3.09\%) & 3496.29 (-3.8\%) \\
 &  &  &  & \textbf{} & \textbf{} & \textbf{} & \textbf{} \\
 & \multirow{3}{*}{0.6} & \textsc{mean} & 3298.46 (-0.15\%) & 3289.08 (-0.3\%) & 3253.57 (-0.86\%) & 3210.81 (-1.49\%) & 3173.9 (-2.04\%) \\
 &  & \textsc{var} & 3306.92 (-0.02\%) & 3306 (-0.03\%) & 3302 (-0.1\%) & 3289.69 (-0.29\%) & 3268.08 (-0.63\%) \\
 &  & \textsc{mean}+\textsc{var} & 3295.75 (-0.2\%) & 3281.79 (-0.42\%) & 3249.22 (-0.91\%) & 3206.09 (-1.57\%) & 3155.75 (-2.31\%) \\
 &  &  &  & \textbf{} & \textbf{} & \textbf{} & \textbf{} \\
 & \multirow{3}{*}{0.8} & \textsc{mean} & 3058.86 (-0.55\%) & 3044.11 (-1.02\%) & 3011.75 (-2.08\%) & 2981.05 (-3.07\%) & 2954.78 (-3.93\%) \\
 &  & \textsc{var} & 3068.14 (-0.24\%) & 3061.46 (-0.46\%) & 3044.88 (-1\%) & 3027.69 (-1.56\%) & 3015.8 (-1.95\%) \\
 &  & \textsc{mean}+\textsc{var} & 3052.75 (-0.74\%) & 3038.33 (-1.21\%) & 3004.99 (-2.3\%) & 2972.66 (-3.35\%) & 2951.89 (-4.02\%) \\
 &  &  &  & \textbf{} & \textbf{} & \textbf{} & \textbf{} \\
 & \multirow{3}{*}{1.0} & \textsc{mean} & 2777.21 (-0.25\%) & 2709.25 (-1.39\%) & 2719.44 (-1.22\%) & 2663.14 (-2.12\%) & 2631.25 (-2.62\%) \\
 &  & \textsc{var} & 2774.09 (-0.31\%) & 2786.42 (-0.1\%) & 2774.25 (-0.31\%) & 2751.93 (-0.68\%) & 2744.39 (-0.81\%) \\
 &  & \textsc{mean}+\textsc{var} & 2749.96 (-0.71\%) & 2714.22 (-1.3\%) & 2726.37 (-1.1\%) & 2667.96 (-2.05\%) & 2566.55 (-3.6\%) \\ &  &  &  &  &  &  &  \\ \hline
\end{tabular}%
}
\end{table}

\begin{table}[p]
\centering
\caption{Number of the remaining false negative predictions from uncertainty-informed diagnosis schemes for the Messdior-2 dataset. The reduction from the baseline (where no uncertainty information is exploited) is shown as percentage numbers in the parentheses.}
\label{tab:num-false-negatives-messidor}
\resizebox{\textwidth}{!}{%
\begin{tabular}{cccccccc}
\hline
\textbf{\begin{tabular}[c]{@{}c@{}}Ensemble\\ Method\end{tabular}} & \textbf{$\rho$} & \textbf{\begin{tabular}[c]{@{}c@{}}Uncertainty\\ Metric\end{tabular}} & \textbf{$q=1$} & \textbf{$q=2$} & \textbf{$q=5$} & \textbf{$q=10$} & \textbf{$q=15$} \\ \hline
\multirow{21}{*}{\textbf{\begin{tabular}[c]{@{}c@{}}Stacking\\ Ensemble\end{tabular}}} &  &  &  &  &  &  &  \\
 & \multirow{3}{*}{0.2} & \textsc{mean} & 201.95 (-7.93\%) & 192.75 (-12.13\%) & 160.75 (-26.72\%) & 118.95 (-45.77\%) & 80.15 (-63.46\%) \\
 &  & \textsc{var} & 205.58(-7.64\%) & 193.52 (-11.77\%) & 173.39 (-20.95\%) & 138.34(-36.93\%) & 95.68 (-56.38\%) \\
 &  & \textsc{mean}+\textsc{var} & 199.55 (-9.03\%) & 188.55 (-14.04\%) & 149.55 (-31.82\%) & 104.95 (-52.15\%) & 70.35 (-67.93\%) \\
 &  &  & \textbf{} & \textbf{} & \textbf{} & \textbf{} & \textbf{} \\
 & \multirow{3}{*}{0.4} & \textsc{mean} & 164.4 (-4.47\%) & 160.5 (-6.74\%) & 154.5 (-10.23\%) & 139.3 (-19.06\%) & 135.7 (-21.15\%) \\
 &  & \textsc{var} & 164.30 (-4.53\%) & 162.97 (-5.30\%) & 159.6 (-7.25\%) & 146.78 (-14.71\%) & 142.05(-17.46\%) \\
 &  & \textsc{mean}+\textsc{var} & 163.3 (-5.11\%) & 156.9 (-8.83\%) & 142.5 (-17.2\%) & 143.7 (-16.5\%) & 129.3 (-24.87\%) \\
 &  &  &  & \textbf{} & \textbf{} & \textbf{} & \textbf{} \\
 & \multirow{3}{*}{0.6} & \textsc{mean} & 190.84 (-4.13\%) & 188.76 (-5.17\%) & 186.69 (-6.21\%) & 181.18 (-8.98\%) & 1174.97 (-12.28\%) \\
 &  & \textsc{var} & 191.24 (-3.92\%) & 190.44 (-4.32\%) & 188.12 (-5.49\%) & 181.98 (-8.58\%) & 174.79 (-12.1\%) \\
 &  & \textsc{mean}+\textsc{var} & 189.88 (-4.61\%) & 188.12 (-5.49\%) & 185.96 (-6.58\%) & 180.18 (-9.48\%) & 172.23 (-13.48\%) \\
 &  &  &  & \textbf{} & \textbf{} & \textbf{} & \textbf{} \\
 & \multirow{3}{*}{0.8} & \textsc{mean} & 165.1 (-4.67\%) & 164.78 (-4.86\%) & 163.98 (-5.33\%) & 157.78 (-8.9\%) & 152.67 (-11.85\%) \\
 &  & \textsc{var} & 165.25 (-4.59\%) & 165.42 (-4.5\%) & 163.26 (-5.74\%) & 160.17 (-7.52\%) & 153.28 (-11.5\%) \\
 &  & \textsc{mean}+\textsc{var} & 164.78 (-4.86\%) & 163.34 (-5.7\%) & 160.46 (-7.36\%) & 150.68 (-13\%) & 141.8 (-18.13\%) \\
 &  &  &  & \textbf{} & \textbf{} & \textbf{} & \textbf{} \\
 & \multirow{3}{*}{1.0} & \textsc{mean} & 164.45 (-1.62\%) & 161.15 (-3.59\%) & 157.7 (-5.65\%) & 156.35 (-6.46\%) & 152.45 (-8.79\%) \\
 &  & \textsc{var} & 165.25 (-1.14\%) & 163.85 (-1.97\%) & 161.25 (-3.53\%) & 158.95 (-4.91\%) & 159.15 (-4.79\%) \\
 &  & \textsc{mean}+\textsc{var} & 164.1 (-1.88\%) & 159.2 (-4.76\%) & 152.15 (-8.97\%) & 151.55 (-9.33\%) & 147.5 (-11.76\%) \\
 &  &  &  &  &  &  &  \\ \hline
\multirow{20}{*}{\textbf{\begin{tabular}[c]{@{}c@{}}MC\\ Dropout\end{tabular}}} &  &  &  &  &  &  &  \\
 & \multirow{3}{*}{0.2} & \textsc{mean} & 316.39 (-0.41\%) & 315.73 (-0.62\%) & 314.25 (-1.09\%) & 312.64 (-1.59\%) & 310.76 (-2.18\%) \\
 &  & \textsc{var} & 317.28 (-0.13\%) & 317.25 (-0.14\%) & 317.07 (-0.2\%) & 316.77 (-0.29\%) & 316.87 (-0.26\%) \\
 &  & \textsc{mean}+\textsc{var} & 315.96 (-0.55\%) & 315.06 (-0.83\%) & 313.35 (-1.37\%) & 311.78 (-1.86\%) & 310.13 (-2.38\%) \\
 &  &  & \textbf{} &  &  &  &  \\
 & \multirow{3}{*}{0.4} & \textsc{mean} & 294.32 (-0.37\%) & 293.4 (-0.68\%) & 290.12 (-1.79\%) & 288.82 (-2.23\%) & 287.35 (-2.73\%) \\
 &  & \textsc{var} & 295.19 (-0.07\%) & 294.95 (-0.15\%) & 294.48 (-0.31\%) & 294.73 (-0.23\%) & 294.65 (-0.25\%) \\
 &  & \textsc{mean}+\textsc{var} & 294.17 (-0.42\%) & 293.19 (-0.75\%) & 289.75 (-1.91\%) & 288.03 (-2.49\%) & 286.4 (-3.05\%) \\
 &  &  &  &  &  &  &  \\
 & \multirow{3}{*}{0.6} & \textsc{mean} & 279.96 (-0.62\%) & 279.04 (-0.94\%) & 275.84 (-2.08\%) & 271.66 (-3.56\%) & 267.78 (-4.94\%) \\
 &  & \textsc{var} & 281.3 (-0.14\%) & 281.12 (-0.21\%) & 280.2 (-0.53\%) & 280.18 (-0.54\%) & 280.3 (-0.5\%) \\
 &  & \textsc{mean}+\textsc{var} & 279.72 (-0.7\%) & 278.62 (-1.09\%) & 274.72 (-2.48\%) & 270.26 (-4.06\%) & 266.8 (-5.29\%) \\
 &  &  &  &  &  &  &  \\
 & \multirow{3}{*}{0.8} & \textsc{mean} & 266.75 (-0.24\%) & 265.85 (-0.58\%) & 263.8 (-1.35\%) & 261.45 (-2.23\%) & 259.48 (-2.96\%) \\
 &  & \textsc{var} & 267.18 (-0.08\%) & 267.08 (-0.12\%) & 267.01 (-0.15\%) & 266.81 (-0.22\%) & 266.9 (-0.19\%) \\
 &  & \textsc{mean}+\textsc{var} & 266.41 (-0.37\%) & 265.68 (-0.64\%) & 263.61 (-1.42\%) & 260.67 (-2.52\%) & 259.81 (-2.84\%) \\
 &  &  &  &  &  &  &  \\
 & \multirow{3}{*}{1.0} & \textsc{mean} & 251.78 (-0.29\%) & 251.18 (-0.52\%) & 249.06 (-1.36\%) & 246.93 (-2.21\%) & 245.54 (-2.76\%) \\
 &  & \textsc{var} & 252.22 (-0.11\%) & 251.97 (-0.21\%) & 251.67 (-0.33\%) & 251.81 (-0.27\%) & 251.95 (-0.22\%) \\
 &  & \textsc{mean}+\textsc{var} & 251.32 (-0.47\%) & 250.74 (-0.7\%) & 248.81 (-1.46\%) & 246.53 (-2.36\%) & 245 (-2.97\%) \\
\multicolumn{1}{l}{} & \multicolumn{1}{l}{} & \multicolumn{1}{l}{} & \multicolumn{1}{l}{} & \multicolumn{1}{l}{} & \multicolumn{1}{l}{} & \multicolumn{1}{l}{} & \multicolumn{1}{l}{} \\ \hline
\multirow{20}{*}{\textbf{TTA}} & \multicolumn{1}{l}{} & \multicolumn{1}{l}{} & \multicolumn{1}{l}{} & \multicolumn{1}{l}{} & \multicolumn{1}{l}{} & \multicolumn{1}{l}{} & \multicolumn{1}{l}{} \\
 & \multirow{3}{*}{0.2} & \textsc{mean} & 350.36 (-0.19\%) & 349.91 (-0.31\%) & 348.44 (-0.73\%) & 345.99 (-1.43\%) & 343.69 (-2.09\%) \\
 &  & \textsc{var} & 350.74 (-0.08\%) & 350.2 (-0.23\%) & 349.61 (-0.4\%) & 348.48 (-0.72\%) & 347.84 (-0.9\%) \\
 &  & \textsc{mean}+\textsc{var} & 349.56 (-0.41\%) & 348.76 (-0.64\%) & 345.71 (-1.51\%) & 341.31 (-2.76\%) & 340.66 (-2.95\%) \\
 & \multicolumn{1}{l}{} &  &  & \textbf{} & \textbf{} & \textbf{} & \textbf{} \\
 & \multirow{3}{*}{0.4} & \textsc{mean} & 207.1 (-0.59\%) & 206.13 (-1.06\%) & 203.53 (-2.3\%) & 200.9 (-3.56\%) & 198.93 (-4.51\%) \\
 &  & \textsc{var} & 207.94 (-0.19\%) & 207.68 (-0.31\%) & 207.14 (-0.57\%) & 205.89 (-1.17\%) & 205.13 (-1.54\%) \\
 &  & \textsc{mean}+\textsc{var} & 206.03 (-1.12\%) & 205.63 (-1.3\%) & 202.3 (-2.89\%) & 199.08 (-4.48\%) & 198.3 (-4.81\%) \\
 & \multicolumn{1}{l}{} &  &  & \textbf{} & \textbf{} & \textbf{} & \textbf{} \\
 & \multirow{3}{*}{0.6} & \textsc{mean} & 197.06   (-0.61\%) & 196.26 (-1.02\%) & 193.37 (-2.47\%) & 190.44 (-3.95\%) & 188.97   (-4.69\%) \\
 &  & \textsc{var} & 197.57 (-0.35\%) & 197.04 (-0.62\%) & 195.78 (-1.26\%) & 194.24 (-2.03\%) & 193.8 (-2.25\%) \\
 &  & \textsc{mean}+\textsc{var} & 196.42 (-0.93\%) & 195.81 (-1.24\%) & 193.36 (-2.48\%) & 189.72 (-4.31\%) & 188.44 (-4.96\%) \\
 & \multicolumn{1}{l}{} &  &  & \textbf{} & \textbf{} & \textbf{} & \textbf{} \\
 & \multirow{3}{*}{0.8} & \textsc{mean} & 202.48 (-0.49\%) & 201.2 (-1.12\%) & 200.15 (-1.63\%) & 196.4 (-3.48\%) & 193.8 (-4.75\%) \\
 &  & \textsc{var} & 203.16 (-0.15\%) & 202.96 (-0.25\%) & 201.73 (-0.86\%) & 200.19 (-1.62\%) & 198.74 (-2.33\%) \\
 &  & \textsc{mean}+\textsc{var} & 201.93 (-0.76\%) & 201 (-1.22\%) & 197.73 (-2.83\%) & 196.83 (-3.27\%) & 194.53 (-4.4\%) \\
 & \multicolumn{1}{l}{} &  &  & \textbf{} & \textbf{} & \textbf{} & \textbf{} \\
 & \multirow{3}{*}{1.0} & \textsc{mean} & 162.23 (-0.55\%) & 161.28 (-1.13\%) & 160.1 (-1.85\%) & 156.98 (-3.77\%) & 156.33 (-4.17\%) \\
 &  & \textsc{var} & 162.58 (-0.34\%) & 162.18 (-0.58\%) & 161.74 (-0.85\%) & 160.61 (-1.54\%) & 159.61 (-2.15\%) \\
 &  & \textsc{mean}+\textsc{var} & 161.85 (-0.78\%) & 161.18 (-1.2\%) & 159.95 (-1.95\%) & 157.23 (-3.62\%) & 157.35 (-3.54\%) \\
 & \multicolumn{1}{l}{} &  &  & \textbf{} & \textbf{} & \textbf{} & \textbf{} \\\hline
\end{tabular}%
}
\end{table}

\begin{figure}[p]
  \centering
  \begin{subfigure}[t]{0.19\linewidth}
    \centering
    \includegraphics[height=1.5cm]{img/beta/10238_left.png}
    \includegraphics[height=2.6cm]{img/beta/10238_left_deep.pdf}
    \includegraphics[height=2.6cm]{img/beta/10238_left_MC.pdf}
    \includegraphics[height=2.6cm]{img/beta/10238_left_TTA.pdf}
    \caption{\texttt{10238\_left} (SL0)}
  \end{subfigure}
  \begin{subfigure}[t]{0.19\linewidth}
    \centering
    \includegraphics[height=1.5cm]{img/beta/3017_left.png}
    \includegraphics[height=2.6cm]{img/beta/3017_left_deep.pdf}
    \includegraphics[height=2.6cm]{img/beta/3017_left_MC.pdf}
    \includegraphics[height=2.6cm]{img/beta/3017_left_TTA.pdf}
    \caption{\texttt{3017\_left} (SL1)}
  \end{subfigure}
  \begin{subfigure}[t]{0.19\linewidth}
    \centering
    \includegraphics[height=1.5cm]{img/beta/1290_left.png}
    \includegraphics[height=2.6cm]{img/beta/1290_left_deep.pdf}
    \includegraphics[height=2.6cm]{img/beta/1290_left_MC.pdf}
    \includegraphics[height=2.6cm]{img/beta/1290_left_TTA.pdf}
    \caption{\texttt{1290\_left} (SL2)}
  \end{subfigure}
  \begin{subfigure}[t]{0.19\linewidth}
    \centering
    \includegraphics[height=1.5cm]{img/beta/509_left.png}
    \includegraphics[height=2.6cm]{img/beta/509_left_deep.pdf}
    \includegraphics[height=2.6cm]{img/beta/509_left_MC.pdf}
    \includegraphics[height=2.6cm]{img/beta/509_left_TTA.pdf}
    \caption{\texttt{509\_left} (SL3)}
  \end{subfigure}
  \begin{subfigure}[t]{0.19\linewidth}
    \centering
    \includegraphics[height=1.5cm]{img/beta/2338_left.png}
    \includegraphics[height=2.6cm]{img/beta/2338_left_deep.pdf}
    \includegraphics[height=2.6cm]{img/beta/2338_left_MC.pdf}
    \includegraphics[height=2.6cm]{img/beta/2338_left_TTA.pdf}
    \caption{\texttt{2338\_left} (SL4)}
  \end{subfigure}

  \begin{subfigure}[t]{0.19\linewidth}
    \centering
    \includegraphics[height=1.5cm]{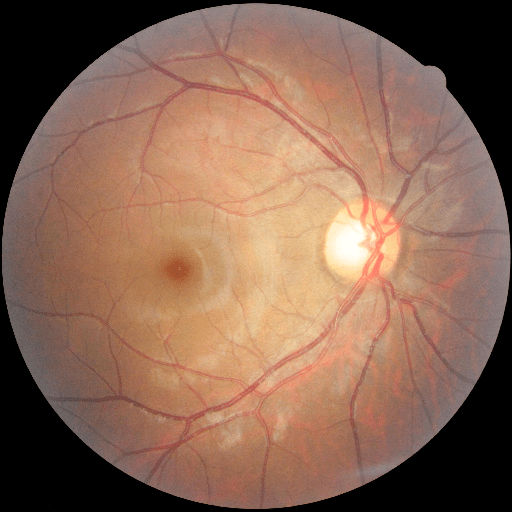}
    \includegraphics[height=2.6cm]{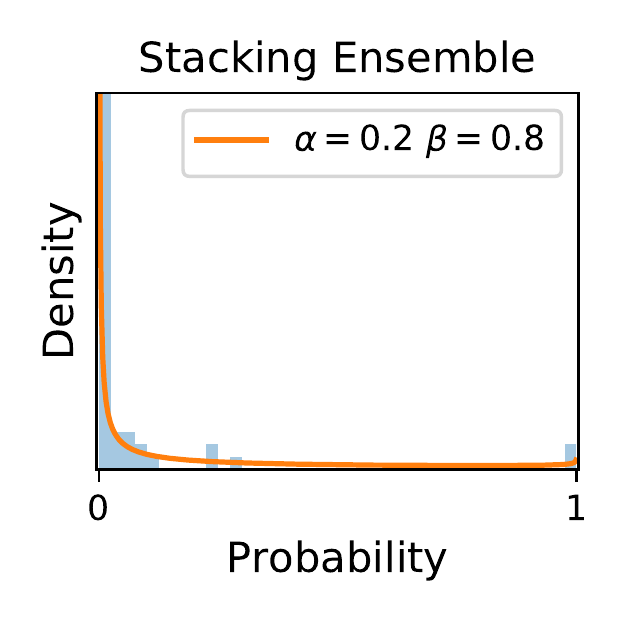}
    \includegraphics[height=2.6cm]{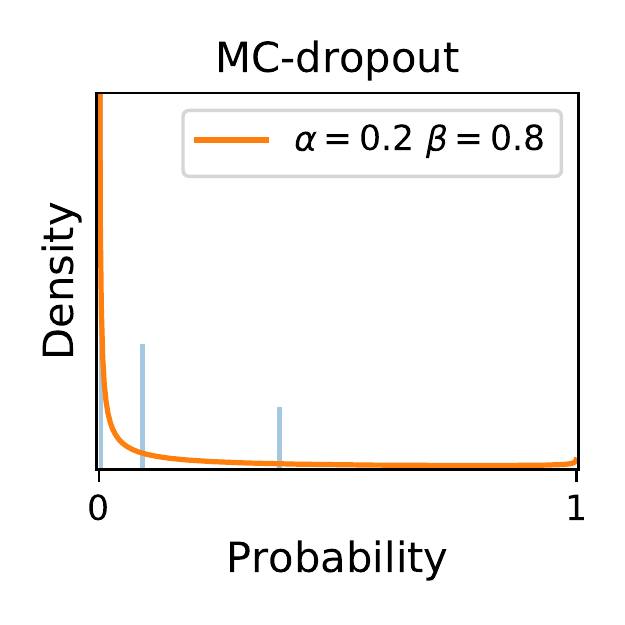}
    \includegraphics[height=2.6cm]{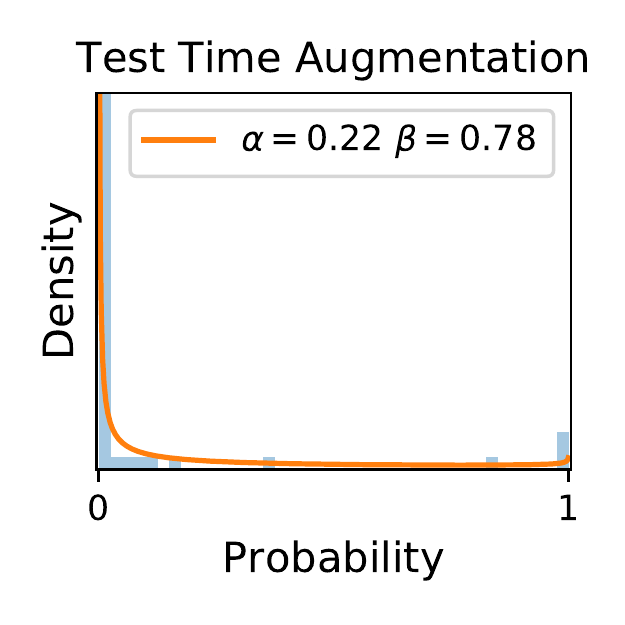}
    \caption{\texttt{13282\_right} (SL0)}
  \end{subfigure}
  \begin{subfigure}[t]{0.19\linewidth}
    \centering
    \includegraphics[height=1.5cm]{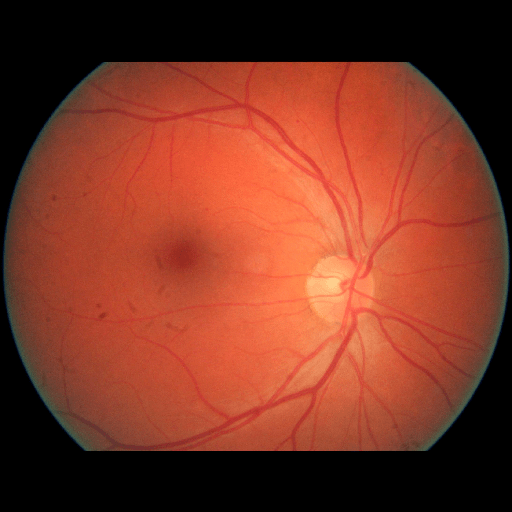}
    \includegraphics[height=2.6cm]{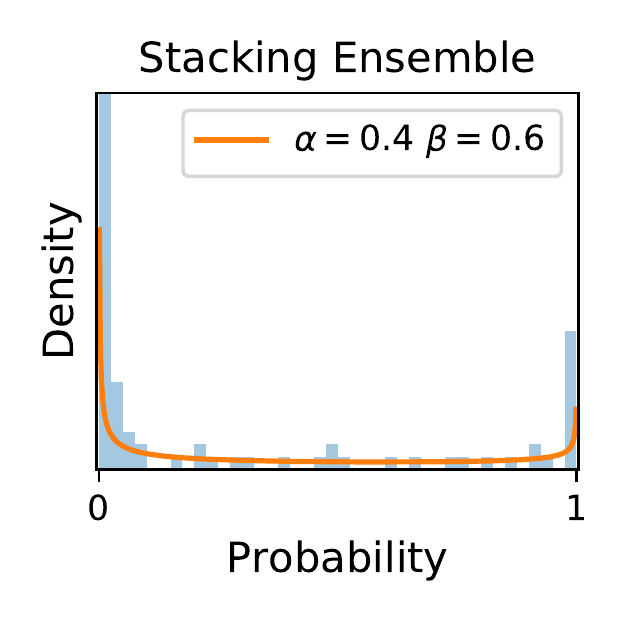}
    \includegraphics[height=2.6cm]{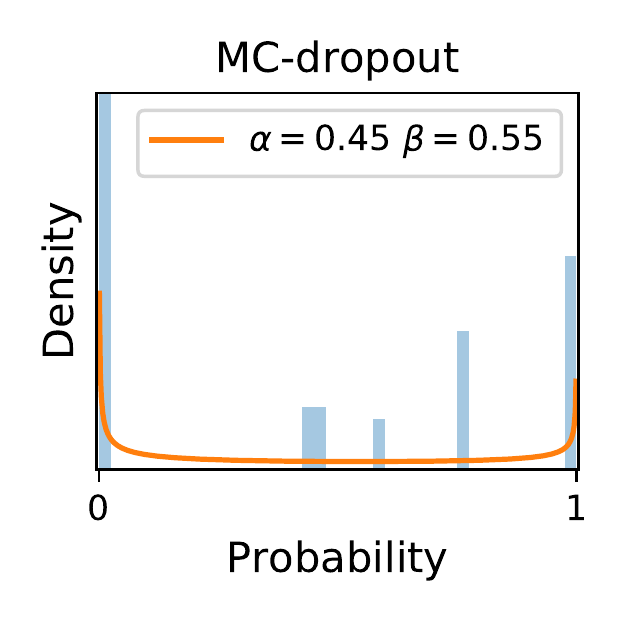}
    \includegraphics[height=2.6cm]{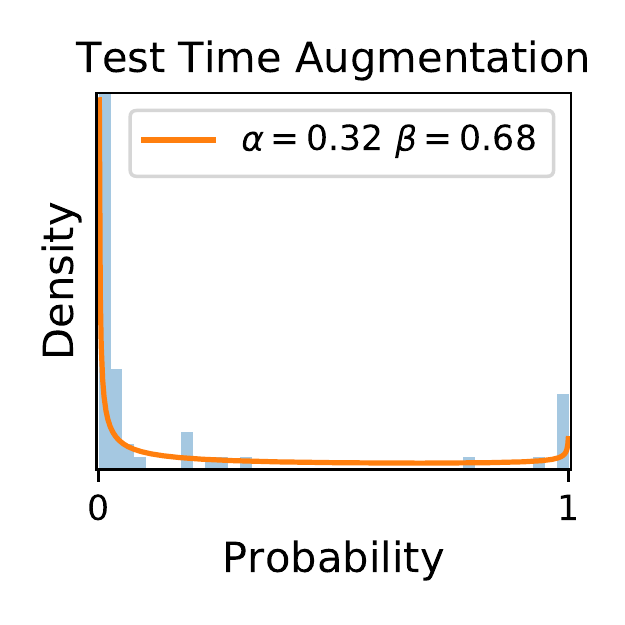}
    \caption{\texttt{3061\_left} (SL1)}
  \end{subfigure}
  \begin{subfigure}[t]{0.19\linewidth}
    \centering
    \includegraphics[height=1.5cm]{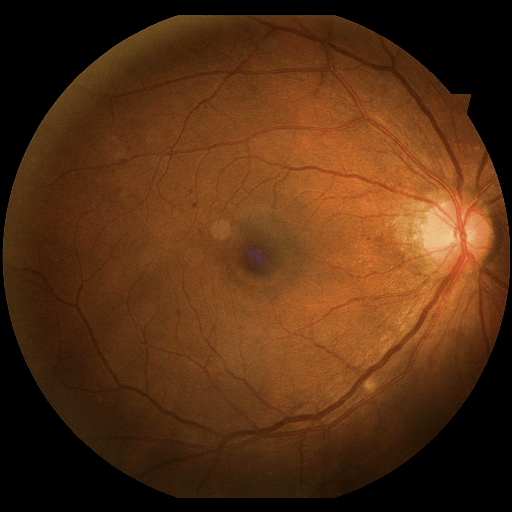}
    \includegraphics[height=2.6cm]{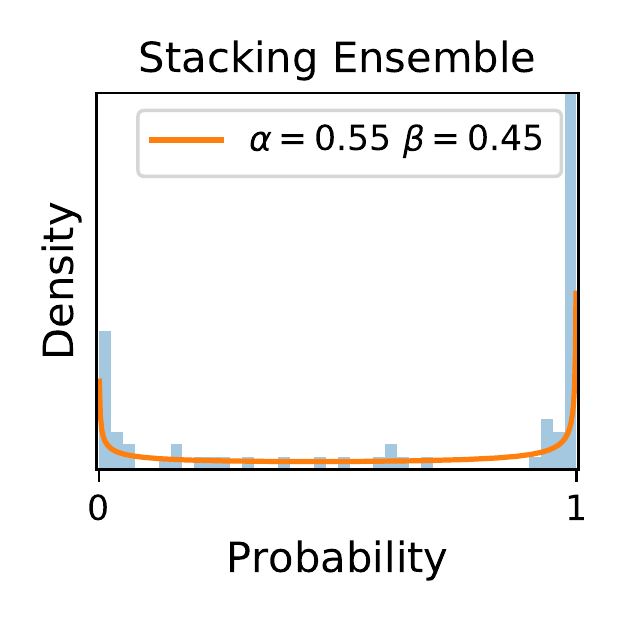}
    \includegraphics[height=2.6cm]{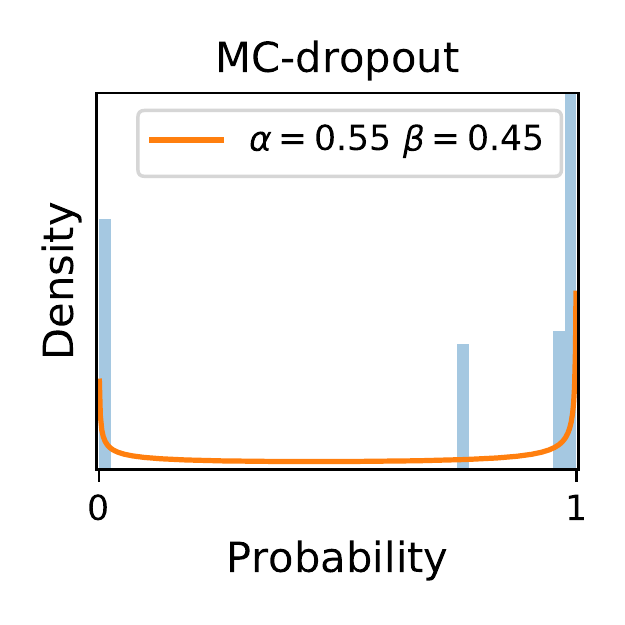}
    \includegraphics[height=2.6cm]{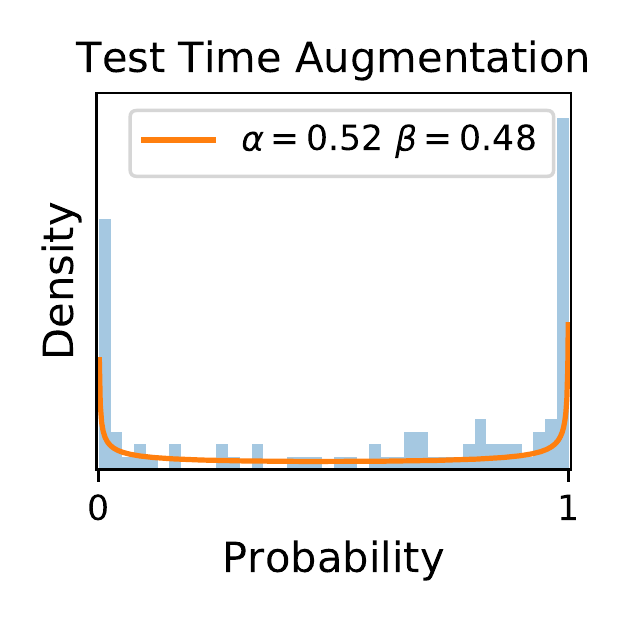}
    \caption{\texttt{1296\_right} (SL2)}
  \end{subfigure}
  \begin{subfigure}[t]{0.19\linewidth}
    \centering
    \includegraphics[height=1.5cm]{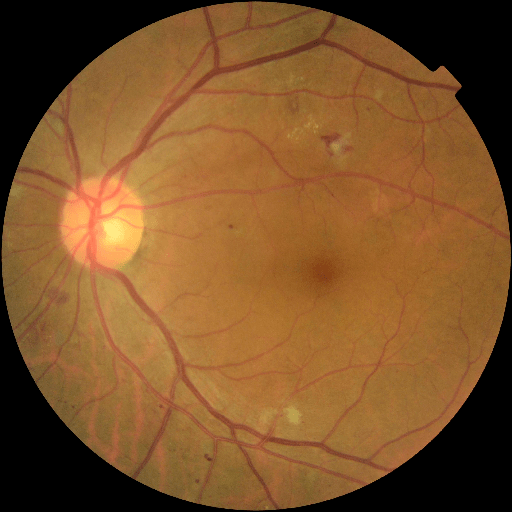}
    \includegraphics[height=2.6cm]{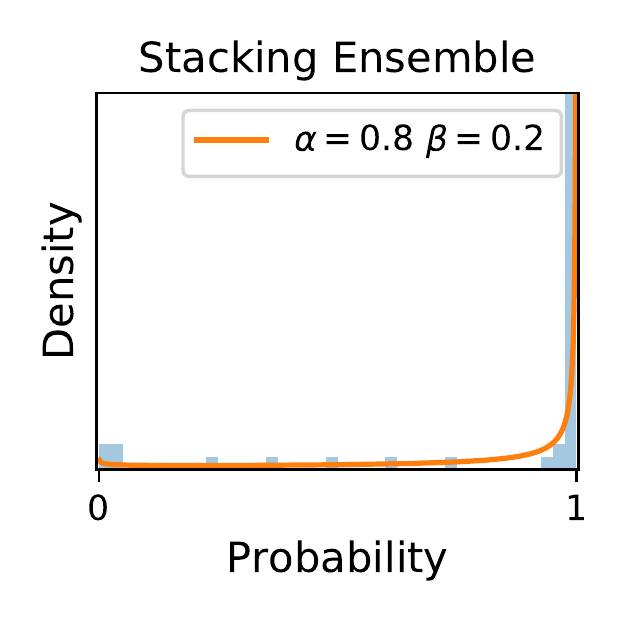}
    \includegraphics[height=2.6cm]{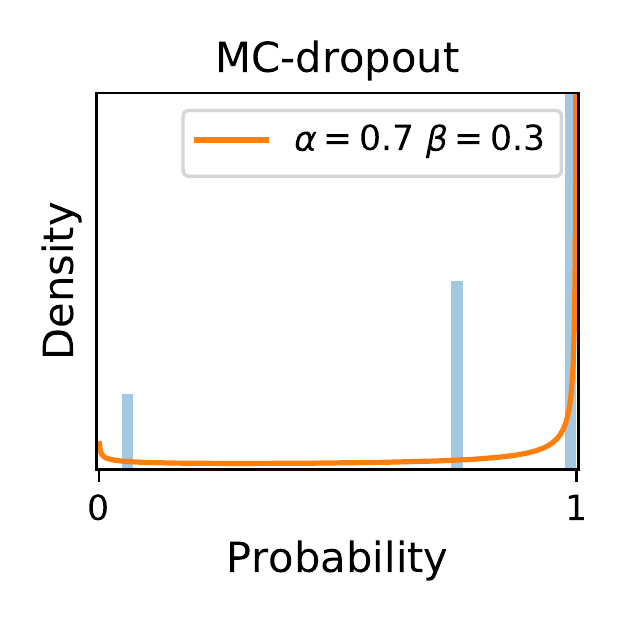}
    \includegraphics[height=2.6cm]{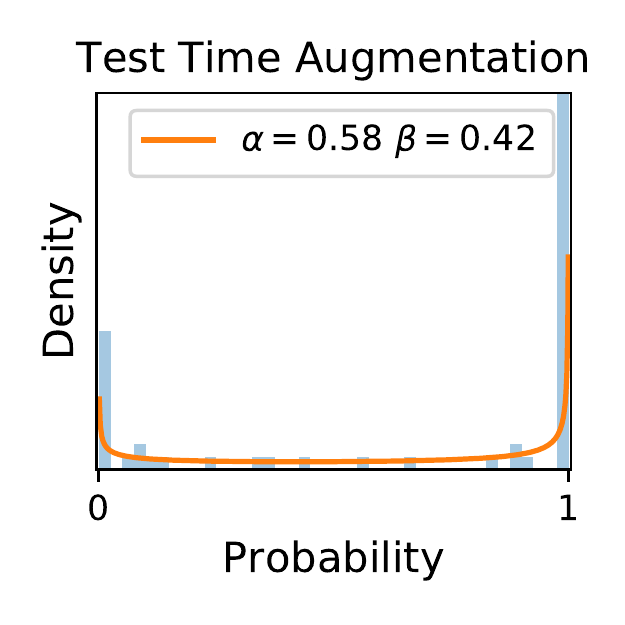}
    \caption{\texttt{5964\_left} (SL3)}
  \end{subfigure}
  \begin{subfigure}[t]{0.19\linewidth}
    \centering
    \includegraphics[height=1.5cm]{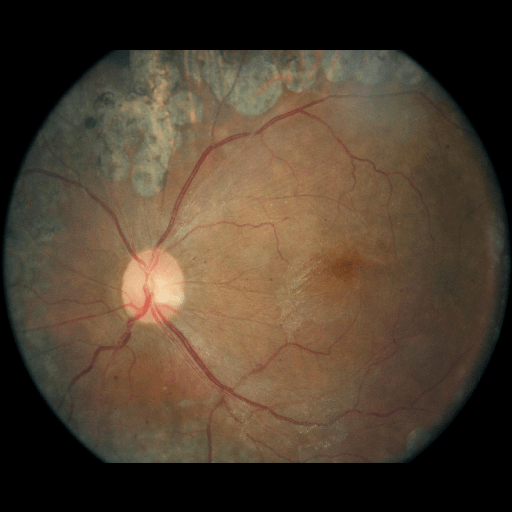}
    \includegraphics[height=2.6cm]{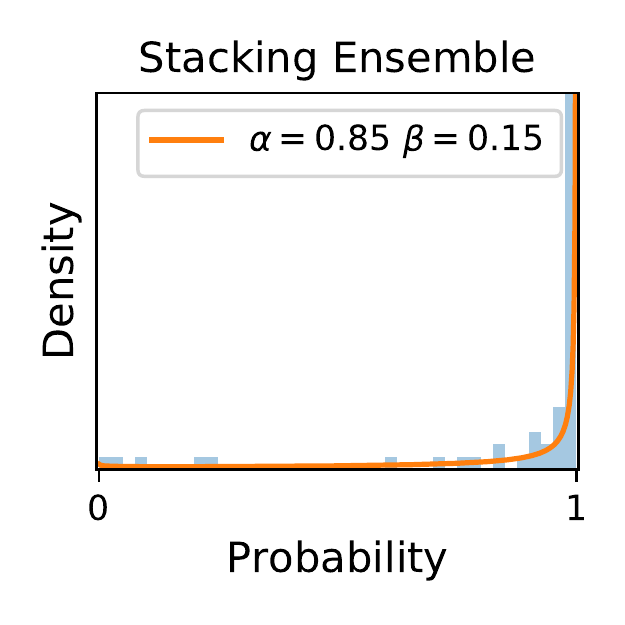}
    \includegraphics[height=2.6cm]{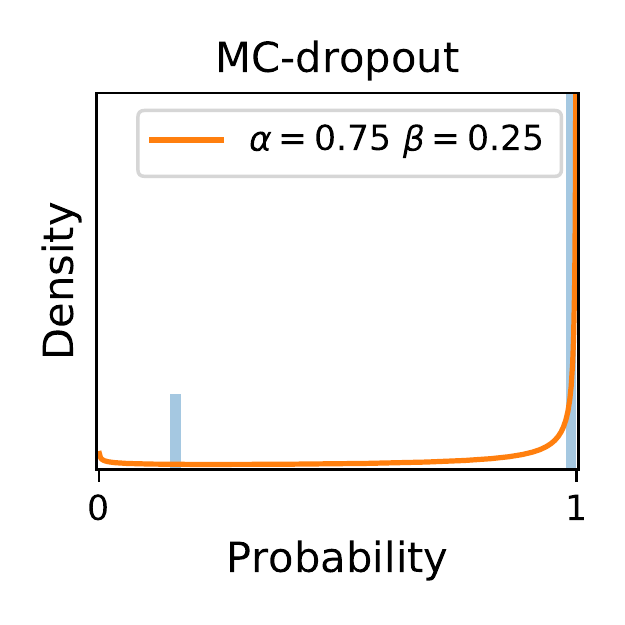}
    \includegraphics[height=2.6cm]{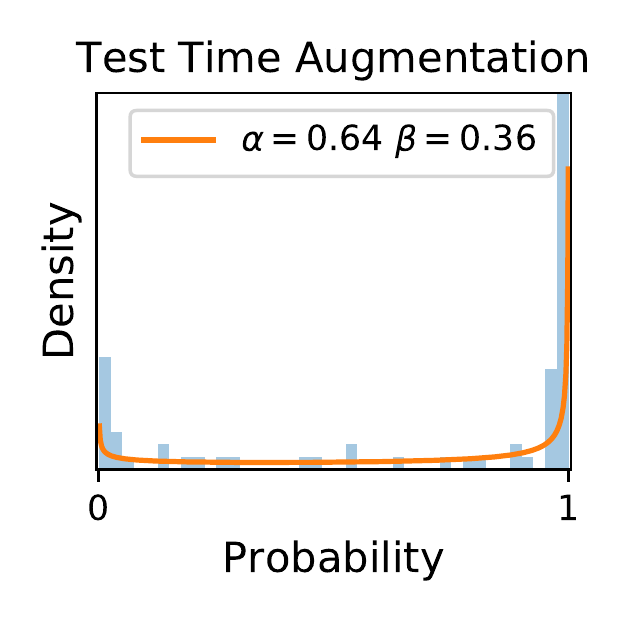}
    \caption{\texttt{16617\_right} (SL4)}
  \end{subfigure}
\end{figure}

\begin{figure}[p]
  \ContinuedFloat
  \centering
  \begin{subfigure}[t]{0.19\linewidth}
    \centering
    \includegraphics[height=1.5cm]{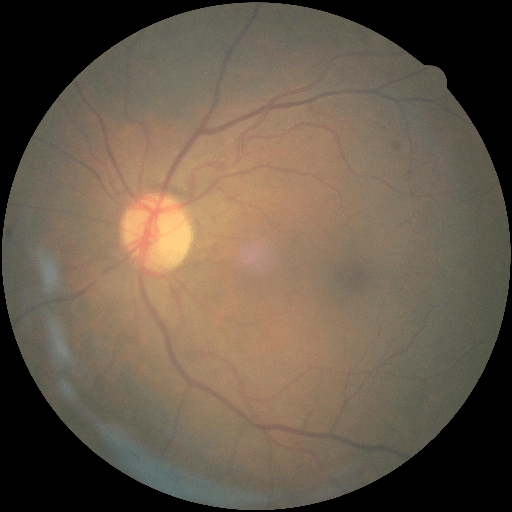}
    \includegraphics[height=2.6cm]{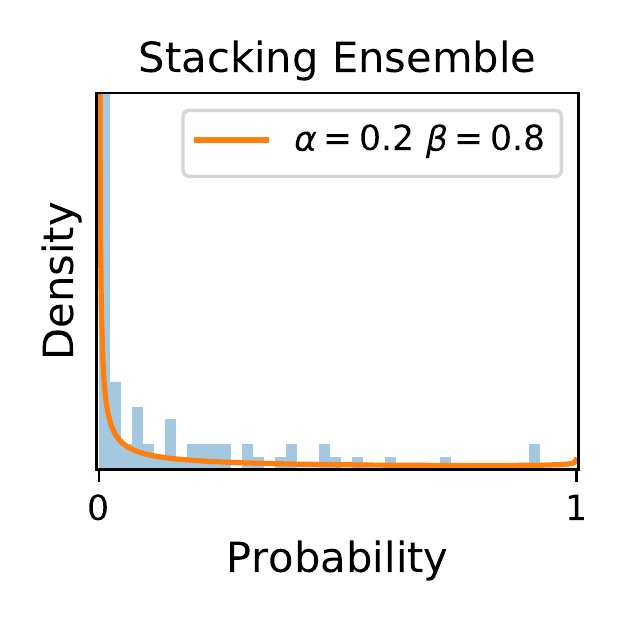}
    \includegraphics[height=2.6cm]{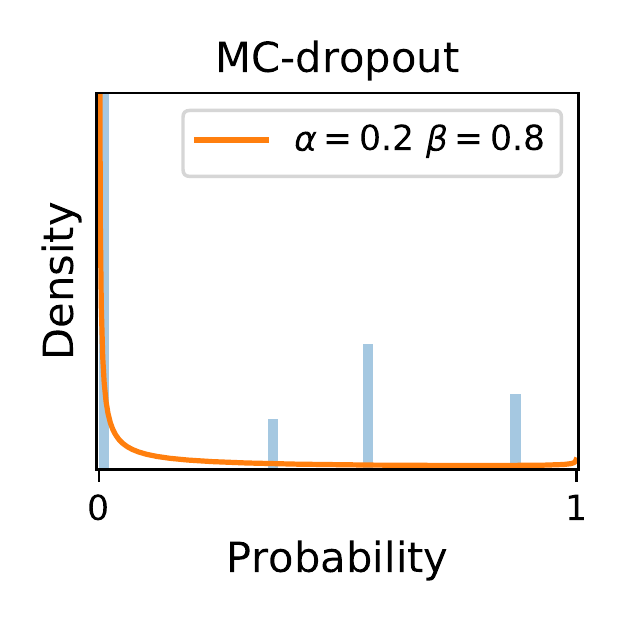}
    \includegraphics[height=2.6cm]{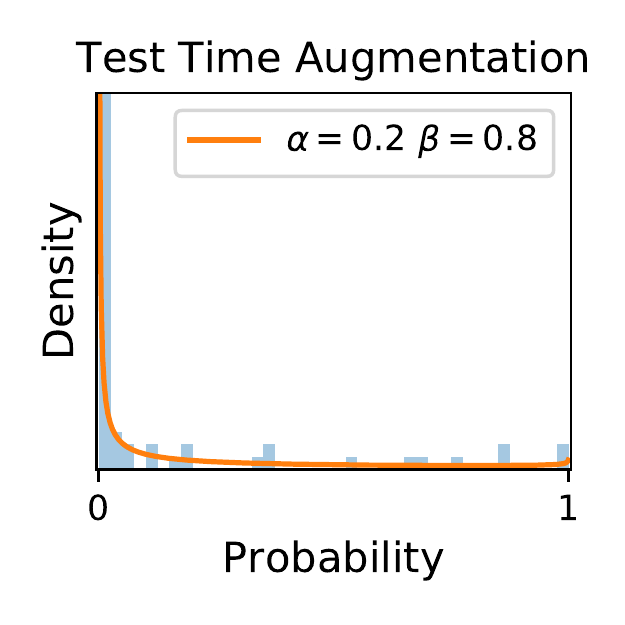}
    \caption{\texttt{2840\_left} (SL0)}
  \end{subfigure}
  \begin{subfigure}[t]{0.19\linewidth}
    \centering
    \includegraphics[height=1.5cm]{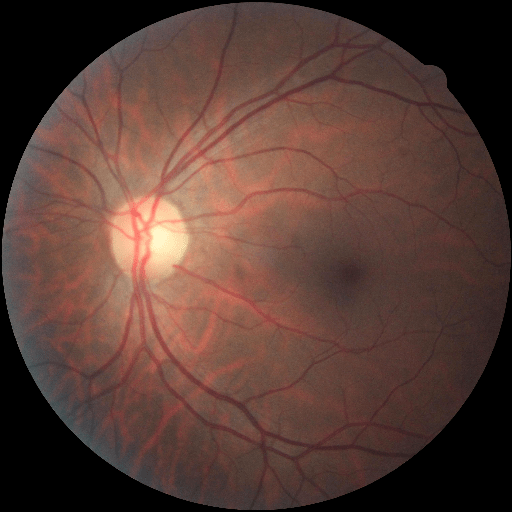}
    \includegraphics[height=2.6cm]{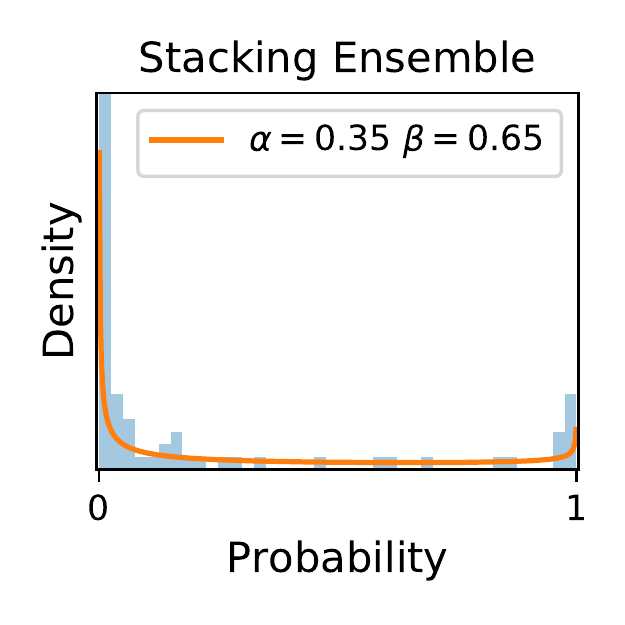}
    \includegraphics[height=2.6cm]{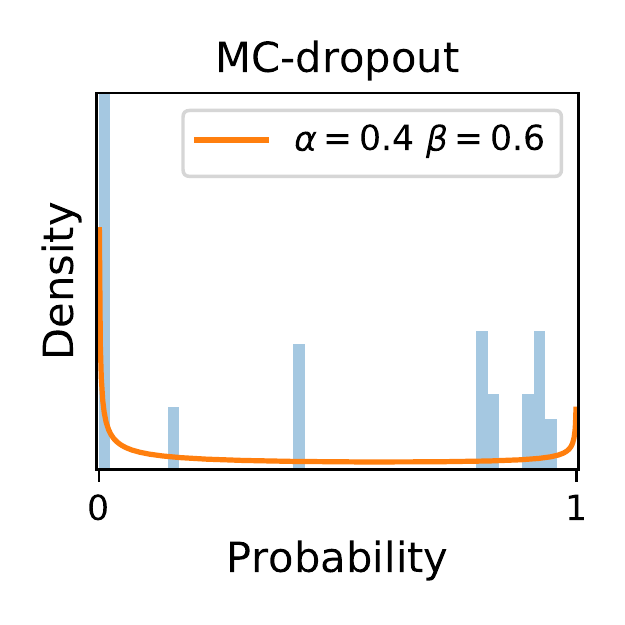}
    \includegraphics[height=2.6cm]{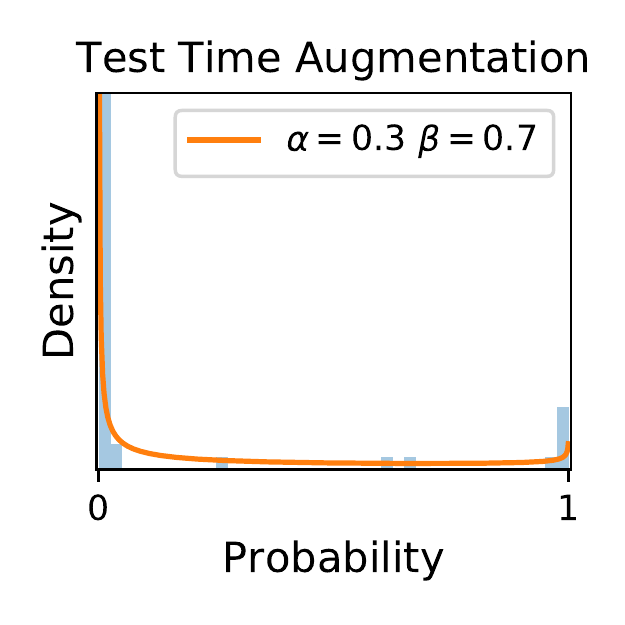}
    \caption{\texttt{3124\_left} (SL1)}
  \end{subfigure}
  \begin{subfigure}[t]{0.19\linewidth}
    \centering
    \includegraphics[height=1.5cm]{img/beta/1296_right.png}
    \includegraphics[height=2.6cm]{img/beta/1296_right_deep.pdf}
    \includegraphics[height=2.6cm]{img/beta/1296_right_MC.pdf}
    \includegraphics[height=2.6cm]{img/beta/1296_right_TTA.pdf}
    \caption{\texttt{1296\_right} (SL2)}
  \end{subfigure}
  \begin{subfigure}[t]{0.19\linewidth}
    \centering
    \includegraphics[height=1.5cm]{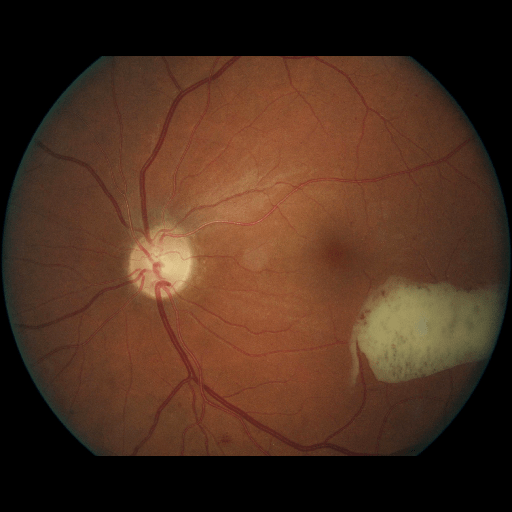}
    \includegraphics[height=2.6cm]{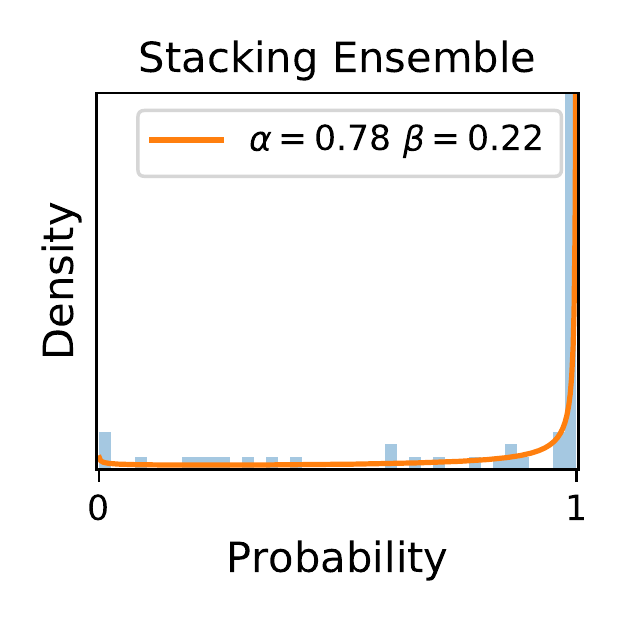}
    \includegraphics[height=2.6cm]{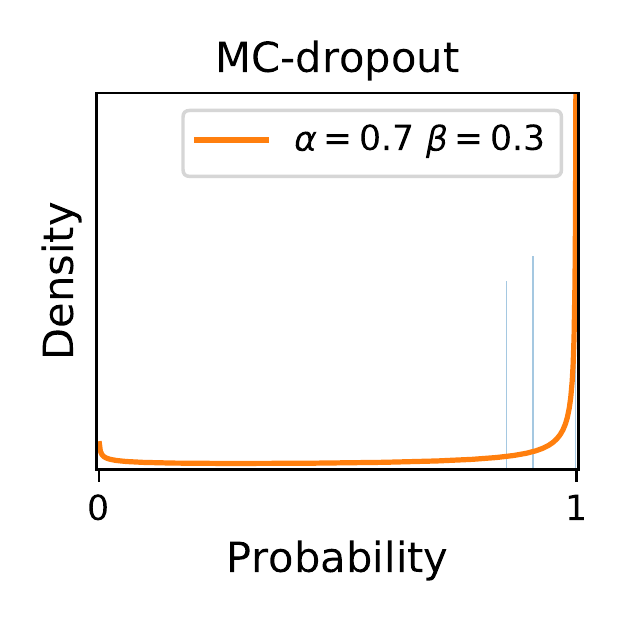}
    \includegraphics[height=2.6cm]{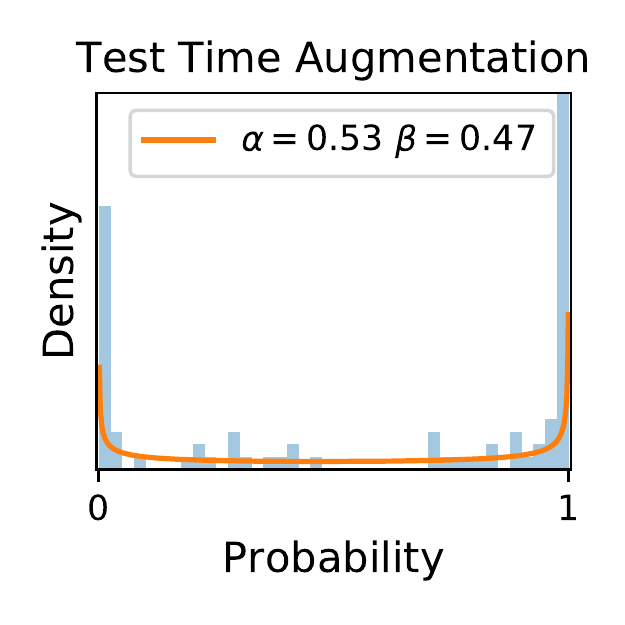}
    \caption{\texttt{21975\_right} (SL3)}
  \end{subfigure}
  \begin{subfigure}[t]{0.19\linewidth}
    \centering
    \includegraphics[height=1.5cm]{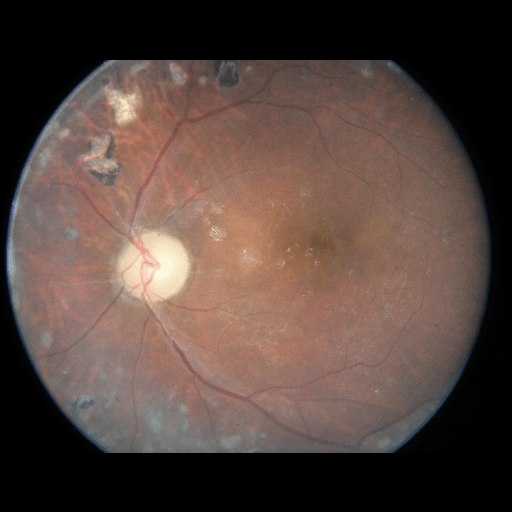}
    \includegraphics[height=2.6cm]{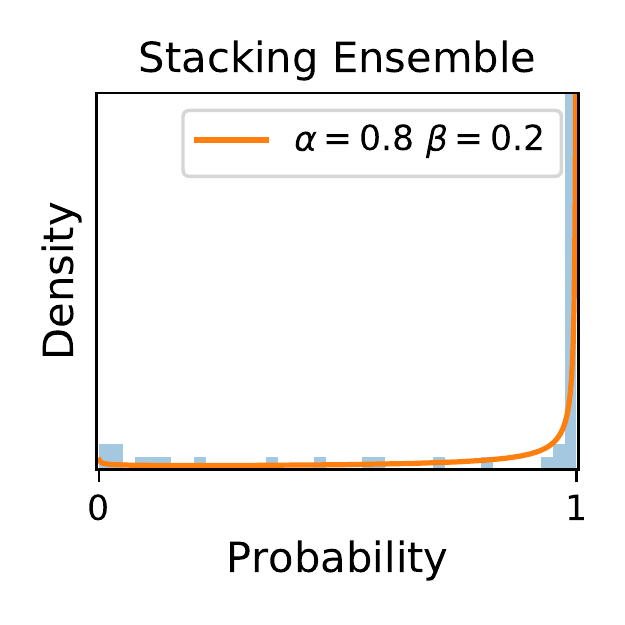}
    \includegraphics[height=2.6cm]{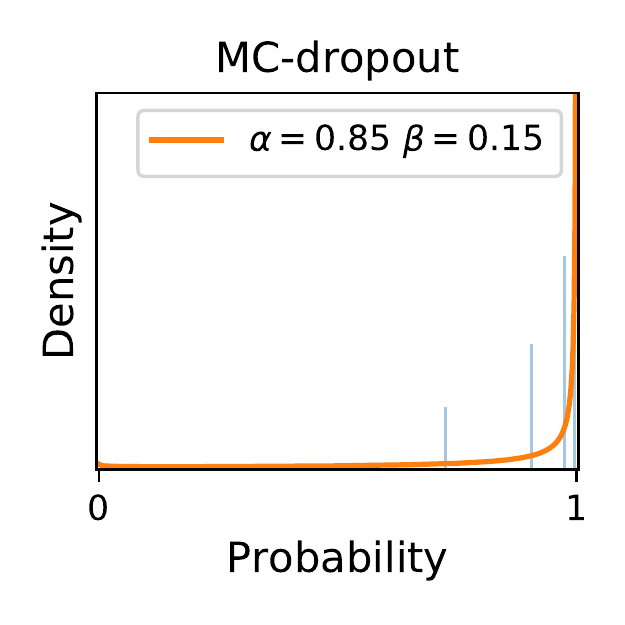}
    \includegraphics[height=2.6cm]{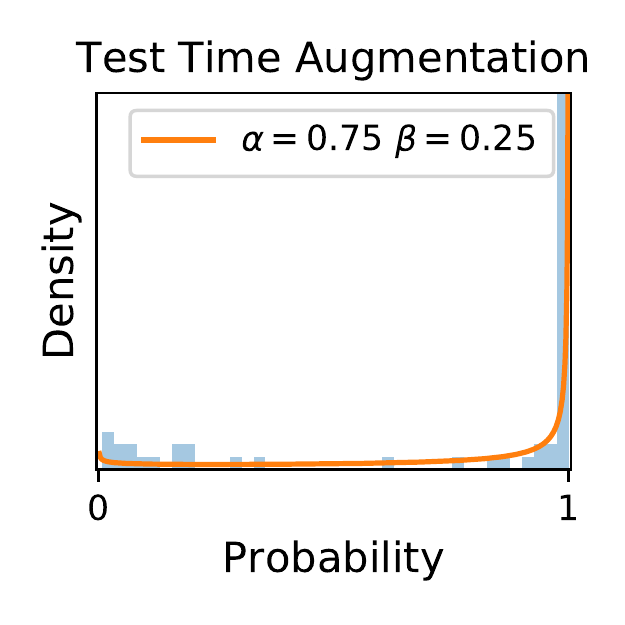}
    \caption{\texttt{34543\_right} (SL4)}
  \end{subfigure}
  
  \begin{subfigure}[t]{0.19\linewidth}
    \centering
    \includegraphics[height=1.5cm]{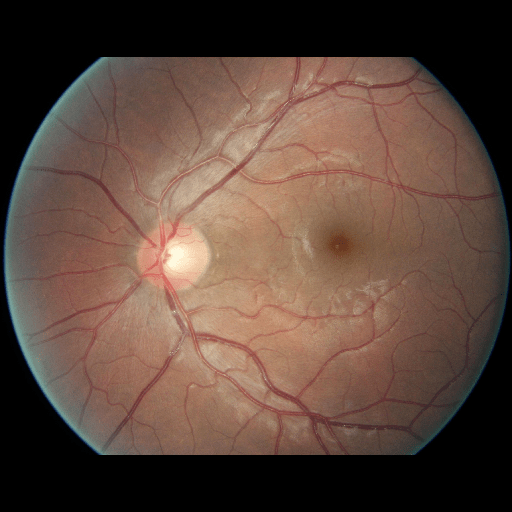}
    \includegraphics[height=2.6cm]{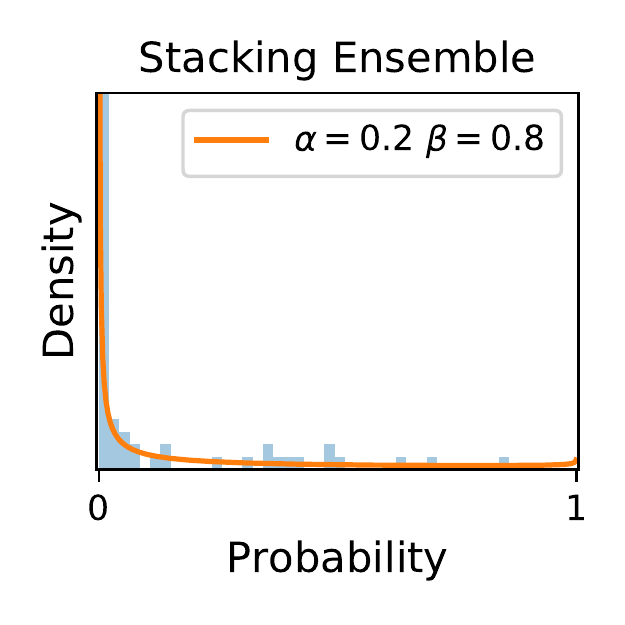}
    \includegraphics[height=2.6cm]{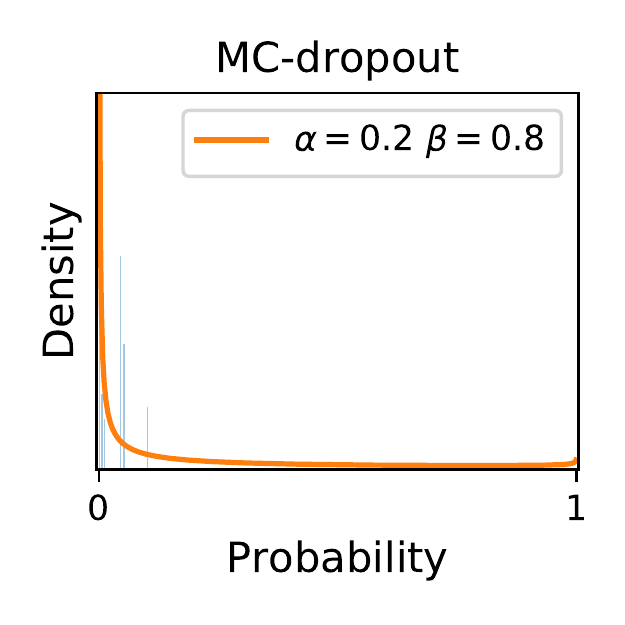}
    \includegraphics[height=2.6cm]{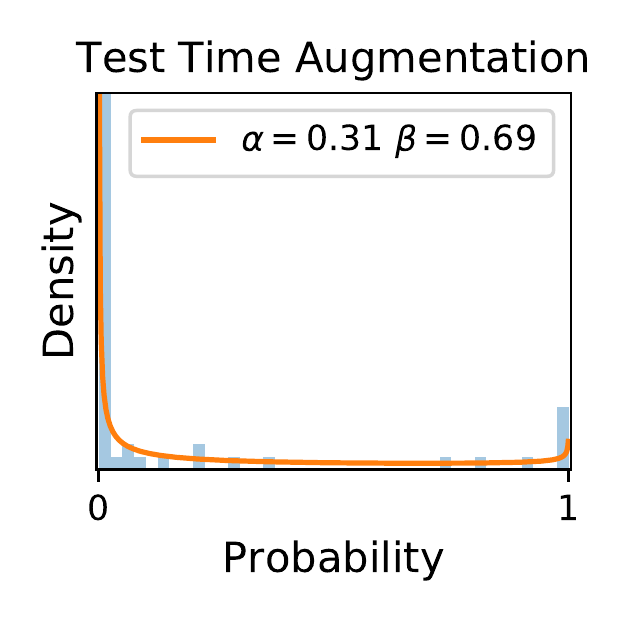}
    \caption{\texttt{2737\_right} (SL0)}
  \end{subfigure}
  \begin{subfigure}[t]{0.19\linewidth}
    \centering
    \includegraphics[height=1.5cm]{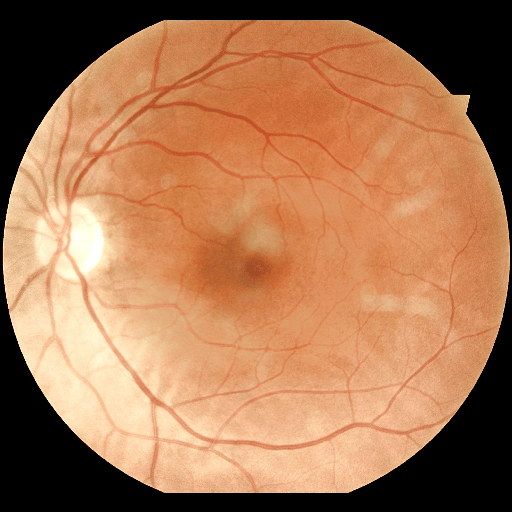}
    \includegraphics[height=2.6cm]{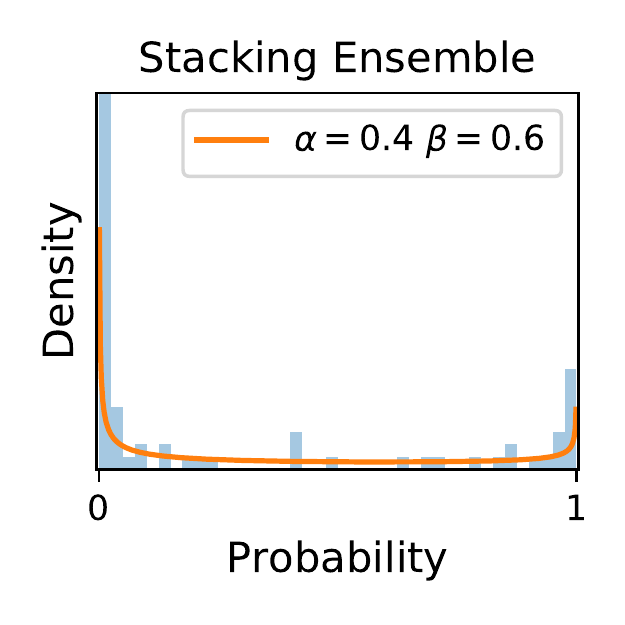}
    \includegraphics[height=2.6cm]{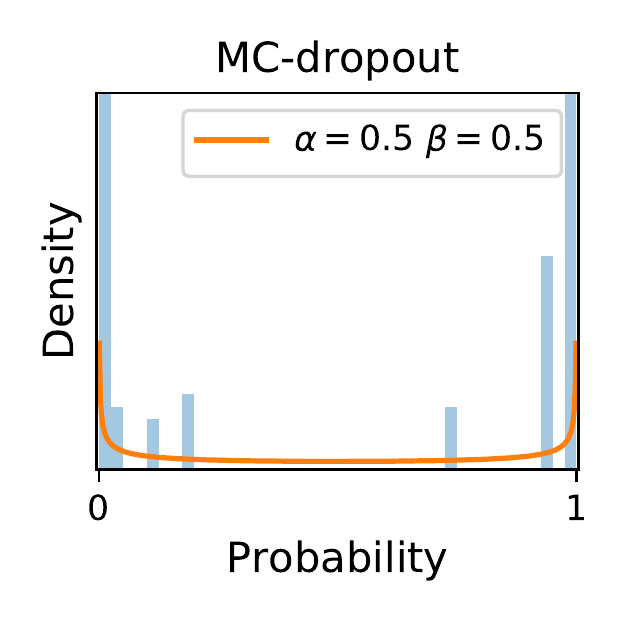}
    \includegraphics[height=2.6cm]{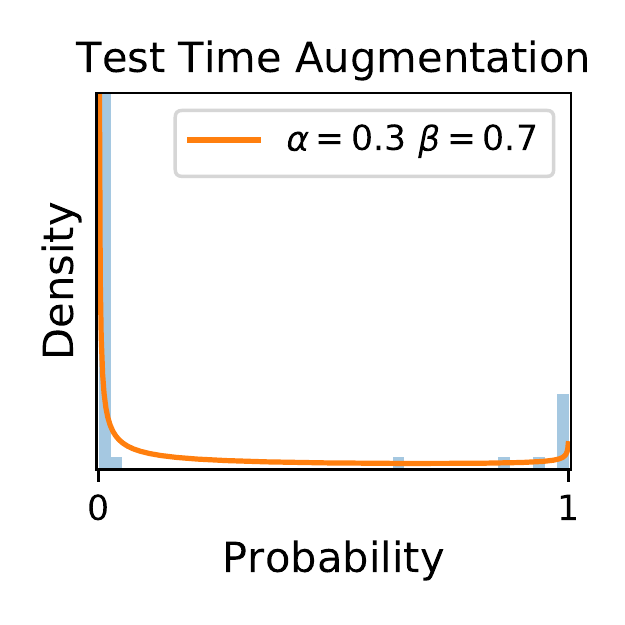}
    \caption{\texttt{3605\_left} (SL1)}
  \end{subfigure}
  \begin{subfigure}[t]{0.19\linewidth}
    \centering
    \includegraphics[height=1.5cm]{img/beta/1296_right.png}
    \includegraphics[height=2.6cm]{img/beta/1296_right_deep.pdf}
    \includegraphics[height=2.6cm]{img/beta/1296_right_MC.pdf}
    \includegraphics[height=2.6cm]{img/beta/1296_right_TTA.pdf}
    \caption{\texttt{1296\_right} (SL2)}
  \end{subfigure}
  \begin{subfigure}[t]{0.19\linewidth}
    \centering
    \includegraphics[height=1.5cm]{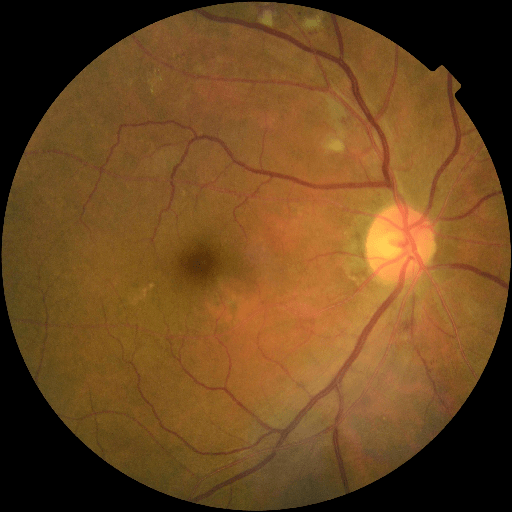}
    \includegraphics[height=2.6cm]{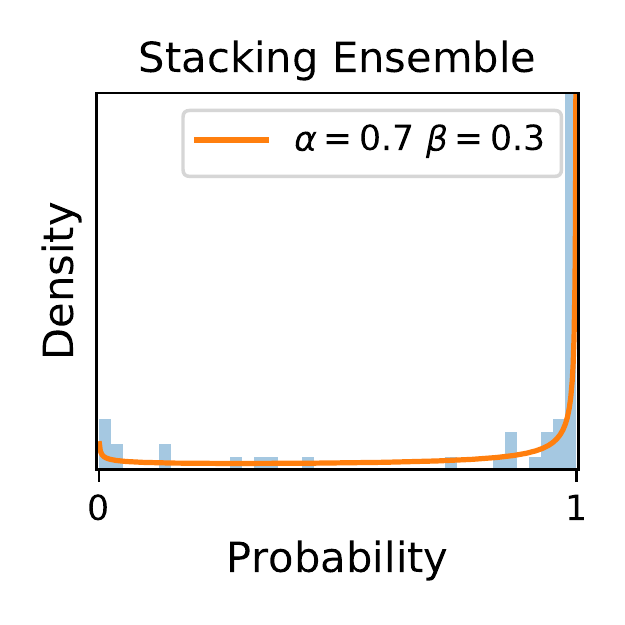}
    \includegraphics[height=2.6cm]{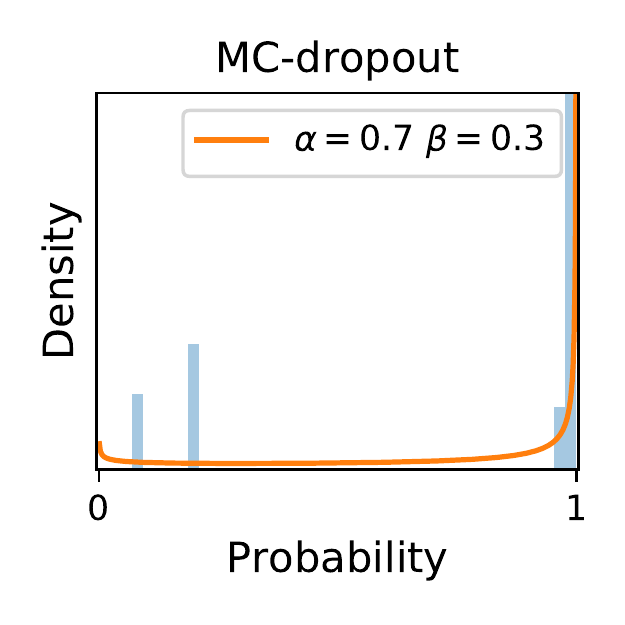}
    \includegraphics[height=2.6cm]{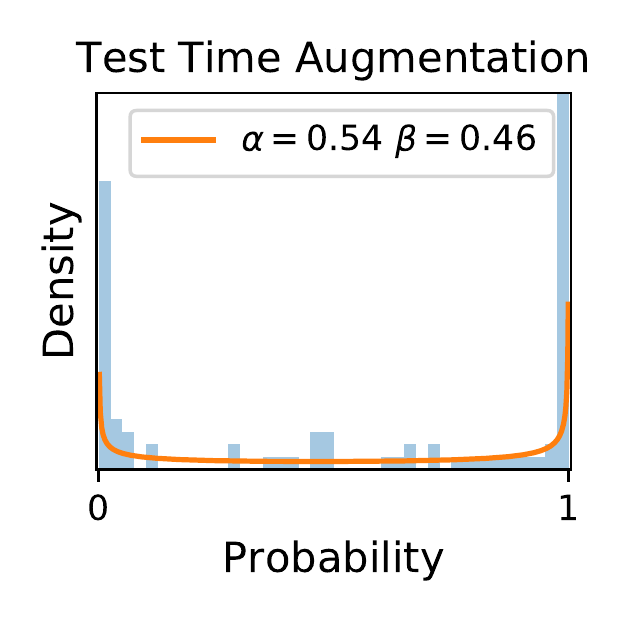}
    \caption{\texttt{7565\_right} (SL3)}
  \end{subfigure}
  \begin{subfigure}[t]{0.19\linewidth}
    \centering
    \includegraphics[height=1.5cm]{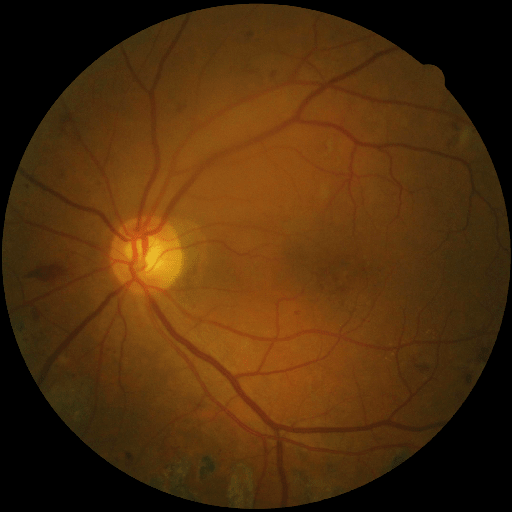}
    \includegraphics[height=2.6cm]{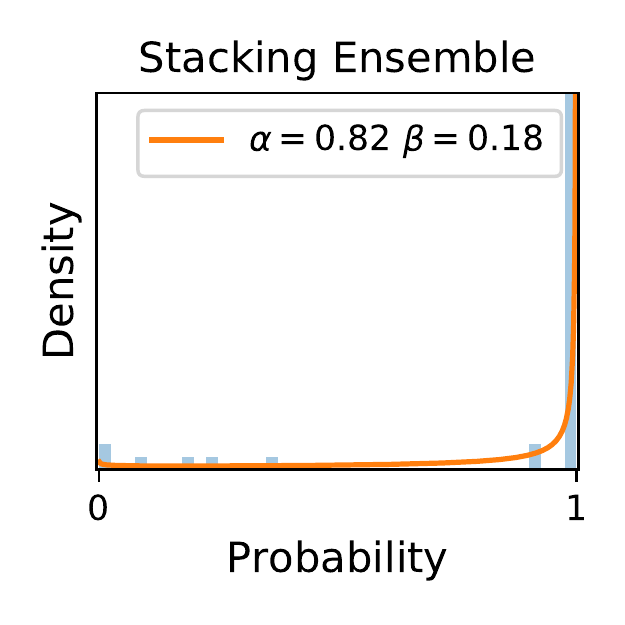}
    \includegraphics[height=2.6cm]{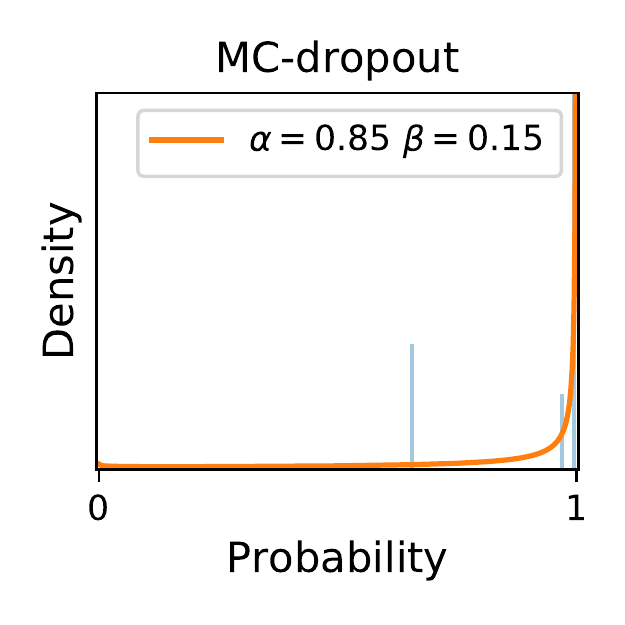}
    \includegraphics[height=2.6cm]{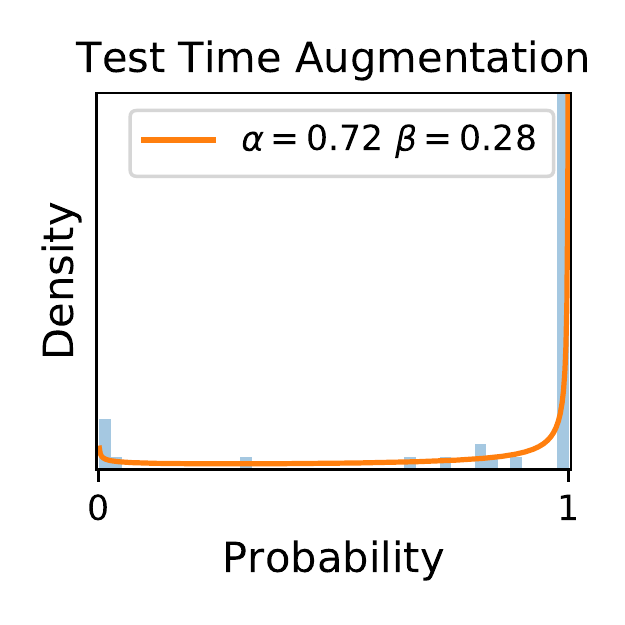}
    \caption{\texttt{23403\_left} (SL4)}
  \end{subfigure}
  \caption{The four rows of each panel correspond to original images, stacking ensemble, MC-dropout, \ac{TTA}, respectively. The orange curves indicate the beta distributions fitted to the distributions of single learners' predictions displayed by the histograms.}
  \label{fig:beta-dist-five-SL-plus}
\end{figure}

\end{document}